\documentclass[pdflatex, sn-mathphys-num, iicol]{sn-jnl}

\usepackage{graphicx}%
\usepackage{multirow}%
\usepackage{amsmath,amssymb,amsfonts}%
\usepackage{amsthm}%
\usepackage[title]{appendix}%
\usepackage{xcolor}%
\usepackage{textcomp}%
\usepackage{manyfoot}%
\usepackage{booktabs}%
\usepackage{algorithm}%
\usepackage{algorithmicx}%
\usepackage{algpseudocode}%
\usepackage{listings}%
\usepackage{caption}
\usepackage{subcaption}
\usepackage{xfrac}
\usepackage{nicefrac}
\usepackage{comment}
\usepackage{cuted}

\theoremstyle{thmstyleone}%
\theoremstyle{thmstyletwo}%

\theoremstyle{thmstylethree}%

\begin{document}

\title[SADFED Koopman Eigenfunction Identification]{Spatially Aware Dictionary-Free Koopman Eigenfunction Identification for Modeling and Control}


\author*[1]{\fnm{David} \sur{Grasev}}\email{david.grasev@unob.cz}

\affil*[1]{\orgdiv{Department of Aviation Technology}, \orgname{University of Defence}, \orgaddress{\city{Brno}, \state{Czech Republic}}}


\abstract{A spatially aware dictionary-free eigenfunction discovery (SADFED) framework is proposed for identification of low-rank Koopman models from data without prescribing a lifting dictionary, kernel, or neural-network eigenfunction architecture. A reference trajectory is selected and used to determine the Koopman modes by regularized least squares (LS). Then, a transformed temporal basis allows the eigenfunction values at all sampled initial conditions to be obtained by a second regularized LS projection. Consequently, only the real and imaginary parts of the eigenvalues remain as the optimization variables. Interpolation of the identified eigenfunction samples reveals their spatial structure, enabling numerical estimation of their gradients. A joint objective combines trajectory reconstruction error with a normalized Koopman partial differential equation (KPDE) residual, promoting spatial consistency with the KPDE over the sampled region and serving as a physics-informed regularizer. The method is evaluated on a system with analytical Koopman eigenfunctions, the FitzHugh-Nagumo system, the van der Pol oscillator, the Duffing system, and a two-spool turbojet engine. The examples demonstrate recovery of known eigenfunctions, sensitivity to reference trajectory and hyperparameters, limit-cycle harmonics and isochrons, discontinuous indicator eigenfunctions and isostables, symmetry exploitation, and construction of state-dependent lifted input dynamics. For the turbojet example, the identified model is further used for state estimation and design of a gain-scheduled tracking linear quadratic Gaussian controller. The results indicate the applicability of SADFED to Koopman spectral identification and control-oriented modeling of nonlinear dynamical systems.}


\maketitle

\section{Introduction}
Data-driven modeling has become an essential tool for the analysis and control of nonlinear systems, particularly in situations where the governing dynamics are unknown, incomplete, or difficult to derive. In many practical applications, high-fidelity physical models are either unavailable or too computationally demanding. By learning directly from measured or simulated data, data-driven methods provide an attractive alternative to modeling system behavior and predicting its evolution. Recent advances in sensing technologies, data storage, and computational power have further accelerated the adoption of these approaches across a wide range of scientific and engineering disciplines.

In recent years, the Koopman operator framework has gained popularity in the machine learning (ML) and nonlinear dynamics communities. Although the theory was originally developed in the 1930s by Koopman and von Neumann~\cite{Koopman1931, Koopman1932}, its widespread practical use has become feasible with the advent of modern computational resources~\cite{BruntonBook2019, Mauroy2020}. The central idea is to represent a nonlinear dynamical system through a transformation of its original state space into an infinite-dimensional space of observable functions (also called a lifted state space). In this space, the temporal evolution of observables is governed by a linear operator - the Koopman operator~\cite{Surana2020, Mauroy2020}. In practice, finite-dimensional approximations of this operator in the form of a Koopman matrix are required. An equivalent viewpoint is to seek a finite set of nonlinear observable functions that span a lifted state space in which the dynamics can be well approximated by a linear system. Ideally, the original state variables should lie within the span of the Koopman lifted state space, enabling their linear reconstruction. This linear structure can also facilitate the use of linear optimal control methods~\cite{Mauroy2020, BruntonBook2019, Brunton2021, Kaiser2021}.

Koopman identification techniques may be broadly divided into two main classes. The first class consists of extended dynamic mode decomposition (EDMD)-based methods, which rely on a predefined library of observable functions, typically constructed from polynomials or radial basis functions. The Koopman matrix is then obtained by solving a regularized least-squares (LS) regression problem on time-shifted data. The most common representative is naturally the classical EDMD, which yields a Koopman approximation optimized for one-step prediction~\cite{Williams2015}. Numerous variants have been proposed, including kernel DMD~\cite{Williams2015-2}, EDMD in reproducing kernel Hilbert spaces~\cite{Jiang2022, Ishikawa2024}, sparse DMD methods~\cite{Jovanovich2014, Schlosser2022}, generator EDMD~\cite{Klus2020}, residual DMD~\cite{Colbrook2023}, or symmetry-constrained EDMD for systems with multiple disjoint invariant sets~\cite{Pan2024}. The extension to controlled systems, EDMDc, was introduced in~\cite{Proctor2018}. A related family of approaches leverages time-delay embeddings and Hankel matrices, such as HAVOK~\cite{BruntonHAVOK, Li2024HAVOK}.

The primary challenge of DMD methods is the selection of observable functions for lifting. Additionally, to cope with strong nonlinearities, the libraries are often very large, scaling up to 100 functions or more. Lower-order solutions can be promoted using, e.g., the above-mentioned sparsity promotion~\cite{Jovanovich2014}, or via singular value decomposition and rank truncation~\cite{Williams2015}. Kernel-based EDMD replaces inner products in the classical EDMD by kernels, avoiding the explicit construction of an observable library~\cite{Williams2015-2}. However, the kernel type, its parameters, and the regularization settings must still be selected. Among other disadvantages of classical EDMD are spurious eigenfunctions, ill-conditioning~\cite{Kaiser2021}, and spectral pollution~\cite{Colbrook2023}.

Another option is to employ dictionary learning methods, where only the structure of observables is decided a priori. Observables can be represented, e.g., by neural networks, especially autoencoders~\cite{Takeishi2017, Lusch2018, Li2020, Jin2024}, which allow discovery of low-rank latent spaces that are conjugate to the underlying dynamics; a powerful concept for analysis of dynamical systems. Despite avoiding the explicit construction of an observable library, the network architecture, including its depth, width, latent dimension, or activation functions, and cost function weights, must still be specified. Also, the resulting models may be less interpretable. Nonlinear decoders can provide compact latent representations, especially for high-dimensional data, but they make output reconstruction nonlinear and are therefore less suitable for optimal linear control. Metaheuristic optimization can also be used for dictionary learning as demonstrated in~\cite{GrasevAccess2025}.

The second major class of approaches consists of spectral methods, which aim at a direct decomposition of the nonlinear dynamics in terms of Koopman eigenvalues and eigenfunctions~\cite{Mezic2020_Spectral}. These methods are of central interest in this paper. They are conceptually deeply connected with the proper orthogonal decomposition~\cite{Mezic2012}, where system trajectories are decomposed into orthogonal spatial modes, temporal modes, and modal amplitudes. Within the Koopman framework, these spatial and temporal structures are directly associated with the eigenfunctions and eigenvalues of the Koopman operator, yielding a Koopman mode decomposition~\cite{Surana2020}.

A key feature of spectral approaches is that the Koopman eigenfunctions are explicitly identified, and the corresponding Koopman matrix becomes a diagonal matrix of Koopman eigenvalues. This formulation can aid dictionary-free approaches, avoiding the difficulties associated with library selection in EDMD. Moreover, Koopman eigenfunctions evolve independently, spanning a Koopman-invariant subspace. Thus, their use mitigates the closure problems associated with arbitrary EDMD dictionaries~\cite{Kaiser2021, Brunton2016InvSubs}. Critically, these methods strive to identify low-rank representations with the most prominent eigenfunctions that explain the underlying dynamics.

Mezi\'{c} and co-authors proposed a generalized Laplace analysis for constructing eigenfunctions from nonlinear trajectories~\cite{Mezic2012, Mauroy2013}. This approach allows exact eigenfunctions to be recovered in certain settings and provides important geometric insight, such as the computation of isostables, which will also be discussed later. However, the method is primarily applicable to dissipative systems, and its numerical implementation may suffer from ill-conditioning. It also involves numerical backward integration, which can be unstable~\cite{Mauroy2013}.

Neural-network-based approaches have also been developed. Lusch et al. employed deep autoencoders to identify Koopman eigenfunctions, including an auxiliary network accounting for continuous spectra~\cite{Lusch2018}. Physics-informed neural networks for Koopman eigenfunction discovery were introduced in~\cite{Liu2022}. While these methods are powerful, their performance and complexity additionally depend on the selected network architecture, training procedure, and dataset size. Moreover, the interpretability of neural networks is limited.

Dictionary-based spectral approaches using sparse identification of nonlinear dynamics (SINDy) were proposed in~\cite{Kaiser2021} to identify and validate Koopman eigenfunctions through the Koopman partial differential equation (KPDE) and thus reduce the likelihood of spurious eigenfunctions. The same framework enabled successful feedback control design using state-dependent Riccati equations. The work of Grasev~\cite{GrasevSpringer2026} further developed this idea for gas turbine engine dynamics, combining SINDy-based control-affine model identification, eigenvalue optimization using projection onto exponential time bases, and eigenfunction identification by solving the KPDE using a dictionary of logistic functions and the Adam optimizer~\cite{Kingma2015}. Finally, the Koopman Kalman observer and linear quadratic Gaussian (LQG) tracking controller were designed.  

Another important spectral approach was proposed by Korda et al.~\cite{Korda2020}, where Koopman eigenfunctions are constructed from autonomous trajectory data using boundary functions defined on a non-recurrent set. The method establishes a mapping between the initial conditions of the nonlinear trajectories and the initial conditions in the eigenfunction space, and it strongly inspires the temporal component of the present work. It is primarily formulated for transient off-attractor dynamics. Recurrent dynamics, such as limit cycles, are therefore not directly accommodated by the same assumptions.

A recent dictionary-free method, combined with model predictive control (MPC), was proposed in~\cite{Cibulka2025}. This method treats the predictor matrices and values of lifting functions as optimization variables. The optimization is carried out using multiple segments of nonlinear trajectories and gradient descent. The method also learns nonlinear input transformations preserving the convex nature of the MPC. Because the lifting-function values, predictor matrices, and input transformation are optimized jointly, the dimension of the optimization problem can grow substantially with the number of data samples and lifting coordinates. It was demonstrated that symmetries of the systems can be exploited to speed up the optimization. The results also show a successful application to the control of systems with multiple fixed points and discontinuous eigenfunctions. 

The approaches discussed above generally require either a prescribed function library, function class, kernels, network architecture, boundary functions defined on a non-recurrent set, or large-scale optimization of lifting function initial conditions and predictor parameters. Moreover, temporal prediction accuracy alone does not necessarily ensure that the identified eigenfunctions are consistent with the KPDE. Motivated by this, the present paper proposes a spatially aware\footnote{The term "spatially aware" refers to the state-space dependence of the identified eigenfunctions and their gradients rather than physical spatial coordinates.} dictionary-free eigenfunction discovery (SADFED) framework, in which, for given candidate eigenvalues, the identification of Koopman modes and eigenfunction initial conditions is recast into LS problems. The eigenvalues are subsequently optimized using a population-based algorithm, minimizing cost functions that enforce accuracy of temporal prediction and KPDE consistency. The proposed framework primarily targets low-dimensional dynamical systems encountered in practical engineering applications, where compact Koopman representations can facilitate identification, prediction, and control design.

The main contributions are as follows:
\begin{itemize}
    \item A dictionary-free Koopman eigenfunction identification framework is proposed that does not require a predefined observable library, kernel, or neural network parameterization.
    
    \item For given eigenvalues, the Koopman modes and eigenfunction initial conditions are obtained via regularized LS problems. This reduces optimization variables to the real and imaginary parts of the eigenvalues.

    \item State prediction error is combined with a physics-informed KPDE residual that promotes spatial consistency with the KPDE. The resulting eigenfunctions and their numerically estimated gradients can also be used to construct state-dependent lifted input dynamics. 
    
    \item The framework is demonstrated on systems with analytical eigenfunctions, nontrivial fixed points, invariant subsets, nonlinear limit cycles, noisy and sparsely sampled data, and controlled nonlinear dynamics. The examples also illustrate the use of the identified model for estimation and optimal control design.
\end{itemize} 

The rest of the paper is organized as follows: Section~\ref{sec_koop} provides a concise introduction to the Koopman operator theory. Section~\ref{sec_SADFED} presents the core derivation of the SADFED approach and discusses its contributions, while Section~\ref{sec4} provides practical implementation guidelines and remarks. The application of the proposed methodology to selected benchmark nonlinear systems, along with the corresponding results and discussion, is provided in Section~\ref{sec_res}. Finally, Section~\ref{sec_conc} summarizes the key findings, highlights the strengths and limitations of the approach, and outlines potential directions for future research.

\section{Koopman operator fundamentals} \label{sec_koop}
The Koopman operator theory provides an alternative approach to the analysis of nonlinear dynamical systems. The core principle is based on a coordinate transformation of the system's state with nonlinear time evolution to a new, infinite-dimensional (functional) space of observables that are evolving linearly along the trajectories of the nonlinear system. 

Consider a discrete-time system obtained by sampling a continuous-time process, $\mathbf{\dot{x} = f(x)}$, with $\mathbf{f(x)}$ the drift vector field, with period $\tau$:
\begin{align} \label{eq_sys_1}
    \mathbf{x}_{k+1} = \mathbf{F}^\tau(\mathbf{x}_k)\,,
\end{align}
where $\mathbf{x} \in \mathbb{R}^m$ is the state vector and $\mathbf{F}^\tau:\mathbb{R}^m\rightarrow\mathbb{R}^m$ is a nonlinear discrete-time map. 

Let $\mathbf{\Psi(x)} \in \mathbb{C}^{n_\Psi}$ be a generic nonlinear vector observable function, with $n_\Psi$ being the number of observables. The Koopman operator, $\mathcal{K}^\tau$, is a linear operator that evolves the observable by composing it with the flow of the discrete-time system~\cite{Mezic2012}:
\begin{align}
    \mathbf{\Psi}(\mathbf{x}_{k+1}) = [\mathcal{K}^\tau\mathbf{\Psi}](\mathbf{x}_{k}) = \mathbf{\Psi(F}^\tau(\mathbf{x}_{k}))\,.
\end{align}

Given that $\mathcal{K}^\tau$ is linear, the nonlinear trajectory produced by $\mathbf{F^\tau(x)}$ is mapped to a new infinite-dimensional lifted state space, where it evolves in a linear manner.

For a continuous-time process, there is an entire family of these operators depending on the sampling period $\tau$. This family of operators forms a Koopman semigroup generated by the infinitesimal Koopman generator, $\mathcal{L}$. It is defined as~\cite{Mauroy2020, Klus2020}
\begin{align}
    [\mathcal{L}\mathbf{\Psi}](\mathbf{x}) = \lim_{\tau \rightarrow 0}\frac{ [\mathcal{K}^\tau \mathbf{\Psi}]\mathbf{(x)} - \mathbf{\Psi(x)}}{\tau} = \frac{\mathrm{d}\mathbf{\Psi}}{\mathrm{d}t}\,.
\end{align}

This paper will utilize this continuous-time generator formulation. Since the operator is linear, it is reasonable to consider its spectral properties. For this, special observables, $\mathbf{\Phi(x)} = [\varphi_1(\mathbf{x})\ \varphi_2(\mathbf{x})\ \dots\ \varphi_{n_\varphi}(\mathbf{x})]^\intercal\,, \mathbf{\Phi}:\mathbb{R}^m \to \mathbb{C}^{n_\varphi}$, called eigenfunctions, are defined as
\begin{align}
    \mathcal{L} \mathbf{\Phi} = \mathbf{\Lambda} \mathbf{\Phi} = \frac{\mathrm{d}\mathbf{\Phi}}{\mathrm{d}t}\,,
\end{align}
where $\mathbf{\Lambda} \in \mathbb{C}^{n_\varphi \times n_\varphi}$ is a diagonal matrix of eigenvalues of the Koopman generator, and $n_\varphi$ denotes the dimension of the lifted system. 

Applying the chain rule, the dynamics of $\mathbf{\Phi}$ can be expressed using the dynamics of the underlying continuous-time process $\mathbf{f(x)}$, as
\begin{align} \label{eqKPDE}
    \frac{\mathrm{d}\mathbf{\Phi}}{\mathrm{d}t} = \mathbf{\nabla_x \Phi(x)\,f(x) = \Lambda \Phi(x)}\,,
\end{align}
with $\mathbf{\nabla_x\Phi}$ the eigenfunction gradient.

The equation represents the KPDE; a fundamental equation capturing a deep underlying connection between the evolution of $\mathbf{\Phi}$ along trajectories, $\mathbf{f(x)}$, and its spatial structure. 

The action of $\mathcal{L}$ on eigenfunctions thus yields a solution to the following linear system:
\begin{align} \label{eqPhiDotSys}
    \mathbf{\dot{\Phi}}(t) = \mathbf{\Lambda\Phi}(t), \quad \mathbf{\Phi}(0) = \mathbf{\Phi_0}\,.
\end{align}

Since the system~(\ref{eqPhiDotSys}) is diagonal, the space of eigenfunctions is invariant to the action of the Koopman generator represented by $\mathbf{\Lambda}$. Finally, any observable of interest, $\mathbf{\Psi}$, can be projected onto the span of eigenfunctions and expressed using the Koopman mode decomposition as~\cite{Surana2020}
\begin{align} \label{KMD}
    \mathbf{\Psi}(\mathbf{x}(t)) \approx \mathbf{C_\Psi} \mathbf{\Phi}(\mathbf{x}(t)) = \sum_{l=1}^{n_\varphi} \mathbf{c}_{\mathbf{\Psi},l}\,\varphi_l(\mathbf{x_0})\,e^{\lambda_l t},
\end{align}
where $\mathbf{C_\Psi}=[\mathbf{c}_{\mathbf{\Psi},1}\ \cdots\ \mathbf{c}_{\mathbf{\Psi},n_\varphi}]$ is the Koopman mode matrix and its $l$-th column $\mathbf{c}_{\mathbf{\Psi},l}\in\mathbb{C}^{n_\Psi}$ is the Koopman mode associated with $\varphi_l$. Each row of $\mathbf{C_\Psi}$ contains the reconstruction coefficients for one component of $\mathbf{\Psi}$.

\begin{figure}[!ht]
    \centering
    \includegraphics[width=0.8\linewidth]{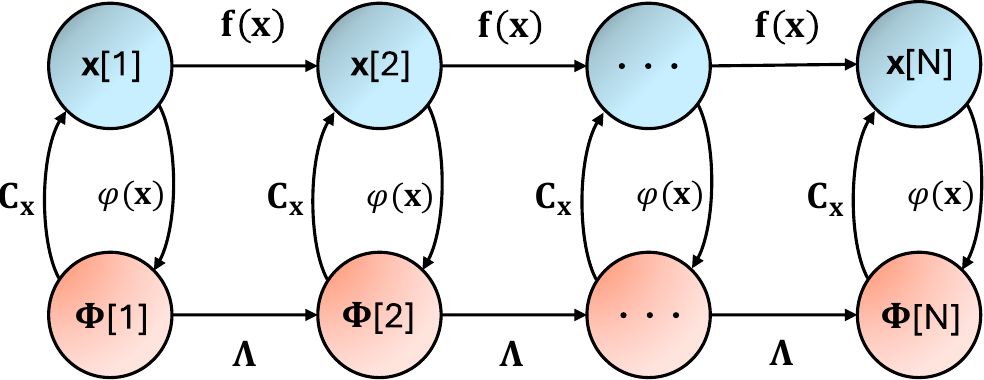}
    \caption{Schematic diagram of Koopman eigenfunction lifting and evolution.}
    \label{Diag_Koopman}
\end{figure}

In practice, the primary demand is that the set of eigenfunctions of the finite-dimensional approximation of $\mathcal{L}$ is a suitable basis for reconstruction of observables of interest, e.g., $\mathbf{x \approx C_x} \mathbf{\Phi}$ with $\mathbf{C_x}$ the modes corresponding to the states. Fig.~\ref{Diag_Koopman} shows the interplay between the evolution of $\mathbf{\Phi}$ given by~(\ref{eqPhiDotSys}) and the original states evolved by $\mathbf{f(x)}$ and reconstructed using their output matrix, $\mathbf{C_x}$.

\section{The proposed approach: Spatially Aware Dictionary-Free Eigenfunction Discovery (SADFED)} \label{sec_SADFED}

Suppose that observables of interest can be accurately represented by a function of spectral components $(\mathbf{\Phi_0, \Lambda, C})$ according to~(\ref{KMD}). Consider the autonomous trajectory data
\begin{align}    
    \mathbf{X} = \begin{bmatrix}
    \mathbf{x}^{(1)}(0) & \cdots & \mathbf{x}^{(1)}(t_N) \\
    \vdots & \ddots & \vdots \\
    \mathbf{x}^{(n_\mathrm{t})}(0) & \cdots & \mathbf{x}^{(n_\mathrm{t})}(t_N)
    \end{bmatrix}
    \in \mathbb{R}^{m n_\mathrm{t} \times N},
\end{align}
where $m$ is the state dimension, $N$ is the number of samples, and $n_\mathrm{t}$ denotes the number of trajectories initialized from a grid of initial conditions $\mathbf{x_0}^{(i)}$.

The objective is to identify a finite-dimensional Koopman spectral representation and the corresponding sampled eigenfunction map $\mathbf{\Phi_0} = \mathbf{\Phi}(\mathbf{x_0})$, such that the observables lie in the span of the eigenfunctions and are accurately reconstructed using~(\ref{KMD}). Here, the main observables are the state variables of the nonlinear system. Additionally, it is desirable to avoid treating $\mathbf{\Lambda}$, $\mathbf{C}$, and $\mathbf{\Phi_0}$ for all trajectories as independent optimization variables.

Furthermore, all eigenfunctions in $\mathbf{\Phi(x)}$ should satisfy the KPDE, which should be explicitly addressed during the optimization, as good accuracy of the mode decomposition~(\ref{KMD}) alone does not guarantee valid eigenfunctions. Thus, inspired by~\cite{Kaiser2021, Liu2022}, the KPDE residual should be introduced into the cost function. It should promote consistency with the KPDE over the sampled region, ideally without the need to parameterize the eigenfunctions.

The present method leverages the linear dependence of the Koopman spectral decomposition on the eigenfunction initial conditions and Koopman modes. A transformation of the temporal basis is proposed, enabling both quantities to be obtained through two regularized LS projections. Consequently, only the eigenvalues remain as decision variables of the outer optimization loop.

\paragraph{Roadmap of the approach}
\begin{enumerate}
    \item Collect autonomous trajectories from a grid of initial conditions and select a suitable reference trajectory.
    \item For a candidate set of eigenvalues, construct an exponential temporal basis.
    \item Determine the reference Koopman modes by LS projection of the reference trajectory.
    \item Transform the temporal basis and project the remaining trajectories onto it to obtain $\mathbf{\Phi_0}$.
    \item Compute the eigenfunction gradient using interpolation and numerical differentiation of eigenfunction landscapes.
    \item Evaluate temporal state reconstruction and KPDE cost functions. 
    \item Close the loop with an eigenvalue optimization algorithm minimizing the joint cost function. 
    \item After convergence, refine the eigenfunctions on a denser grid of initial conditions.
\end{enumerate}
All steps will be explained in the following sections.

\subsection{Reference trajectory and reference modes}
The first step of the proposed approach is the selection of a reference trajectory, $\mathbf{x_{ref}} \in \mathbb{R}^{m \times N}$. The reference trajectory should cover a substantial portion of the investigated state-space region and sufficiently excite the most prominent modes of the low-rank Koopman decomposition~(\ref{KMD}). Ideally, it should capture both the fast transients and the slow asymptotic dynamics. 

Consider a given reference trajectory $\mathbf{x_{ref}}$ starting from a reference initial condition $\mathbf{x_{ref,0}}$. The selection criteria for this trajectory will be discussed later. Also consider a single pair of complex conjugate eigenvalues, $\lambda_{1,2} = \alpha \pm \beta i$, with $\alpha = \Re(\lambda)$ and $\beta = \Im(\lambda)$, and the corresponding eigenfunctions $\varphi_{1,2}(\mathbf{x}) = \varphi_\mathrm{R}(\mathbf{x}) \pm i \varphi_\mathrm{I}(\mathbf{x})$, with $\varphi_\mathrm{R} = \Re(\varphi)$ and $\varphi_\mathrm{I} = \Im(\varphi)$. Further, assume a real-valued representation, where $\varphi_\mathrm{R}$ and $\varphi_\mathrm{I}$ are represented as separate real-valued lifted states~\cite{GrasevSpringer2026}. For a single complex eigenpair, the fundamental temporal basis of~(\ref{eqPhiDotSys}) is
\begin{align} \label{eq_E_basis}
    \mathbf{E} = \begin{bmatrix}
        e^{\alpha t}(\varphi\mathrm{_{R0}}\cos(\beta t) - \varphi\mathrm{_{I0}}\sin(\beta t) ) \\[6pt]  e^{\alpha t}(\varphi\mathrm{_{R0}}\sin(\beta t) + \varphi\mathrm{_{I0}}\cos(\beta t) )
    \end{bmatrix} \in \mathbb{R}^{2 \times N }\,,
\end{align}
where $\varphi_\mathrm{R0} = \varphi_\mathrm{R}(\mathbf{x_0})$ and $\varphi_\mathrm{I0} = \varphi_\mathrm{I}(\mathbf{x_0})$.

For $\mathbf{x_{ref}}$, the initial values are normalized and fixed as $\varphi_\mathrm{R}(\mathbf{x_{ref,0}}) = \varphi_\mathrm{I}(\mathbf{x_{ref,0}}) = 1$, yielding a reference basis $\mathbf{E_{ref}}$. The corresponding reference Koopman mode matrix is obtained as
\begin{align} \label{eq_Cref}
    \mathbf{C_{ref} = \left((E_{ref} E_{ref}^\intercal + \rho I)^\mathrm{-1} E_{ref} x_{ref}^\intercal\right)}^\intercal \in \mathbb{R}^{m \times 2}\,,
\end{align}
where $\mathbf{I}$ is the identity matrix and $\rho > 0$ is an L2 regularization parameter.

\subsection{Change of temporal basis}
The main idea is to fix the matrix of Koopman modes, $\mathbf{C_{ref}}$, and find the eigenfunction initial conditions by projecting each trajectory onto a new temporal basis given by the known exponential trajectories and modes.

Suppose that the observables are $\mathbf{\Psi(x)}  =\mathbf{x}\in\mathbb R^m$. Considering the single complex eigenpair $(\varphi_\mathrm{R}, \varphi_\mathrm{I}, \alpha, \beta)$ from above, the decomposition~(\ref{KMD}) for an $i$-th trajectory from $\mathbf{X}$ yields
\begin{align} \label{eq_ref_KMD}
    \mathbf{x}^{(i)}  \approx &\,\mathbf{C_{ref} E}^{(i)} \nonumber\\[6pt]
    = &\,e^{\alpha t} \big(\mathbf{c_1^{ref}}\varphi_\mathrm{R0}^{(i)}  \cos(\beta t)  - \mathbf{c_1^{ref}}\varphi_\mathrm{I0}^{(i)} \sin(\beta t)\nonumber\\ & \ \ +\mathbf{c_2^{ref}}\varphi_\mathrm{R0}^{(i)} \sin(\beta t) + \mathbf{c_2^{ref}}\varphi_\mathrm{I0}^{(i)} \cos(\beta t) \big)\,,
\end{align} 
with $\begin{bmatrix} \mathbf{c_1^{ref}} & \mathbf{c_2^{ref}} \end{bmatrix} = \mathbf{C_{ref}}$ and unknown $\varphi_\mathrm{R0}^{(i)},\varphi_\mathrm{I0}^{(i)}$.

The decomposition can be vectorized, e.g., using the \texttt{reshape} function in MATLAB, so each spectral element of the decomposition is of size either $1 \times mN$ or $mN \times 1$. Rearranging the terms in~(\ref{eq_ref_KMD}), the decomposition can be expressed as
\begin{align}
    \mathbf{x_r}^{(i)} \approx \left(\mathbf{\Phi_0}^{(i)}\right)^\intercal \mathbf{B}\,,
\end{align} 
where $\mathbf{x_{r}}^{(i)} = \mathrm{vec}(\mathbf{x}^{(i)})^\intercal \in \mathbb{R}^{1 \times mN}$,
\begin{align}
    \mathbf{\Phi_0}^{(i)} = \begin{bmatrix}
        \varphi_\mathrm{R0}^{(i)} & \varphi_\mathrm{I0}^{(i)}
    \end{bmatrix}^\intercal \in \mathbb{R}^{2 \times 1}\,,
\end{align}
and
\begin{align} \label{eq_B}
    \mathbf{B} = \begin{bmatrix}
        \mathrm{vec}\!\left(\ \ \mathbf{c_1^{ref}} e^{\alpha t} \cos(\beta t) + \mathbf{c_2^{ref}} e^{\alpha t} \sin(\beta t) \right)^\intercal \\[6pt]
        \mathrm{vec}\!\left(-\mathbf{c_1^{ref}} e^{\alpha t} \sin(\beta t) + \mathbf{c_2^{ref}} e^{\alpha t} \cos(\beta t) \right)^\intercal
    \end{bmatrix} \,.
\end{align}

The matrix $\mathbf{B} \in \mathbb{R}^{2 \times mN}$ represents a new vectorized temporal basis. The presented construction applies to an arbitrary number of eigenfunctions, $n_\varphi$, and to any other observable $\psi(\mathbf{x})$ of interest.

\subsection{Optimal eigenfunction initial conditions}
Finally, for each vectorized trajectory $\mathbf{x_{r}}^{(i)}, i=1,2,...,n_\mathrm{t}$, the optimal eigenfunction initial values, $\mathbf{\Phi_0}^{(i)}=\mathbf{\Phi}(\mathbf{x_0}^{(i)})$, are obtained by projection onto the new basis $\mathbf{B}$ as
\begin{align} \label{eq_Phi0}
    \mathbf{\Phi_0}^{(i)} = \left( \mathbf{B B}^\intercal + \rho\mathbf{I} \right)^{-1}\mathbf{B} (\mathbf{x_r}^{(i)})^\intercal \,.
\end{align}

The obtained initial conditions $\mathbf{\Phi_0}^{(i)}$ are collected into a single matrix $\mathbf{\Phi_0} \in \mathbb{R}^{n_\varphi \times n_\mathrm{t}}$. Subsequently, the temporal cost function can be computed as
\begin{align} \label{eq_Jtemp}
    \mathcal{J}_\mathrm{temp} = \frac{1}{n_\mathrm{t}} \sum_{i = 1}^{n_\mathrm{t}} \left\| \mathbf{x}^{(i)} - \mathbf{C_{ref} E}^{(i)} \right\|_\mathrm{F}^2\,,
\end{align}
with $\mathbf{\Lambda}$ now the block-diagonal matrix
\begin{align}\label{eq_lambda_mat}
    \mathbf{\Lambda} = \begin{bmatrix}
        \alpha_1 & -\beta_1 & & & \\ \beta_1 & \alpha_1 & & & \\ & & \ddots & & \\  & & & \alpha_{n_\lambda} & -\beta_{n_\lambda} \\ & & & \beta_{n_\lambda} & \alpha_{n_\lambda}
    \end{bmatrix}_{n_\varphi \times n_\varphi}, 
\end{align}
where $n_\lambda$ is the number of eigenvalues and, for real eigenvalues, $\beta = 0$.

\subsection{Spatial integrity and KPDE cost function}
As mentioned before, the eigenfunctions must also satisfy the KPDE~(\ref{eqKPDE}), ensuring spatio-temporal integrity on the sampled state-space region. The collected initial condition data $\{\mathbf{\Phi_0}^{(i)},\mathbf{x_0}^{(i)}\}_{i=1}^{n_\mathrm{t}}$ represent samples of the candidate eigenfunctions and encode their shape.

To verify adherence of the candidate eigenfunctions to the KPDE, the gradient, $\mathbf{\nabla_x \Phi}$, is required. For this, the eigenfunction samples are interpolated on a finer grid $\{\mathbf{x_{int}}^{(k)}\}_{k=1}^{n_{\mathrm{int}}}$ to obtain the relation $\mathbf{\Phi(x)}$, and numerical differentiation is employed to estimate the gradient. In this work, smoothing-spline interpolation is utilized to suppress noise and spatial irregularities. Each column of the gradient is subsequently computed using 4th-order-accurate central differences. The derivative of $\mathbf{\Phi}$ at the $k$-th point of the interpolation grid w.r.t. the $j$-th state variable is given as
\begin{equation} \label{eq_graddiff}
    {\frac{\partial \mathbf{\Phi}}{\partial x_j}}^{(k)} \approx \frac{-\mathbf{\Phi}_\mathbf{int}^{k+2} + 8\mathbf{\Phi}_\mathbf{int}^{k+1} - 8\mathbf{\Phi}_\mathbf{int}^{k-1} + \mathbf{\Phi}_\mathbf{int}^{k-2}}{12 \Delta x_j}\,,
\end{equation}
where $\mathbf{\Phi}_\mathbf{int}^{k \pm a} = \mathbf{\Phi}(\mathbf{x}^{(k)}_\mathbf{int} \pm a\Delta x_j\mathbf{e}_j)$ with $\mathbf{e}_j$ the $j$-th unit vector.

Note that fitting the samples by an analytical function and computing the gradient analytically would introduce additional parameters and significantly increase the computational cost of closed-loop eigenvalue optimization.

Assuming the autonomous drift vector field $\mathbf{f(x)}$ is known, the normalized KPDE cost term is
\begin{align} \label{eq_JKPDE}
    \mathcal{J}_\mathrm{KPDE} = \frac{1}{n_\mathrm{int}}\sum_{k = 1}^{n_\mathrm{int}} \left\| \mathbf{N}^{-1} \mathbf{r}^{(k)}_\mathbf{KPDE} \right\|_2^2\,,
\end{align} 
where $n_\mathrm{int}$ denotes the total number of points on the interpolation grid, $\mathbf{N}=\mathrm{diag}(s_1,\ldots,s_{n_\varphi})$, with $s_l = \max_{k=1,\ldots,n_{\mathrm{int}}}|\varphi_l^{(k)}|$, normalizes the KPDE residual of each eigenfunction by its maximum absolute value over the interpolation grid, and $\mathbf{r}^{(k)}_\mathbf{KPDE} \in \mathbb{R}^{n_\varphi \times 1}$ is the KPDE residual obtained as 
\begin{align} \label{eq_rKPDE}
    \mathbf{r}^{(k)}_\mathbf{KPDE} = (\mathbf{\nabla_x \Phi})^{(k)} \mathbf{f}^{(k)} - \mathbf{\Lambda \Phi}_\mathbf{int}^{(k)}\,,
\end{align}
with $\mathbf{f}^{(k)} = \mathbf{f}(\mathbf{x_{int}}^{(k)})$. 

\subsubsection{Rationale for KPDE normalization}
The $\mathbf{r_{KPDE}}$ normalization using $\mathbf{N}$ reduces the dependence of $\mathcal{J}_{\mathrm{KPDE}}$ on the selected reference trajectory $\mathbf{x_{ref}}$. Since the eigenfunctions are normalized to 1 at $\mathbf{x_{ref,0}}$, their overall scale depends on their absolute value at this point, $|\varphi(\mathbf{x_{ref,0}})|$. If $|\varphi_l(\mathbf{x_{ref,0}})| = |\varphi_{l}|_\mathrm{max}$ over the sampled state-space region, then $|\varphi_l| \leq 1$ in this region. Conversely, given the same shape of the eigenfunction, selecting $\mathbf{x_{ref,0}}$ where $|\varphi_l(\mathbf{x_{ref,0}})|$ is small scales the entire eigenfunction, and thus its gradient, to larger values. Since $\mathbf{r_{KPDE}}$ scales linearly with the eigenfunction, its magnitude depends on the selection of $\mathbf{x_{ref}}$ and could bias the optimization. Normalization by maximum absolute values of the eigenfunctions removes this scaling dependence and makes $\mathcal{J}_{\mathrm{KPDE}}$ reflect the KPDE consistency independently of eigenfunction amplitudes.

\subsection{Eigenvalue optimization}
The concept of reference trajectory and change of the temporal basis allows elimination of $\mathbf{C}$ and $\mathbf{\Phi_0}$ from the optimization variables. These quantities are now computed analytically inside the objective function using LS, and only the real and imaginary parts of the eigenvalues, $\alpha$ and $\beta$, remain as independent decision variables, $\boldsymbol{\theta} = [\alpha_1,\beta_1,\dots,\alpha_{n_\lambda},\beta_{n_\lambda}]^\intercal$.

\subsubsection{Cost function}
The final cost function is
\begin{align} \label{final_cost}
    \mathcal{J}(\boldsymbol{\theta}) = \mathcal{J}_\mathrm{temp}(\boldsymbol{\theta}) + \gamma \mathcal{J}_\mathrm{KPDE}(\boldsymbol{\theta})\,,
\end{align}
where $\gamma \ge 0$ is a KPDE cost weighting parameter.

The temporal cost function $\mathcal{J}_\mathrm{temp}$ captures how well all the trajectories of the nonlinear system are represented by the candidate spectral components $(\mathbf{\Phi_0, \Lambda, C})$ according to the Koopman mode decomposition~(\ref{KMD}). It also promotes identification of relevant decay rates and fundamental frequencies, corresponding to long-term behavior of the system.

The KPDE cost $\mathcal{J}_\mathrm{KPDE}$ acts as a physics-informed regularizer\footnote{Here, \textit{physics-informed} refers to the mathematical use of a governing PDE cost function, inspired by physics-informed ML. However, the KPDE may encode physical structure in specific systems, e.g., Hamiltonian dynamics~\cite{Kaiser2021}.}: among the many solutions that provide a good basis for the Koopman mode decomposition and have comparable $\mathcal{J}_\mathrm{temp}$, it promotes eigenfunctions consistent with the KPDE, enhancing the dynamical relevance of the resulting low-rank Koopman model and the uniqueness and correctness of the discovered eigenfunctions. 

The evaluation of the objective function is summarized in Algorithm~\ref{algo1}.
\begin{algorithm}[!ht]
\caption{SADFED objective function evaluation for candidate eigenvalues}
\label{algo1}
\begin{algorithmic}[1]

\State \textbf{Input:}
$\boldsymbol{\theta}$, $\mathbf{X}$, $\{\mathbf{x_0}^{(i)}\}_{i=1}^{n_\mathrm{t}}$, $\mathbf{f(x)}$, $\mathbf{x_{ref}}$,
$\rho$, $\gamma$

\State \textbf{Output:}
$\mathcal{J} = \mathcal{J}_{\mathrm{temp}} + \gamma\mathcal{J}_{\mathrm{KPDE}}$

\State Extract $\{\alpha_l,\beta_l\}_{l=1}^{n_\lambda}$ from $\boldsymbol{\theta}$
and construct the reference temporal basis $\mathbf{E_{ref}}$

\State Compute reference Koopman modes $\mathbf{C_{ref}}$~(\ref{eq_Cref})

\State Construct the transformed temporal basis
$\mathbf{B}(\mathbf{C_{ref}},\boldsymbol{\theta})$~(\ref{eq_B})

\For{$i=1,\ldots,n_\mathrm{t}$}
    \State Vectorize $\mathbf{x}^{(i)}$ and compute $\mathbf{\Phi_0}^{(i)}$~(\ref{eq_Phi0})
    
    \State Reconstruct $\mathbf{x}^{(i)}$ and store the reconstruction residuals
\EndFor

\State Compute $\mathcal{J}_{\mathrm{temp}}$~(\ref{eq_Jtemp})

\State Interpolate
$\{\mathbf{x_0}^{(i)},\mathbf{\Phi_0}^{(i)}\}_{i=1}^{n_\mathrm{t}}$
onto the interpolation grid and estimate $\mathbf{\nabla_x \Phi}$~(\ref{eq_graddiff})

\State Construct
$\mathbf{N}=\mathrm{diag}(s_1,\ldots,s_{n_\varphi})$,
$s_l = \max_{k=1,\ldots,n_{\mathrm{int}}}|\varphi_l^{(k)}|$

\For{$k=1,\ldots,n_{\mathrm{int}}$}
    \State Evaluate $\mathbf{r}^{(k)}_\mathbf{KPDE}$~(\ref{eq_rKPDE})
\EndFor

\State Compute
$\mathcal{J}_{\mathrm{KPDE}}$~(\ref{eq_JKPDE})

\State \Return $\mathcal{J} = \mathcal{J}_{\mathrm{temp}} + \gamma\mathcal{J}_{\mathrm{KPDE}}$

\end{algorithmic}
\end{algorithm}

The optimization problem is summarized as
\begin{equation} \label{eq_problem}
    \boldsymbol{\theta}^* = \arg \min_{\boldsymbol{\theta}} \mathcal{J}(\boldsymbol{\theta}) \,,
\end{equation}
s.t.
\begin{equation}
\begin{aligned}\label{eq_constraints}
     \alpha_\mathrm{min} &\le  \alpha_l  \le \alpha_\mathrm{max}\,, &&l = 1,...,n_\lambda\,, \\
     \beta_\mathrm{min} &\le  \beta_l  \le \beta_\mathrm{max}\,,  &&l = 1,...,n_\lambda \,, \\
     d_{ij} &\ge d_\mathrm{min}\,, &&1 \le i < j \le n_\lambda\,,
\end{aligned}
\end{equation}
with $d_{ij}$ defined as 
\begin{align}
    d_{ij} = \sqrt{(\alpha_i-\alpha_j)^2 + (\beta_i - \beta_j)^2}\,.
\end{align}

The $\alpha$ and $\beta$ bounds restrict the candidate decay rates and frequencies to a dynamically relevant search region. The last condition in~(\ref{eq_constraints}) avoids nearly repeated eigenvalue pairs, promoting exploration and diversity of the solution. Once the optimal solution $\boldsymbol{\theta}^*$ is obtained, the eigenvalue matrix $\mathbf{\Lambda}$ is constructed using~(\ref{eq_lambda_mat}).

\subsubsection{Optimizer}
The objective function is highly nonlinear and nonconvex, and computation of its analytical gradient w.r.t. eigenvalues can be cumbersome. This limits utilization of gradient-based optimization methods, which might also get stuck at local minima. In addition, the low dimensionality of $\boldsymbol{\theta}$, and thus of the search space, and simplicity of the cost function motivate the utilization of population-based global optimization methods. In this work, the particle swarm optimization (PSO)~\cite{Kennedy1995} is used to explore the bounded eigenvalue space and locate the most promising candidate solution. PSO was selected based on preliminary numerical comparisons with genetic algorithms and differential evolution. Subsequently, local search is performed using the gradient-free Nelder--Mead (NM) method~\cite{Nelder1965} to refine the solution. 

\begin{figure*}[!ht]
    \centering
    \includegraphics[width=1\linewidth]{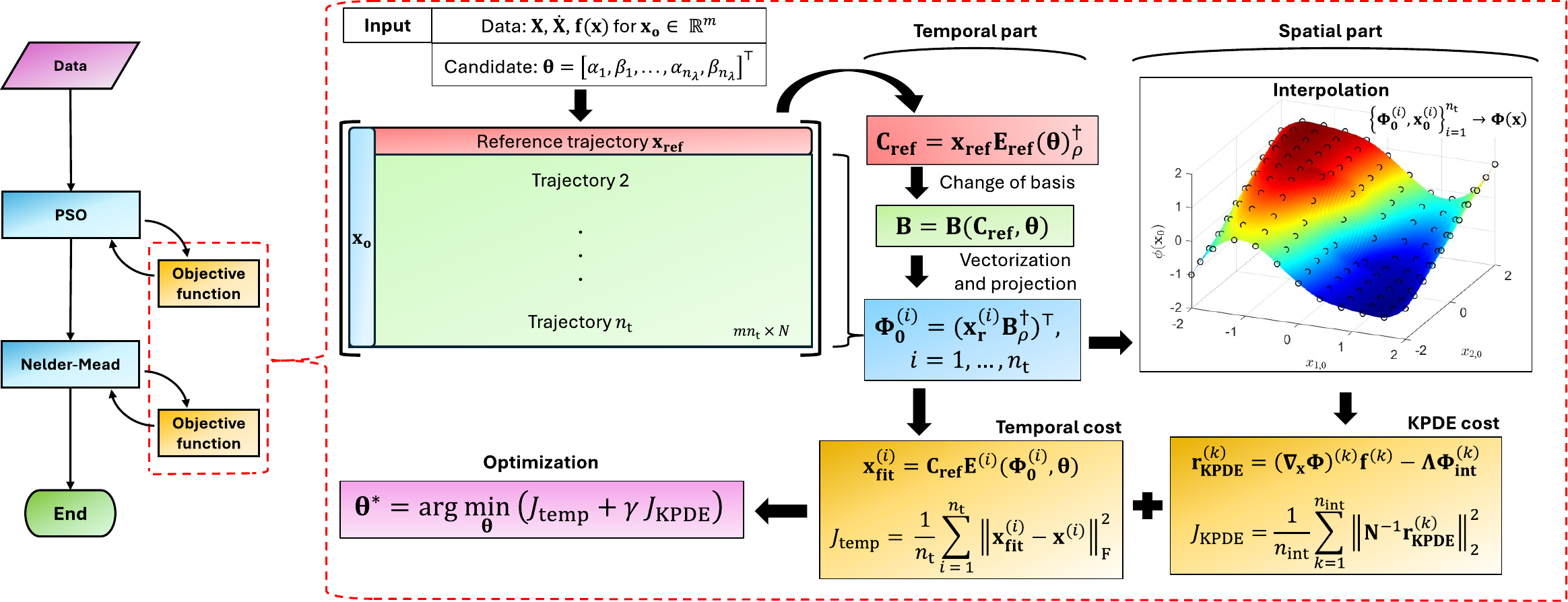}
    \caption{Flowchart of the SADFED eigenvalue optimization and the objective function.}
    \label{ID_Cost_Flows}
\end{figure*}

The entire optimization procedure is depicted in Fig.~\ref{ID_Cost_Flows}. 

\subsection{Eigenfunction post-processing} \label{ssec_refine}
After the optimal eigenvalues are identified, the eigenfunction landscapes can be refined using a denser grid of initial conditions and an even finer interpolation grid. During the eigenvalue optimization, relatively coarse sampling is sufficient to reduce the computational burden, since the eigenfunction reconstruction and KPDE evaluation are repeated many times throughout the optimization process. Once the optimal solution $\boldsymbol{\theta}^*$ is obtained, the trajectories from the denser grid are computed only once and projected onto the already determined temporal basis $\mathbf{B}$, yielding higher-resolution eigenfunction samples. These samples are subsequently interpolated on a finer grid to improve the accuracy of the eigenfunctions and their numerically estimated gradients.

The refined eigenfunction samples do not require an analytical representation and can be stored in a lookup table and evaluated via linear interpolation. For autonomous open-loop prediction, the interpolation is employed only at initialization to obtain $\mathbf{\Phi}(0)$, after which the lifted dynamics evolve linearly according to~(\ref{eqPhiDotSys}). This avoids fitting an additional parametric model, at the expense of interpolation accuracy, memory requirements, and scalability with the state dimension. If an analytical representation is required, the eigenfunction samples may be used to train a separate parametric model. However, this step is external to the core SADFED identification procedure.

\subsection{Summary and relation to existing approaches}
The proposed SADFED framework combines ideas from trajectory-based Koopman eigenfunction construction and physics-informed identification based on the KPDE. The use of autonomous trajectories and the recovery of eigenfunction samples from their temporal evolution are inspired by the trajectory-based construction from~\cite{Korda2020}, while the explicit KPDE cost is motivated by approaches such as~\cite{Kaiser2021} or~\cite{Liu2022}. In contrast to EDMD and sparse regression methods based on prescribed dictionaries~\cite{Williams2015, Kaiser2021}, kernel methods, and neural-network models, SADFED does not assume a predefined lifting dictionary, kernel, or parametric eigenfunction architecture. Compared with~\cite{Korda2020, Cibulka2025}, the identification is reduced to optimization of only the eigenvalues, decreasing the size of the problem, while the spatio-temporal consistency of eigenfunctions is directly promoted through the numerical KPDE cost function. 

The main contribution of SADFED is twofold. First, the notion of reference trajectory, the use of fixed reference Koopman modes, and the change of temporal basis enable the elimination of $\mathbf{\Phi_0}$ and $\mathbf{C}$ from the nonlinear optimization through two-stage regularized LS, leaving only the eigenvalues as decision variables. Second, $\mathcal{J}_\mathrm{KPDE}$ introduces a physics-informed regularization into the same identification framework, reducing the ambiguity among solutions with comparable $\mathcal{J}_\mathrm{temp}$ by promoting eigenfunctions that are consistent with the KPDE over the sampled region, thereby enhancing their correctness and uniqueness.

\section{Practical considerations and remarks} \label{sec4}
The previous section described the core SADFED identification procedure. Its practical implementation can be adapted according to known information about the dynamical structure of the investigated system. This section summarizes practical guidelines, remarks, applicability, limitations, and extension to controlled systems.

\subsection{Selection of reference trajectory} \label{sec_ref_traj_sel}
The reference trajectory should cover a substantial portion of the sampled state-space region and sufficiently excite the prominent modes of the low-rank Koopman system. Ideally, it should capture both fast transients and slow asymptotic dynamics.

In dissipative systems, trajectories converging last are generally suitable candidates because they retain the influence of slow Koopman modes associated with the principal eigenvalues and asymptotic dynamics. These modes govern the long-term behavior and may be weakly represented in rapidly converging trajectories. The selected trajectory should also sufficiently cover transient dynamics.

For systems with a stable limit cycle, the trajectory should include the transient toward the cycle, span at least one period to correctly capture the fundamental frequency, and sufficiently cover its phase. A slowly converging trajectory is again suitable because it captures both the periodic dynamics and the decaying transient modes.

The sensitivity to the reference trajectory follows from its role in determining the reference Koopman modes and the transformed basis $\mathbf{B}$. Weak excitation of relevant modes may lead to poorly estimated mode amplitudes, an ill-conditioned basis $\mathbf{B}$, and distorted eigenfunction landscapes. These distortions may be further amplified by numerical differentiation and increase $\mathcal{J}_{\mathrm{KPDE}}$.

Therefore, the selection is guided by state-space coverage and convergence times. In practice, several promising candidates should be evaluated, and the final choice should exhibit the lowest total cost and a well-conditioned solution.

\subsection{Principal, zero, and repeated eigenvalues} \label{sec_princ_zero_eig}
For many systems, some Koopman eigenvalues can be determined a priori and included directly in the spectrum. For systems with fixed points, principal Koopman eigenvalues can be obtained from the linearization Jacobian at the corresponding fixed points. In particular, those with real parts closest to zero describe the slow asymptotic dynamics within the corresponding basin of attraction~\cite{Mauroy2013}.

For systems with stable limit cycles and point spectrum, the fundamental frequencies of the limit cycles can be obtained as $\omega_0 = 2\pi/T$, with $T$ the cycle period, yielding fundamental imaginary eigenvalues with $\beta = \omega_0$. Integer multiples of $\omega_0$ correspond to higher harmonics. Decaying transient modes associated with Floquet exponents may also be included to represent convergence to the cycle.

A zero eigenvalue with $\alpha_0 = 0$ and $\beta_0 = 0$ can be included to represent constant or invariant eigenfunctions. Alternatively, for a single nontrivial fixed point, the state can first be shifted to the origin, and, after optimization, the spectral components can be updated as
\begin{equation}
    \begin{aligned}
        \mathbf{\Phi_0}  &\leftarrow \begin{bmatrix} \mathbf{1}_{1 \times n_\mathrm{t}} \\ \mathbf{\Phi_0} \end{bmatrix}\,, \\ 
        \mathbf{\Lambda} &\leftarrow \mathrm{blkdiag}\left(0,\mathbf{\Lambda}\right)\,, \\ 
        \mathbf{C_{ref}} &\leftarrow \begin{bmatrix} \mathbf{x_{FP}} & \mathbf{C_{ref}} \end{bmatrix}\,,
    \end{aligned}
\end{equation}
where $\mathbf{x_{FP}}$ denotes the coordinates of the fixed point. 

For systems with multiple invariant subsets, a zero-eigenvalue eigenfunction may act as an indicator of the corresponding basin of attraction~\cite{Pan2024}, as will be demonstrated later.

If repeated eigenvalues with nontrivial Jordan blocks are expected, the formulation can be extended by including the secular terms $t^q e^{\lambda t}$, where $q = 0, \dots, s-1$ for a Jordan block of size $s$, and the corresponding Jordan blocks in $\mathbf{\Lambda}$.

\subsection{Selection of model order} 
\label{sec_model_order}
The model order depends primarily on the acceptable prediction error, and can be selected as follows:

\begin{enumerate}
    \item Begin with principal eigenvalues if available and zero-eigenvalue components if needed.
    \item Incrementally increase $n_\lambda$.
    \item Evaluate prediction errors, preferably using unseen trajectories.
    \item Select the lowest $n_\lambda$ for which the error is acceptable and increasing it yields only marginal improvement.
\end{enumerate}

Furthermore, Koopman modes $\mathbf{C}$ can be inspected, and spectral components with negligible mode amplitudes can be excluded. Also, excessive $n_\lambda$ may result in some optimal eigenvalues located at the decision-variable bounds or too close to each other, leading to redundant spectral components, providing additional indicators of over-parameterization. The effect of the model order is discussed in Section \ref{sec_sens_anal}.

\subsection{Regularization and KPDE weight}
The parameters $\rho$ and $\gamma$ can be tuned empirically. A small $\rho$ is initially selected and increased if the LS projections become ill-conditioned or produce distorted eigenfunction landscapes. Owing to the normalization of $\mathcal{J}_{\mathrm{KPDE}}$, $\gamma$ primarily controls the trade-off between temporal reconstruction and spatial KPDE consistency and should be adjusted to match the magnitude of $\gamma \mathcal{J}_\mathrm{KPDE}$ to that of $\mathcal{J}_\mathrm{temp}$. Its influence, together with that of $\rho$, is investigated in the sensitivity analysis presented later.

\subsection{Basins of attraction and discontinuous eigenfunctions} \label{sec_basins}
For systems with multiple basins of attraction, numerical gradient computation fails near separatrices, compromising $\mathcal{J}_\mathrm{KPDE}$. Therefore, these regions should be excluded from the KPDE evaluation. Separatrix shape and location can be estimated by inspecting the behavior of the collected trajectories, e.g., by assigning different colors to initial points based on their corresponding fixed point. Subsequently, the gradient and $\mathcal{J}_\mathrm{KPDE}$ can be computed only at points sufficiently far from the separatrices. 

The zero-eigenvalue eigenfunction is thus a piecewise-constant function parameterizing the basins of attraction~\cite{Williams2015, Pan2024} with a step discontinuity at the separatrix.

\subsection{Exploiting symmetry}
For systems with known state-space symmetry, e.g., the van der Pol and Duffing systems, the state space may be sampled only at its symmetry-related subspace and the remaining dynamics reconstructed using the corresponding symmetry transformation. This can substantially reduce the number of trajectories required for identification.

\subsection{Computational scalability} \label{sec_scal}
SADFED is primarily proposed for identification of low-dimensional nonlinear dynamical systems, particularly practical engineering systems where compact models are desirable for prediction and control. The main scalability limitation arises from the state-space sampling and numerical evaluation of the KPDE residual. For $d$ state variables and $q$ nodes per coordinate, the grid of initial conditions yields $q^d$ trajectories, and the finer interpolation grid also grows significantly larger. Consequently, direct application to high-dimensional systems, such as spatially discretized PDEs, becomes computationally demanding. For PDEs in particular, methods such as EDMD or Koopman autoencoders may offer more practical alternatives, as they can offer better scalability, although their complexity depends on the selected observables or network architecture \cite{Kutz2018-EDMD-PDEs, Otto2019-AE-PDEs}. Alternatively, proper orthogonal decomposition could be employed to extract dominant spatio-temporal modes, followed by SADFED identification of the reduced-order dynamics. 

The computational cost of SADFED could also be reduced in several ways. The KPDE term can be disabled setting $\gamma = 0$, retaining the proposed temporal basis transformation and two-stage regularized LS formulation, while sacrificing explicit spatial KPDE consistency. Where applicable, symmetry can also be exploited. Alternatively, the KPDE residual could be evaluated only at carefully selected checkpoints within the operating region, using eigenfunction values interpolated at neighboring perturbed points. Combined with scattered interpolation, this could avoid constructing a dense interpolation grid and substantially reduce the KPDE computational burden. Further validation of SADFED on higher-dimensional systems is beyond the scope of the present paper.

\subsection{Applicability and limitations} 
SADFED is expected to perform best for systems with a point spectrum when a useful low-rank approximation exists, the relevant modes are sufficiently excited, the sampled trajectories sufficiently cover the investigated state-space region, and the vector field $\mathbf{f(x)}$ is known or can be estimated. It is applicable to systems with single or multiple fixed points, nonlinear limit cycles, and control. 

SADFED is a numerical and nonconvex identification procedure. Global convergence of the optimizer is not guaranteed and depends on tuning its hyperparameters, such as the population size or velocity weights in PSO. For given candidate eigenvalues, the two-stage regularized LS provides unique solutions. However, uniqueness of the resulting eigenvalues is not guaranteed, despite the KPDE regularization reducing ambiguity among solutions with comparable $\mathcal{J}_\mathrm{temp}$.

The main limitations include the scalability to higher-order systems as discussed in Section \ref{sec_scal}. Further challenges arise for continuous spectra, strongly chaotic systems, noisy derivative estimates, and extrapolation outside the sampled region. Fixing the eigenvalues, the eigenfunction maps can be extended to new regions if additional trajectories are available in those regions. A small $\mathcal{J}_\mathrm{KPDE}$ indicates KPDE consistency only over the sampled region, not globally.

\subsection{Extension to control}
In the case of \textit{controlled systems}, the gradient plays an important role, as it multiplies the input dynamics in the original state space. Consider the control-affine system
\begin{align}
    \mathbf{\dot{x} = f(x) + G(x) u}\,,
\end{align}
where $\mathbf{G(x)} \in \mathbb{R}^{m \times p}$ is the known input matrix and $\mathbf{u} \in \mathbb{R}^p$ is a vector of $p$ inputs. 

Insertion of this system into the KPDE~(\ref{eqKPDE}) yields
\begin{align}
    {\frac{\mathrm{d} \mathbf{\Phi}}{\mathrm{d} t} = \mathbf{\nabla_x\Phi\cdot\dot{x}} = \mathbf{\Lambda\Phi}+\underbrace{\mathbf{\nabla_x\Phi(x)G(x)}}_{\mathbf{\Gamma(x)}} \mathbf{u}}\,.
\end{align}

Control-nonaffine systems $\mathbf{\dot{x} = f(x,u)}$ can be reformulated in a control-affine form via input-state augmentation, $\mathbf{x_{aug}} = [\mathbf{x}^\intercal\,\,\mathbf{u}^\intercal]^\intercal$, with a new input $\mathbf{v = \dot{u}}$. SADFED can then be applied to the augmented system, yielding input-dependent eigenfunctions $\varphi(\mathbf{x, u})$. Alternatively, a factorization described in \cite{Iacob2024} can be employed.

The gradient is already available from the eigenfunction refinement, and the lifted input matrix $\mathbf{\Gamma(x)}$ can be evaluated over the same grid. For online implementation, it may be interpolated or represented by a compact analytical surrogate, yielding a state-dependent lifted predictor suitable for methods such as the state-dependent Riccati equation~\cite{Kaiser2021}, gain-scheduled LQG~\cite{GrasevSpringer2026}, adaptive LQG~\cite{Annaswamy2021}, or other suitable control approaches.

\subsection{SADFED within the complete modeling and control framework}
\begin{figure}[!ht]
    \centering
    \includegraphics[width=0.95\linewidth]{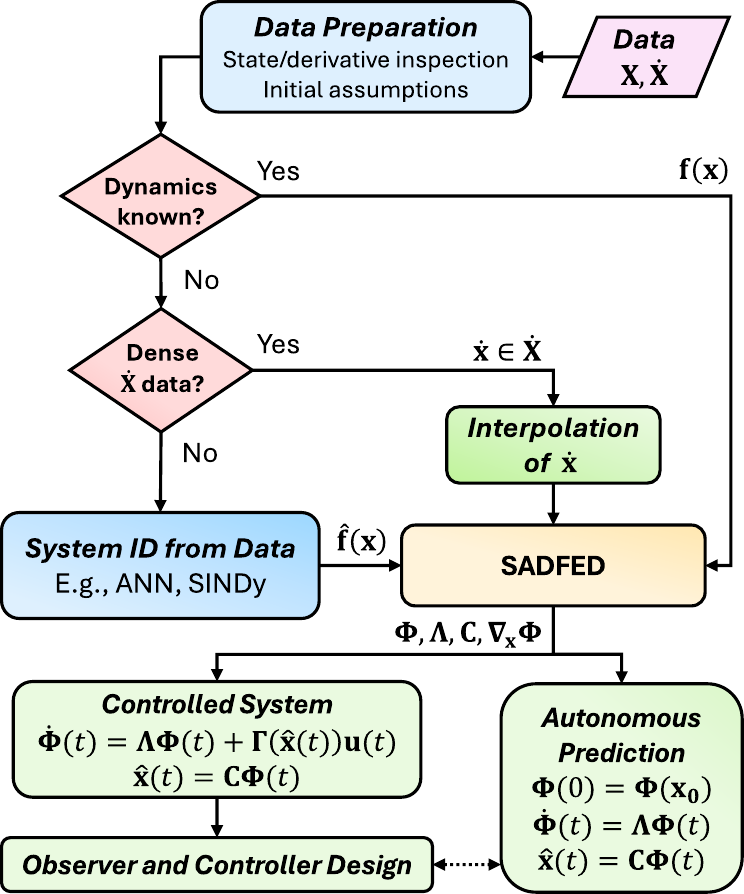}
    \caption{SADFED in the context of a broader data-driven modeling workflow.}
    \label{flowchart_ALL}
\end{figure}

SADFED represents the Koopman identification component of the broader modeling workflow summarized in Fig.~\ref{flowchart_ALL}. Its core inputs are autonomous trajectory data $\mathbf{X}$ and the drift vector field $\mathbf{f(x)}$, required for the evaluation of the KPDE. If $\mathbf{f(x)}$ is known, it can be evaluated directly at the interpolation points. Otherwise, sufficiently dense and accurate time-derivative data may be interpolated to approximate $\mathbf{f(x)}$, while sparse or noisy data generally require separate identification, e.g., using SINDy or neural networks. The resulting estimate is then used only for the KPDE evaluation, whereas the temporal part of SADFED remains based solely on $\mathbf{X}$.

The output of SADFED comprises $\mathbf{\Lambda}$, $\mathbf{C}$, $\mathbf{\Phi}$, and $\mathbf{\nabla_x \Phi}$. For autonomous prediction, the eigenfunction map is required only for initialization and may therefore be stored numerically as described in Section~\ref{ssec_refine}. Also, an observer, e.g., a Kalman filter, can be designed for autonomous prediction using the observer Riccati equation~\cite{Surana2020}. As discussed in the preceding subsection, for control-affine systems, the gradient can additionally be combined with the input vector field to construct $\mathbf{\Gamma(x) = \nabla_x\Phi(x) G(x)}$.  The resulting lifted predictor can then be employed independently of SADFED for observer and controller design.

\section{Results} \label{sec_res}
The proposed approach is demonstrated on several benchmark systems. The systems and motivation for their selection are provided in Table~\ref{tab_systems_motivation}.
\begin{table}[!ht]
\centering
\caption{Example systems and their role in the validation of SADFED.}
\label{tab_systems_motivation}
\begin{tabular}{p{0.2\columnwidth} p{0.25\columnwidth} p{0.4\columnwidth}}
\toprule
\textbf{System} & \textbf{Motivation} &
\textbf{SADFED validation} \\
\midrule
Quadratic manifold & Analytical Koopman representation. & Recovery of analytical eigenfunctions -- analytical verification. \\
\midrule
FitzHugh--Nagumo with inputs & Nontrivial fixed point and nonlinear input dynamics. & Sensitivity analyses, lifted input dynamics, and comparison with EDMDc using a single controlled trajectory. \\
\midrule
Van der Pol & Nonlinear limit cycle and asymptotic phase dynamics. & Spectrum discovery, harmonics, refinement beyond the optimization region, and isochrons. \\
\midrule
Duffing & Multiple fixed points, basins of attraction, and discontinuous eigenfunction. & Identification of invariant subsets, indicator eigenfunction and isostables, and symmetry exploitation. \\
\midrule
Two-spool turbojet & Practical nonlinear system with multiple states and inputs. & Application to engineering systems, lifted input modeling, state estimation, and nonlinear control. \\
\bottomrule
\end{tabular}
\end{table}

\subsection{System with closure and exact eigenfunctions}
The first investigated system admits an exact finite-dimensional Koopman-invariant subspace and therefore provides a ground-truth benchmark. The system is given as
\begin{equation}
\begin{aligned}
    \dot{x}_1 &= \mu x_1  \\
    \dot{x}_2 &= \nu(x_2 - x_1^2)\,,
\end{aligned}
\end{equation}
where $\mu$ and $\nu$ are the principal eigenvalues at the system's single trivial fixed point. The observables $\psi_1 = x_1$, $\psi_2 = x_2$, and $\psi_3 = x_1^2$ form a closed Koopman-invariant subspace because $\dot{\psi}_3 = 2x_1\dot{x}_1 = 2 \mu x_1^2 = 2 \mu\psi_3$. For $\mu = -0.1$ and $\nu = -1$, the linear Koopman system is
\begin{align}
    \frac{\mathrm{d}}{\mathrm{d}t}\begin{bmatrix}
        \psi_1 \\ \psi_2 \\ \psi_3
    \end{bmatrix} = 
    \begin{bmatrix}
        -0.1 & 0 & 0 \\ 0 & -1 & 1 \\ 0 & 0 & -0.2
    \end{bmatrix}
    \begin{bmatrix}
        \psi_1 \\ \psi_2 \\ \psi_3
    \end{bmatrix},
\end{align}
and the eigenfunctions are $\varphi_1 = x_1$, $\varphi_2 = 1.6x_1^2$, and $\varphi_3 = x_2 - 1.25x_1^2$ up to arbitrary scaling. 

Trajectories were simulated from a grid of $21 \times 21$ initial conditions in the range of $[-1, 1]^2$ using Heun numerical integration with a time step of $\Delta t = 0.2$~s for 50 seconds, yielding 251 time samples for each.

\begin{figure}[ht]
    \centering
    \includegraphics[width=0.3\linewidth]{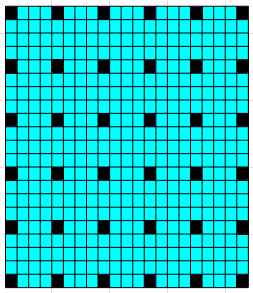}
    \caption{Sparse selection of trajectories from the full $21 \times 21$ (blue) grid to a $6 \times 6$ (black) subgrid using every second odd row and column.}
    \label{fig_grid}
\end{figure}

A $6\times6$ subgrid of trajectories was used for the temporal part, as shown in Fig.~\ref{fig_grid}. The reference trajectory started at $[-1,-1]$ and captured both the slow manifold and the fast transient.  

The principal eigenvalues were prescribed, while two additional complex-conjugate eigenvalue pairs were optimized using PSO followed by NM, with a PSO population of 50, 200 generations, and 2000 NM iterations. The representation thus contains one additional mode compared to the exact solution. The initial population was generated using the hypercube sampling function. The parameters were set to $\rho = 10^{-6}$ and $\gamma = 10^{-4}$, and the KPDE was evaluated on a $100 \times 100$ interpolation grid ($\Delta\mathbf{x} = [0.02, 0.02]$). 

\begin{figure}[ht]
    \centering
    \includegraphics[width=1\linewidth]{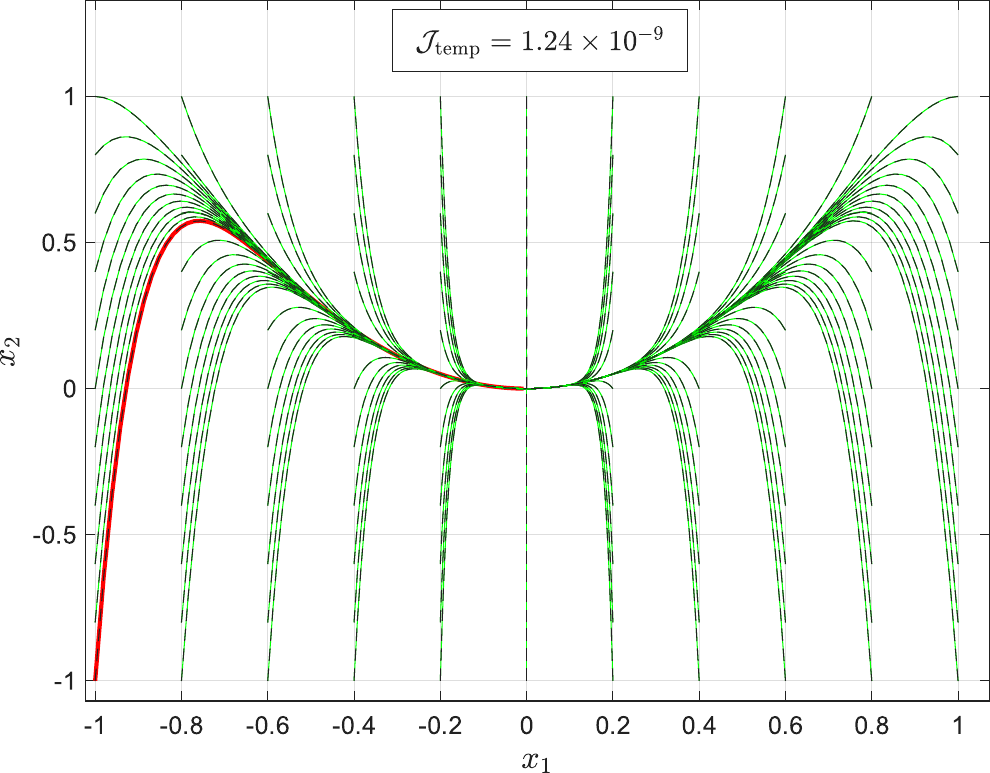}
    \caption{Ground truth (green solid) vs. reconstructed (black dashed) trajectories. The red line highlights the reference trajectory.}
    \label{ex_eig_fit_traj}
\end{figure}

The resulting costs were $\mathcal{J}_\mathrm{temp} = 1.24 \times 10^{-9}$ and $\mathcal{J}_\mathrm{KPDE} = 0.048$, with the reconstructed trajectories shown in Fig.~\ref{ex_eig_fit_traj}. 

\begin{figure}[!ht] 
    \centering 
    \begin{subfigure}{0.45\textwidth}
        \centering
        \includegraphics[width=0.9\linewidth]{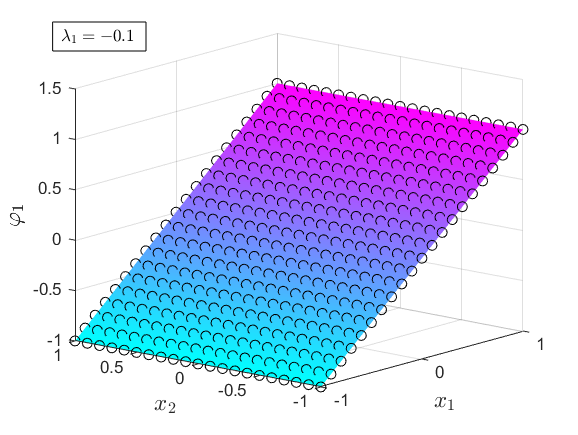}
        \caption{}
        \label{sys1_phi1}
    \end{subfigure}
    \begin{subfigure}{0.45\textwidth}
        \centering
        \includegraphics[width=0.9\linewidth]{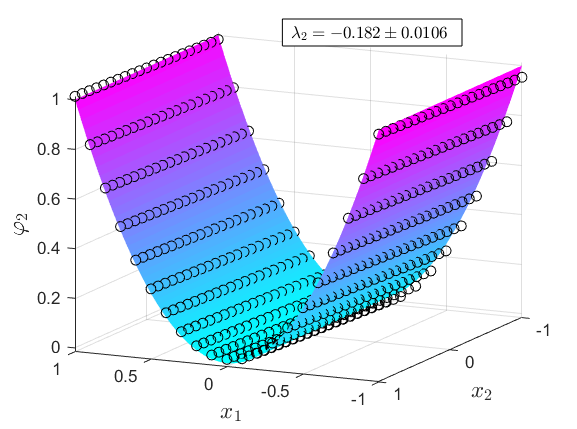}
        \caption{}
        \label{sys1_phi2}
    \end{subfigure} 
    \begin{subfigure}{0.45\textwidth}
        \centering
        \includegraphics[width=0.9\linewidth]{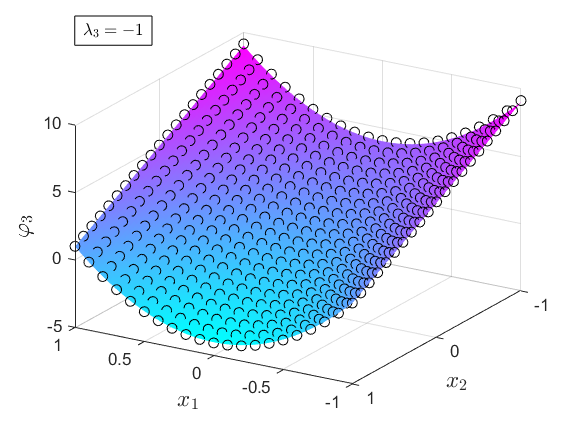}
        \caption{}
        \label{sys1_phi3}
    \end{subfigure}
    \caption{Exact (colored surfaces) vs. estimated (black circles) eigenfunctions: (a) $\varphi_1$, (b) $\varphi_2$, and (c) $\varphi_3$. Eigenfunctions were normalized such that $\varphi_i(1, 1) = 1$.}
    \label{exact_eig_fcns}
\end{figure}

The identified eigenvalues were $\lambda_1 = -0.1$, $\lambda_2 = -0.182 \pm 0.0106i$, $\lambda_3 = -1$, and $\lambda_4 = -0.34$. The mode associated with the additional eigenvalue $\lambda_4$ was negligible and was therefore omitted. The eigenvalue $\lambda_2$ approximates the exact eigenvalue $2\mu = -0.2$, with only a small residual imaginary part. Figure~\ref{exact_eig_fcns} compares the identified $\varphi_1, \varphi_2$, and $\varphi_3$ with the exact solutions, showing close agreement. For visualization only, both exact and identified eigenfunctions are rescaled to unity at $(1,1)$. 

\subsection{Shifted fixed point: FitzHugh-Nagumo system with inputs}
The control-affine FitzHugh-Nagumo model with inputs was selected to demonstrate the use of the numerically estimated eigenfunction gradient for constructing lifted input dynamics. The system is given as
\begin{equation}
\begin{aligned}
    \underbrace{\begin{bmatrix}
         \dot{x}_1 \\[3pt] \dot{x}_2
    \end{bmatrix}}_{\mathbf{\dot{x}}} = 
    &\underbrace{\begin{bmatrix}
        -x_2 - x_1(x_1 - 1)(x_1 - a) + b \\[3pt] c(x_1 - d x_2)
    \end{bmatrix}}_{\mathbf{f(x)}} \\ 
    &+ \underbrace{\begin{bmatrix}
        1 \\[3pt] \cos({x_1^2})
    \end{bmatrix}}_{\mathbf{g(x)}} u \,,
\end{aligned}
\end{equation}
where $a = 0.5$, $b = 0.05$, $c = 0.2$, and $d = 1$. 

\begin{figure}[!ht]
    \centering
    \includegraphics[width=1\linewidth]{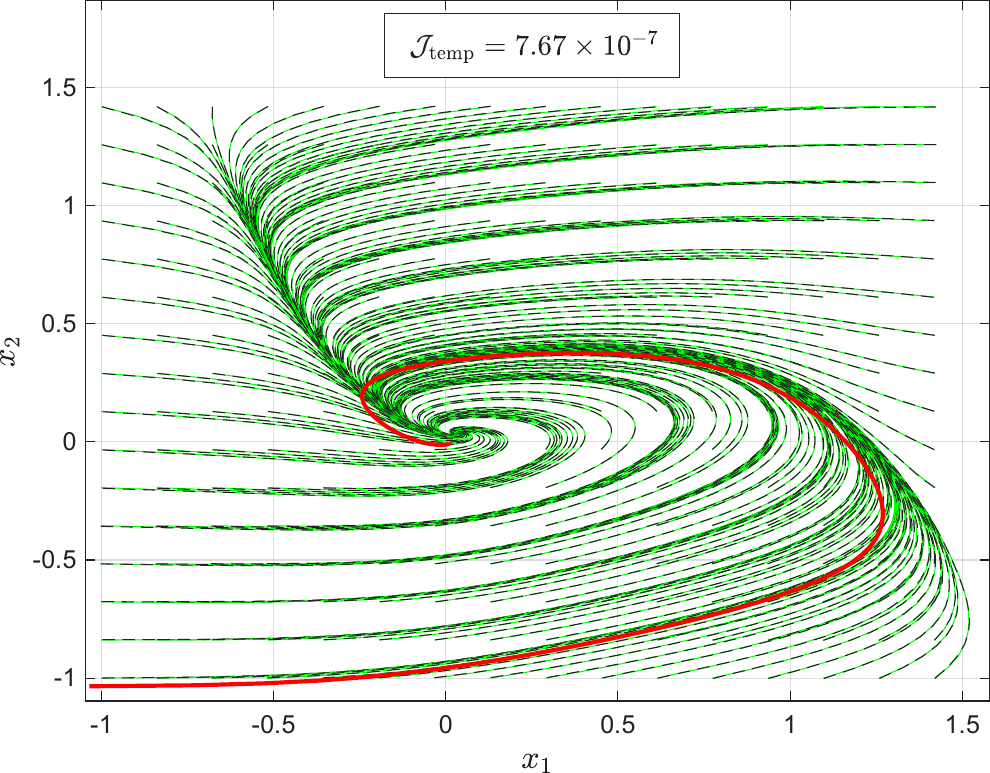}
    \caption{Reconstructed autonomous trajectories of the FitzHugh-Nagumo system: ground truth (green solid), fit (black dashed), reference trajectory (red solid).}
    \label{fig_fitzhugh_auton_traj}
\end{figure}

The autonomous system ($u = 0$) was simulated from a $21 \times 21$ grid of initial conditions over $[-1, 1.5]^2$ for 20~s using Heun's method with $\Delta t = 0.01$~s and a sampling period $T_\mathrm{s} = 0.1$~s. The reference trajectory started at $[-1, -1]$ and captured both the slow manifold and the fast oscillatory transient, as shown in Fig.~\ref{fig_fitzhugh_auton_traj}. The fixed point is $\mathbf{x_{FP}} = (0.0345, 0.0345)$, for which the Jacobian matrix is
\begin{align}
    \mathbf{J} = \begin{bmatrix}
        -0.4 & -1 \\ 0.2 & -0.2
    \end{bmatrix}\,,
\end{align}
yielding the principal eigenvalues $\lambda_{1,2} = -0.3 \pm 0.436i$. 

Since the fixed point is nontrivial, $\mathbf{x_{FP}}$ was subtracted from the trajectories for the temporal identification and added back to the initial-condition coordinates for spatial interpolation and KPDE evaluation. The principal eigenvalues were fixed, while seven additional complex-conjugate eigenvalue pairs were optimized. After optimization, the constant eigenfunction, zero eigenvalue, and modes $\mathbf{x_{FP}}$ were appended to $\mathbf{\Phi_0}$, $\mathbf{\Lambda}$, and $\mathbf{C_{ref}}$, respectively, following Section~\ref{sec_princ_zero_eig}.

The identification parameters were $\rho = 10^{-6}$ and $\gamma = 10^{-4}$, with a PSO population of 100 and 2000 NM iterations. A $100 \times 100$ interpolation grid was used during optimization ($\Delta\mathbf{x} = [0.025, 0.025]$). The resulting costs were $\mathcal{J} = 1.488 \times 10^{-6}$, $\mathcal{J}_\mathrm{temp} = 7.67 \times 10^{-7}$, and $\mathcal{J}_\mathrm{KPDE} = 0.0072$. Once the optimal $\boldsymbol{\theta}^*$ was obtained, the eigenfunctions were refined using $125 \times 125$ trajectories and a $500 \times 500$ interpolation grid ($\Delta\mathbf{x} = 0.005$). The reconstructed trajectories are depicted in Fig.~\ref{fig_fitzhugh_auton_traj}.

Subsequently, the lifted input dynamics were approximated by the compact surrogate
\begin{align} \label{eq_input_fit}
\mathbf{\Gamma(x)} = \mathbf{\nabla_x \Phi(x) g(x)} \approx \mathbf{W_2} \sigma(\mathbf{W_1 x + b_1}) +\mathbf{b_2}\,,
\end{align}
where $\mathbf{W_1}$, $\mathbf{b_1}$, $\mathbf{W_2}$ and $\mathbf{b_2}$ are the input weights, input biases, output weights and output biases, respectively, and $\sigma(z) = 1/(1 + e^{-z})$ is the sigmoid function.

\begin{figure}[!ht]
    \centering
    \includegraphics[width=1\linewidth]{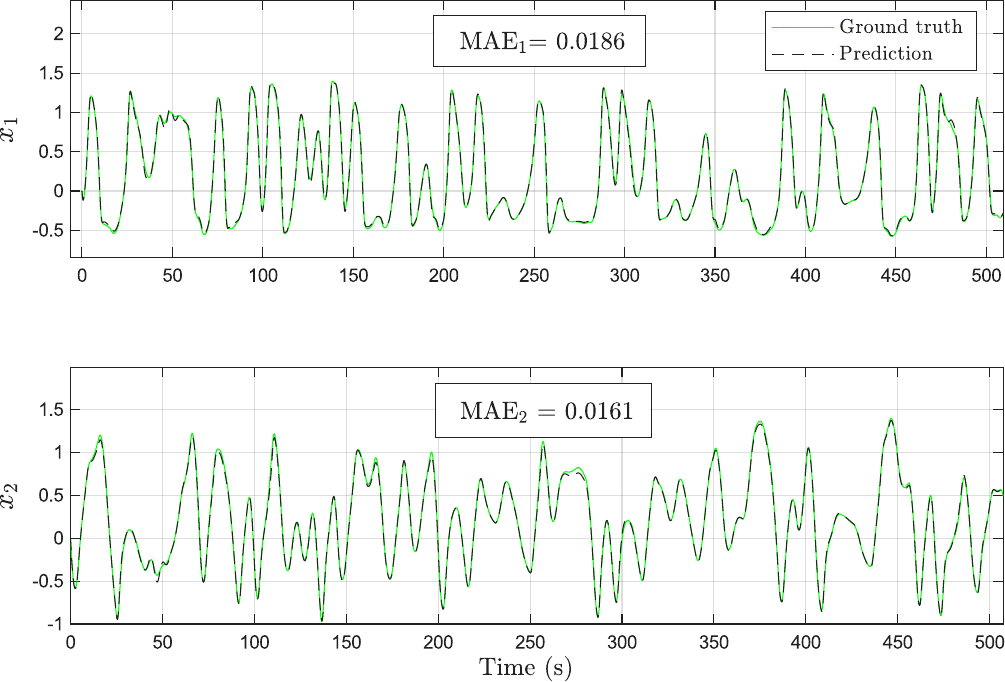}
    \caption{The fitted input-driven dynamics for the first 500 seconds of the validation dataset.}
    \label{fig_fitzhugh_inputdyn_ELM}
\end{figure}

The surrogate parameters were optimized using the Adam optimizer, and Fig.~\ref{fig_fitzhugh_inputdyn_ELM} shows the validation prediction results with $\mathrm{MAE} = \mathrm{MAE}_1 + \mathrm{MAE}_2 = 0.0347$.

\subsubsection{Comparison to EDMDc}
For comparison, the same input-driven training data were used to identify a bilinear EDMDc model:
\begin{equation}
\begin{aligned}
    \mathbf{\dot{\Psi}} &= \mathbf{A \Psi + B[u\,\,\Psi u]^\intercal}\,, \\
    \mathbf{\hat{x}} &= \mathbf{C_{\Psi} \Psi}\,.
\end{aligned}
\end{equation}

Four lifting functions were considered: (i) polynomial functions, (ii) Gaussian radial basis functions (GAU RBFs), (iii) inverse quadratic (IQ) RBFs, and (iv) scaled thin-plate spline (TPS) RBFs. The RBFs are defined as 
\begin{align} \label{eqRBF}
    \mathrm{GAU}(\mathbf{x}) &= \exp{\big(-\epsilon r^2 \big)}\,, \\[3pt]
    \mathrm{IQ}(\mathbf{x}) &= \frac{1}{1 + \epsilon r^2}\,, \\[3pt]
    \mathrm{TPS}(\mathbf{x}) &= r^2 \log |\epsilon r|\,,  
\end{align}
where $r = \|\mathbf{x - x_c}\|_2$, $\mathbf{x_c}$ is the collocation point, and $\epsilon$ is the scale.

In the first comparison, the lifted dimension was fixed to 16, equal to that of the SADFED model. L2 regularization was used to improve numerical conditioning. The RBF collocation points were uniformly distributed over $[-1, 1.5]^2$. For the RBFs, both the regularization and scaling parameters were optimized, whereas only the regularization parameter was optimized for the polynomial case. 

\begin{figure}[!ht]
    \centering
    \includegraphics[width=1\linewidth]{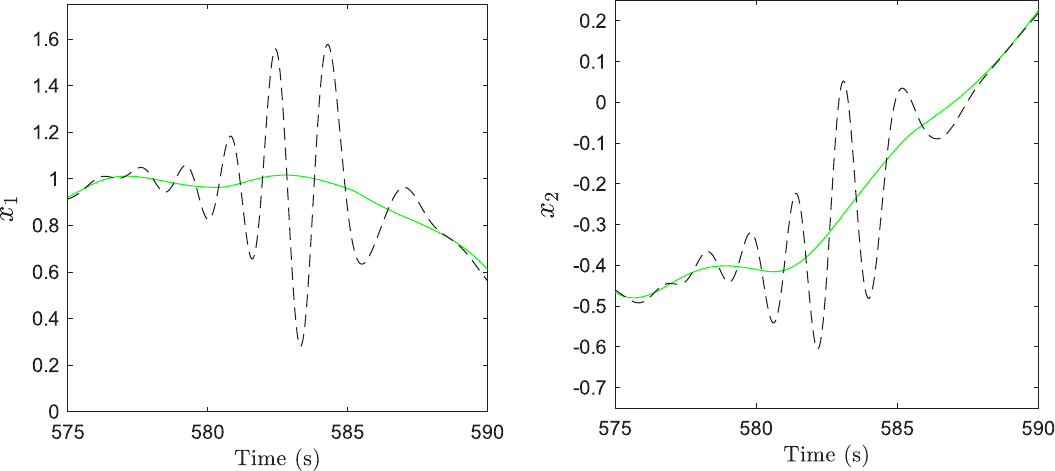}
    \caption{Example of oscillatory behavior for the EDMDc solution with 64 IQ RBFs. Ground truth (green solid) vs. EDMDc prediction (black dashed).}
    \label{FHN_Oscil}
\end{figure}

A second comparison investigated the EDMDc dimension required to outperform the SADFED. The regularization parameter was varied over $[10^{-6}, 10^{-1}]$. The scaling parameter was varied over $[0.5, 10]$ for the Gaussian and TPS RBFs and over $[0.1, 5]$ for the IQ RBFs. The RBF grid was increased from $2 \times 2$ to $8 \times 8$, corresponding to 4--64 RBFs. The best EDMDc result was obtained using 64 IQ RBFs with $\epsilon = 0.75$ and the regularization parameter of $10^{-5}$, yielding an MAE of 0.0270. Thus, EDMDc required a $4\times$ larger lifted dimension to achieve a lower MAE. Moreover, the resulting prediction exhibited an occasional oscillatory behavior, as shown in Fig.~\ref{FHN_Oscil}.

The best results are reported in Table~\ref{tab_edmd}. The dominant eigenvalues of the autonomous lifted dynamics are also reported for comparison. At the same lifted dimension of 16, SADFED achieves the lowest prediction error among all considered models while keeping the a priori principal eigenvalues. In this setting, the EDMDc achieves a lower MAE only after increasing the lifted dimension to 64.
\begin{table*}[!ht] 
\centering
\caption{Comparison of SADFED and EDMDc.}\label{tab_edmd}
\begin{tabular*}{\textwidth}{@{\extracolsep\fill}lccccc}
\toprule
Case & Dimension & L2 reg. par. & Scale, $\epsilon$ & MAE & Dominant $\lambda$ \\
\midrule
SADFED & 16 & $10^{-6}$ & --    & 0.0347 & $-0.3\pm0.436i$ \\
Poly (3rd)     & 16 & $10^{-2}$ & --    & 0.0949 & $-0.19\pm0.38i$ \\
IQ (16)        & 16 & $10^{-1}$ & 0.75  & 0.1063 & $-0.2\pm0.39i$ \\
IQ (64)        & 64 & $10^{-5}$ & 0.75  & 0.0270 & $-0.23\pm0.41i$ \\
Gaussian       & 16 & $10^{-1}$ & 1     & 0.1270 & $-0.2\pm0.39i$ \\
TPS            & 16 & $10^{-1}$ & 10    & 0.1621 & $-0.2\pm0.38i$ \\
\bottomrule
\end{tabular*}
\end{table*}

\subsubsection{Sensitivity analyses} \label{sec_sens_anal}
Sensitivity to the reference trajectory, hyperparameters, model and grid dimensions, and measurement noise was investigated for the FitzHugh-Nagumo example. 

\begin{itemize}
\item \textbf{Reference trajectory selection.} 
To quantify the effect of the reference trajectory selection, trajectories initialized at 17 reference points were selected: the center point, four corners, four intermediate boundary points, and eight intermediate interior points. The optimization was performed using identical settings for all trajectories. For each trajectory, the convergence time $t_\mathrm{conv}$ was the time after which the trajectory remained inside a $0.01 \times 0.01$ neighborhood of the fixed point. As shown in Fig.~\ref{reftraj_t_conv}, the total identification cost decreases approximately exponentially with increasing convergence time. This supports the practical guidelines from Section \ref{sec_ref_traj_sel} that slowly converging reference trajectories are generally preferable. Figure~\ref{reftraj_ss} shows the candidate trajectories, with the color indicating the $\mathcal{J}$. The selected reference trajectory yields the smallest $\mathcal{J}$.
\begin{figure}[!ht] 
    \centering 
    \begin{subfigure}{0.45\textwidth}
        \centering
        \includegraphics[width=0.8\linewidth]{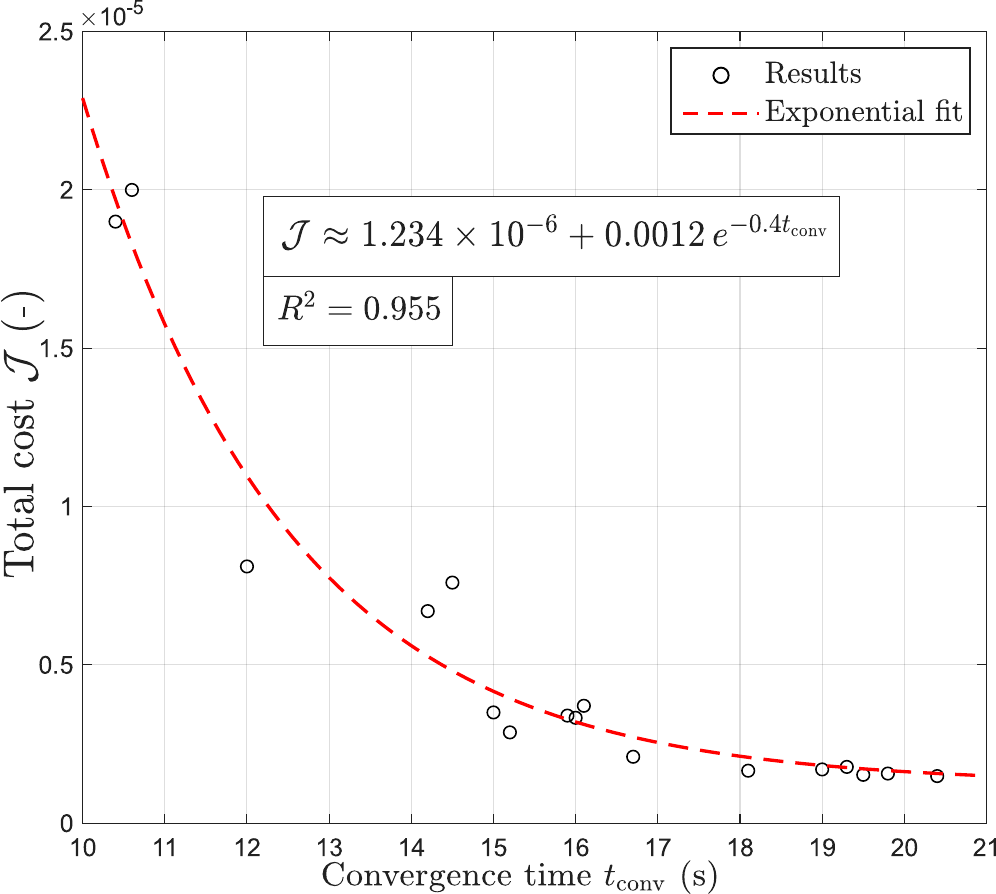}
        \caption{}
        \label{reftraj_t_conv}
    \end{subfigure}
    \begin{subfigure}{0.45\textwidth}
        \centering
        \includegraphics[width=1\linewidth]{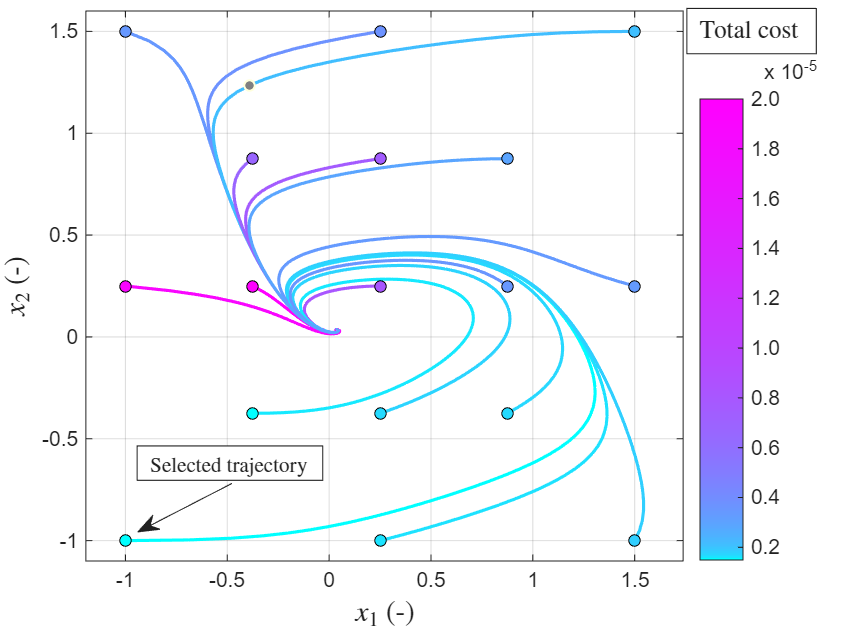}
        \caption{}
        \label{reftraj_ss}
    \end{subfigure}
    \caption{Reference trajectory selection: (a) relationship between $\mathcal{J}$ and $t_\mathrm{conv}$ and (b) the selected trajectories colored according to $\mathcal{J}$ (nonlinear color map for better readability).}
    \label{reftraj}
\end{figure}

\item \textbf{Effect of $\rho$ and $\gamma$ parameters.} 
The effects of $\rho$ and $\gamma$ are shown in Fig.~\ref{alpha_gamma_effects}. With increasing $\gamma$, the algorithm puts more emphasis on minimizing the KPDE cost at the expense of increased temporal cost. Thus, $\gamma$ controls the trade-off between the two objectives. Increasing $\rho$ improves the conditioning of the LS projections, but also introduces regularization bias. In the investigated range, this increased both cost terms.
\begin{figure}[!ht] 
    \centering 
    \begin{subfigure}{0.45\textwidth}
        \centering
        \includegraphics[width=1\linewidth]{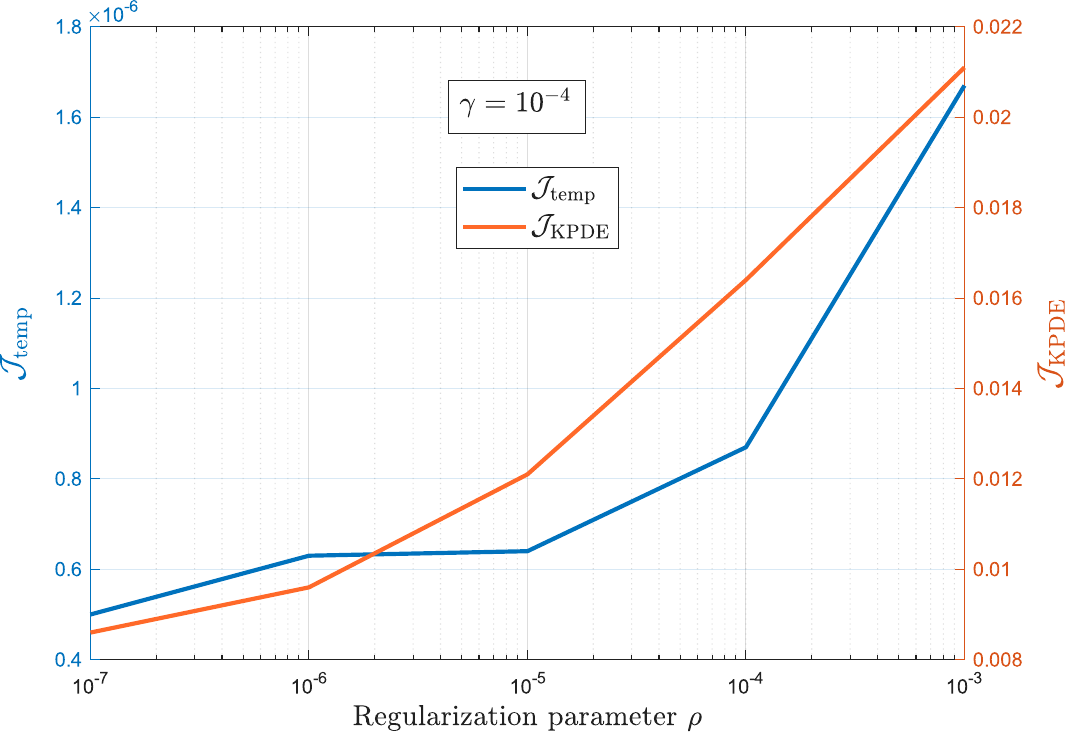}
        \caption{}
        \label{alpha_effects}
    \end{subfigure}
    \begin{subfigure}{0.45\textwidth}
        \centering
        \includegraphics[width=1\linewidth]{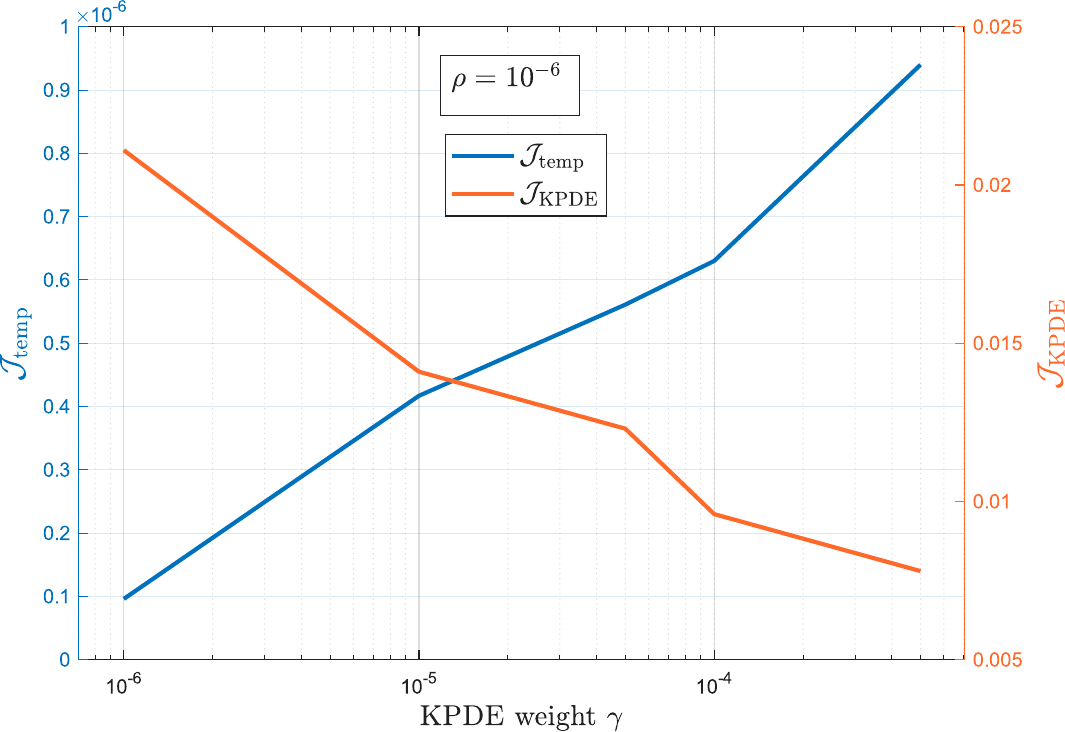}
        \caption{}
        \label{gamma_effects}
    \end{subfigure}
    \caption{The effect of (a) $\rho$ and (b) $\gamma$ on $\mathcal{J}_\mathrm{temp}$ and $\mathcal{J}_\mathrm{KPDE}$.}
    \label{alpha_gamma_effects}
\end{figure}

\item \textbf{Effect of model and grid dimensions.} 
The sensitivity to (i) the number of eigenvalues $n_\lambda$, (ii) the trajectory subgrid used to evaluate $\mathcal{J}_{\mathrm{temp}}$, (iii) the full grid of initial conditions, and (iv) the interpolation grid used for KPDE evaluation was investigated. The results are shown in Fig.~\ref{scaling_analysis}. Increasing $n_\lambda$ decreased both cost terms in the investigated range. The final $n_\lambda$ selection should be performed based on the procedure described in Section \ref{sec_model_order} to find the lowest-rank representation. Increasing the temporal subgrid size provided a more representative evaluation of $\mathcal{J}_\mathrm{temp}$ over the sampled region while having little effect on $\mathcal{J}_{\mathrm{KPDE}}$. Increasing the full initial-condition grid from $11 \times 11$ to $31 \times 31$ only moderately affected both cost functions, indicating a lower sensitivity to the trajectory sampling density within the investigated range. Finally, the interpolation grid was varied from $50 \times 50$ to $200 \times 200$. The KPDE cost increased for denser grids, which corresponds to the finer resolution of regions with larger local KPDE residuals, particularly near the boundaries of the sampled region. 
\begin{figure}[!ht] 
    \centering 
    \begin{subfigure}{0.45\textwidth}
        \centering
        \includegraphics[width=1\linewidth]{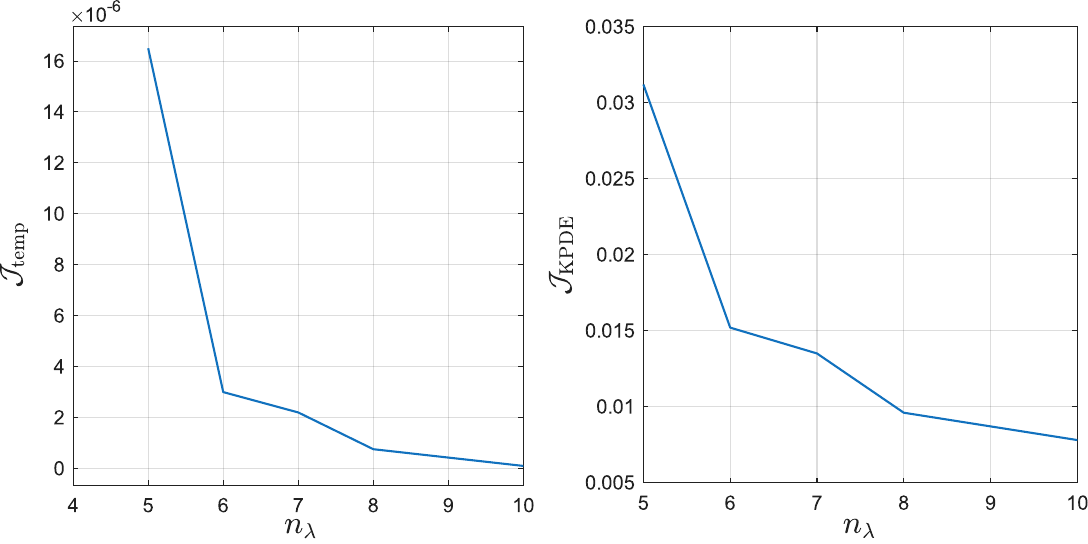}
        \caption{}
        \label{scaling_n_lambda}
    \end{subfigure}
    \begin{subfigure}{0.45\textwidth}
        \centering
        \includegraphics[width=1\linewidth]{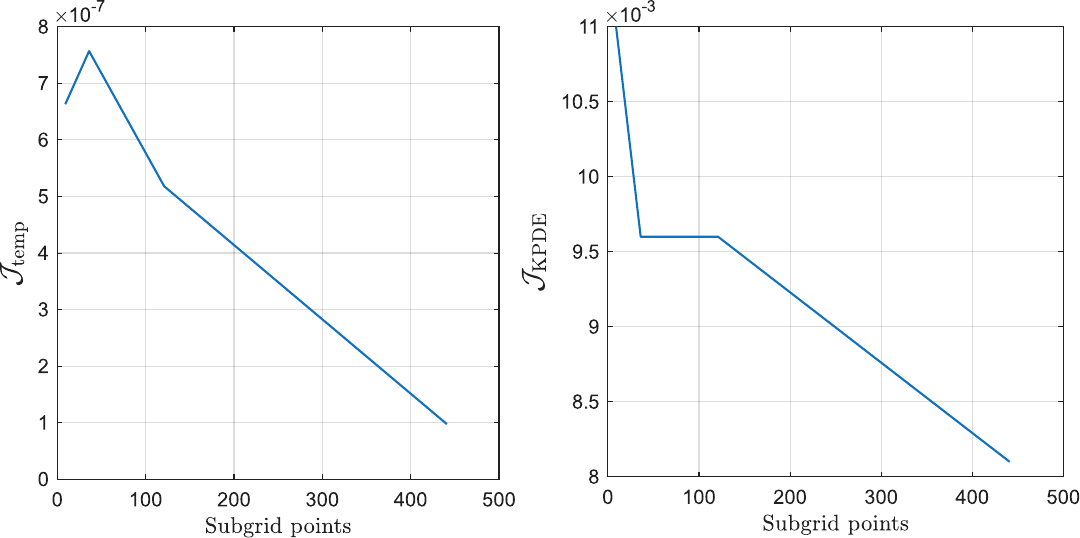}
        \caption{}
        \label{scaling_subgrid}
    \end{subfigure} 
    \begin{subfigure}{0.45\textwidth}
        \centering
        \includegraphics[width=1\linewidth]{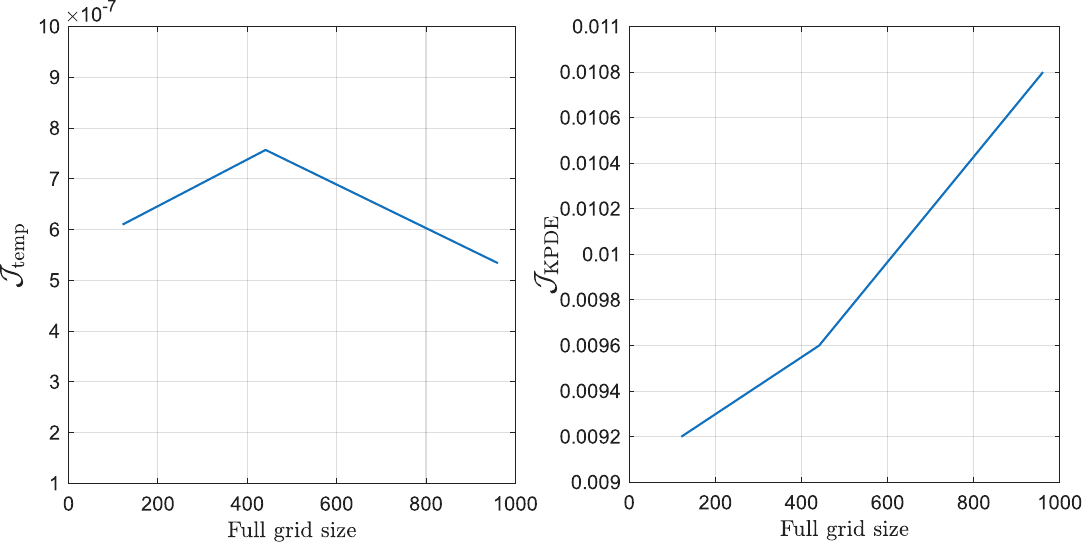}
        \caption{}
        \label{scaling_full_grid}
    \end{subfigure}
    \begin{subfigure}{0.45\textwidth}
        \centering
        \includegraphics[width=1\linewidth]{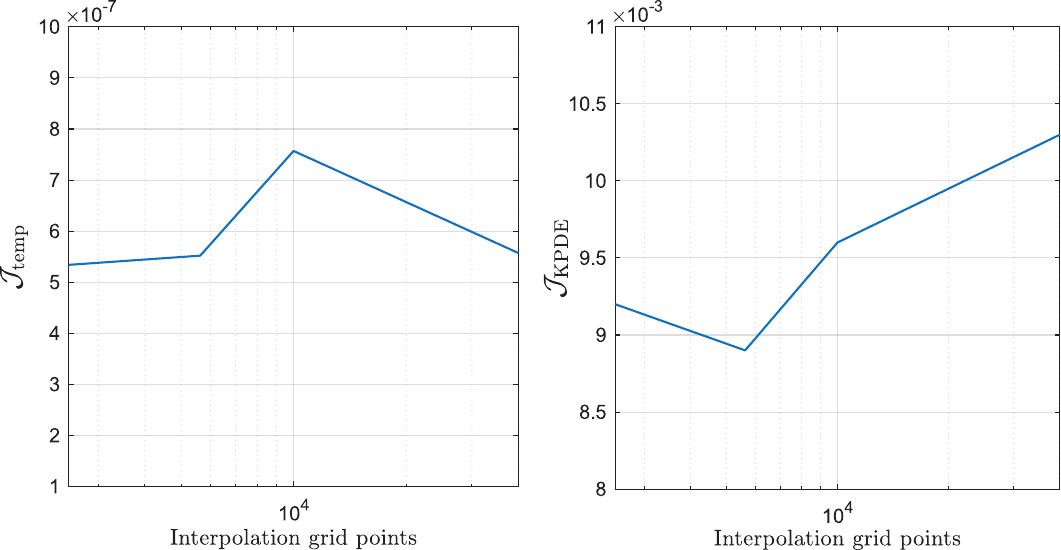}
        \caption{}
        \label{scaling_int_grid}
    \end{subfigure}
    \caption{The effect of (a) $n_\lambda$, (b) temporal subgrid size, (c) full grid size, and (d) interpolation grid size on $\mathcal{J}_\mathrm{temp}$ and $\mathcal{J}_\mathrm{KPDE}$.}
    \label{scaling_analysis}
\end{figure}

\item \textbf{Noise sensitivity.} 
The effect of noise on the eigenfunction identification phase was analyzed considering additive Gaussian white noise with noise variances from $10^{-6}$ to $2.5\times10^{-3}$. The results are shown in Fig.~\ref{noise_sens}. The identification remains accurate for noise variance below $6.25 \times 10^{-4}$ (standard deviation of 0.025). As expected, an increase in $\mathcal{J}_\mathrm{temp}$ follows the increasing variance. Noise also perturbs the coordinates of initial conditions associated with the eigenfunction mapping, which may distort the interpolated eigenfunction landscapes and their numerical gradients. This effect can be partially mitigated by increasing the smoothing used during spline interpolation. 
\begin{figure}[!ht]
    \centering
    \includegraphics[width=1\linewidth]{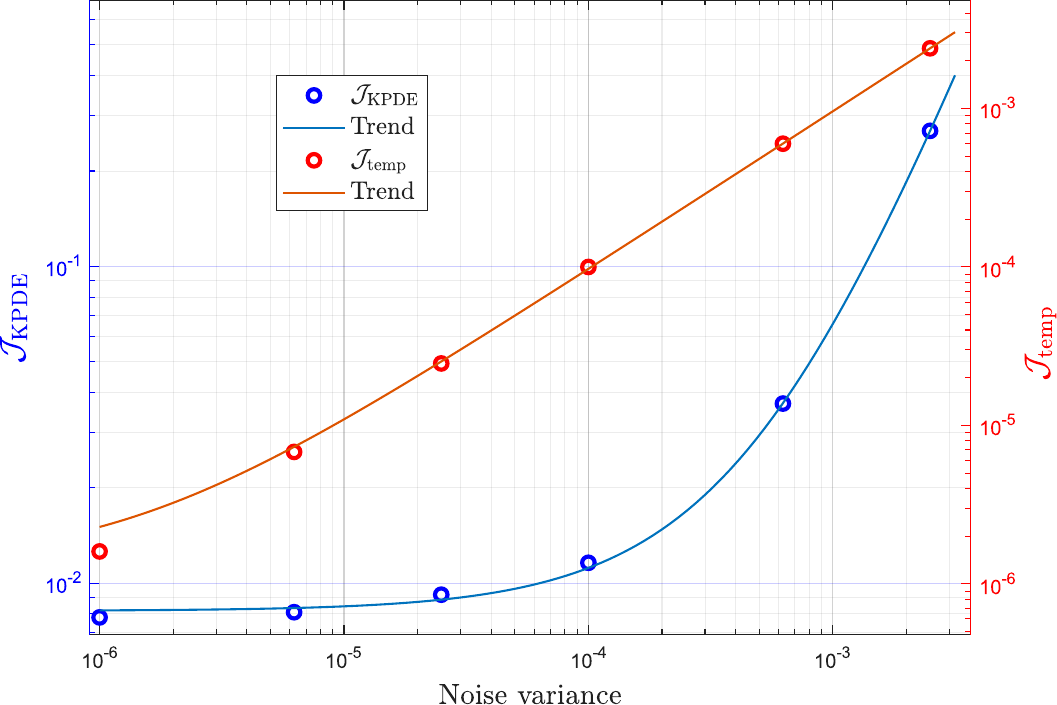}
    \caption{The effect of increasing the variance of additive Gaussian white noise on $\mathcal{J}_\mathrm{temp}$ and $\mathcal{J}_\mathrm{KPDE}$.}
    \label{noise_sens}
\end{figure}

\end{itemize}

\subsection{Limit cycle and isochrons: Van der Pol oscillator}
The van der Pol oscillator was selected to evaluate SADFED for a stable nonlinear limit cycle, including spectrum discovery and asymptotic phase identification. The system is
\begin{equation}
\begin{aligned}
    \dot{x}_1 &= x_2\,, \\
    \dot{x}_2 &= \mu (1 - x_1^2) x_2 - x_1\,,
\end{aligned}
\end{equation}
where $\mu$ was set to 2. The system is centrally symmetric, with $\mathbf{f(-x) = -f(x)}$, and exhibits both transient and periodic dynamics.

Trajectories were simulated from a $21 \times 21$ grid over $[-2.5, 2.5] \times [-4, 4]$ using an integration step of $\Delta t = 0.01$~s and a sampling period $T_\mathrm{s} = 0.1$~s for 20~s (201 samples), covering at least two cycle periods. The fundamental frequency was estimated as $\omega_1 \approx 0.824$~rad/s, corresponding to the eigenvalue pair $\lambda_{1,2} = \pm 0.824i$. Twelve additional complex-conjugate eigenvalue pairs were optimized using a PSO population of 200, 2000 NM iterations, $\rho = 10^{-6}$, $\gamma = 10^{-4}$, and a $100 \times 100$ interpolation grid ($\mathbf{\Delta x} = [0.05, 0.08]$).

\begin{figure}[!ht] 
    \centering 
    \begin{subfigure}{0.5\textwidth}
        \centering
        \includegraphics[width=1\linewidth]{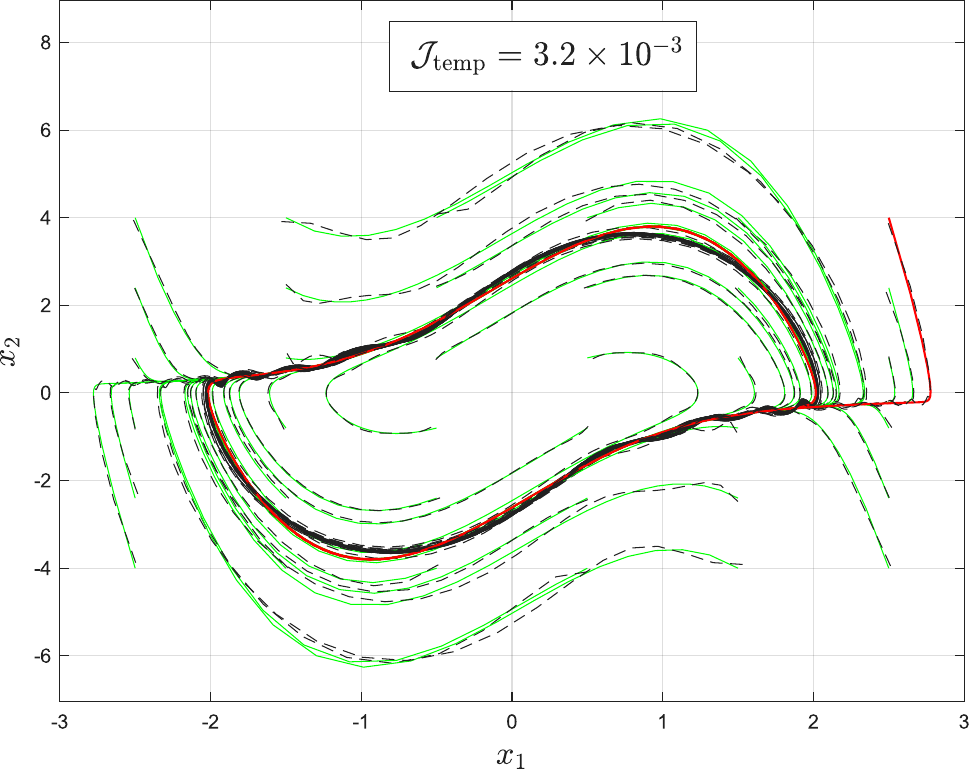}
        \caption{}
        \label{vdp_trajs}
    \end{subfigure}
    \begin{subfigure}{0.5\textwidth}
        \centering
        \includegraphics[width=1\linewidth]{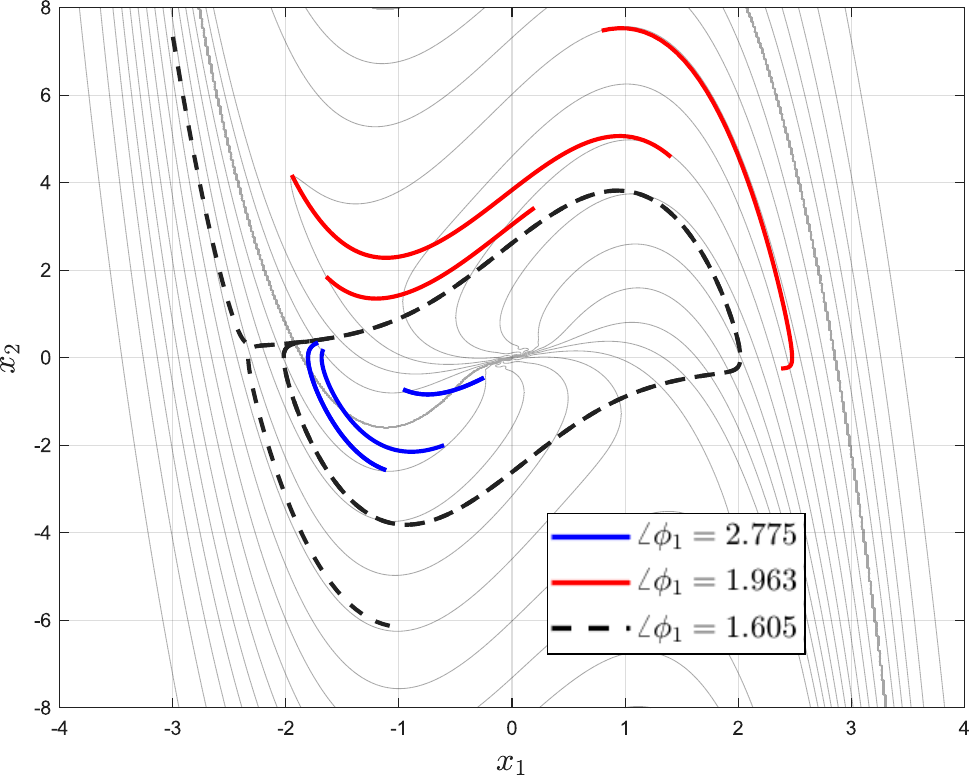}
        \caption{}
        \label{vdp_isochrons}
    \end{subfigure}
    \caption{(a) The reconstructed trajectories: ground truth (green solid), reconstruction (black dashed), and reference trajectory (red). (b) Level sets of the $\angle \varphi_1$ representing approximate isochrons, with selected trajectories propagated for equal time intervals.}
    \label{van_der_pol_fit_isochron}
\end{figure}

The resulting temporal cost was $\mathcal{J}_\mathrm{temp} = 3.2 \times 10^{-3}$. The reconstructed trajectories are shown in Fig.~\ref{vdp_trajs}. Since a large fraction of the training samples lies on the limit cycle, the temporal objective primarily addresses the periodic dynamics, while the fast transient reconstruction is less accurate. After fixing the identified eigenvalues, the eigenfunctions were refined using a $200 \times 200$ grid over $[-3,3] \times [-8,8]$. This demonstrates that the eigenfunction landscapes can be extended to a larger state-space region without repeating the eigenvalue optimization.

The phase of the fundamental eigenfunction, $\theta(\mathbf{x}) = \angle\varphi_1(\mathbf{x}) = \mathrm{atan2}\left(\varphi_\mathrm{1I}(\mathbf{x}),\varphi_\mathrm{1R}(\mathbf{x})\right)$, defines the approximate isochrons shown in Fig.~\ref{vdp_isochrons}. Trajectories initialized on the same level set pass through other level sets approximately at the same time and thus converge toward the limit cycle approximately with the same phase, consistent with the definition of isochrons~\cite{Mauroy2012, Mauroy2013}. For the depicted trajectories, the observed maximum phase-time deviations were $0.03$~s (blue), $0.14$~s (red), and $0.32$~s (black dashed), with larger errors for trajectories initialized farther from the limit cycle.

Additional oscillatory eigenvalue pairs were identified at approximately $\pm2.48i$, $\pm4.12i$, $\pm5.78i$, and $\pm7.42i$, corresponding to the third, fifth, seventh, and ninth harmonics of $\omega_1$, respectively. The identified frequencies approximately correspond to odd harmonics $(2k-1)\omega_1$, $k \in \mathbb{N}$, which is consistent with the central symmetry of the van der Pol dynamics. Once the fundamental frequency is estimated, the selected harmonics can therefore be included in $\mathbf{\Lambda}$, allowing the optimization to focus on transient eigenvalues. This demonstrates that SADFED successfully recovered the harmonic structure of the limit cycle from the data using a low-rank Koopman approximation.

\begin{figure*}[!ht] 
    \centering 
    \begin{subfigure}{0.45\textwidth}
        \centering
        \includegraphics[width=1\linewidth]{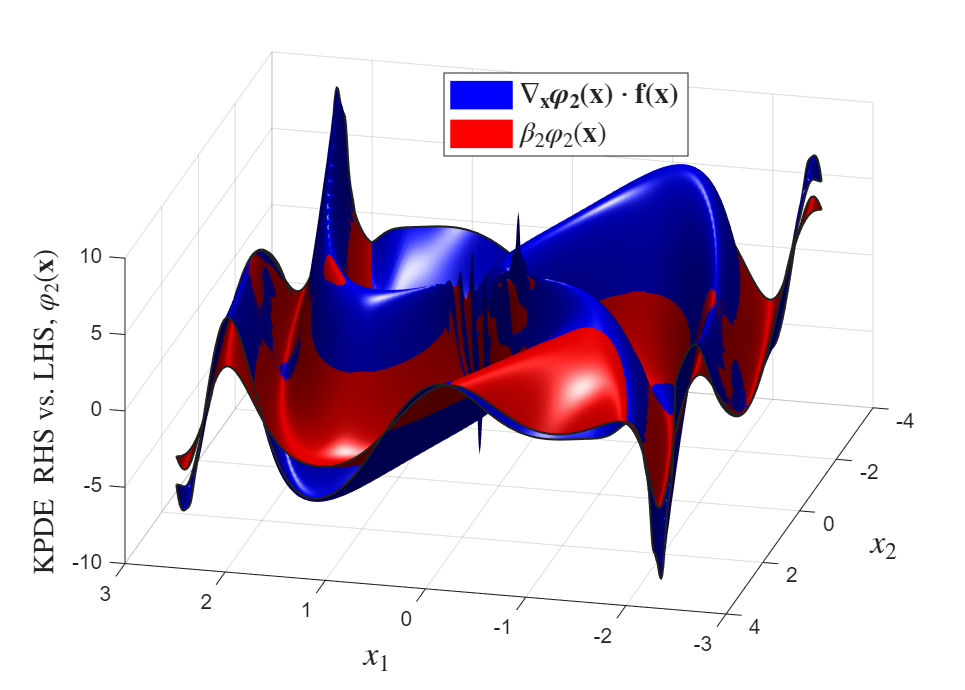}
        \caption{}
        \label{vdP_KPDE_1}
    \end{subfigure}
    \begin{subfigure}{0.45\textwidth}
        \centering
        \includegraphics[width=1\linewidth]{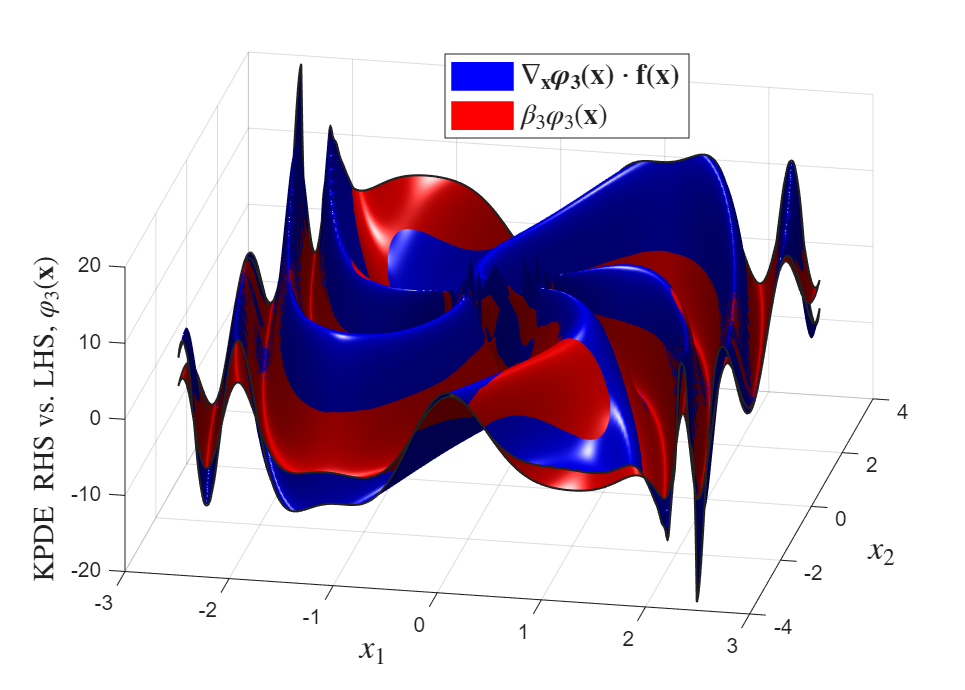}
        \caption{}
        \label{vdP_KPDE_2}
    \end{subfigure} 
    \begin{subfigure}{0.45\textwidth}
        \centering
        \includegraphics[width=1\linewidth]{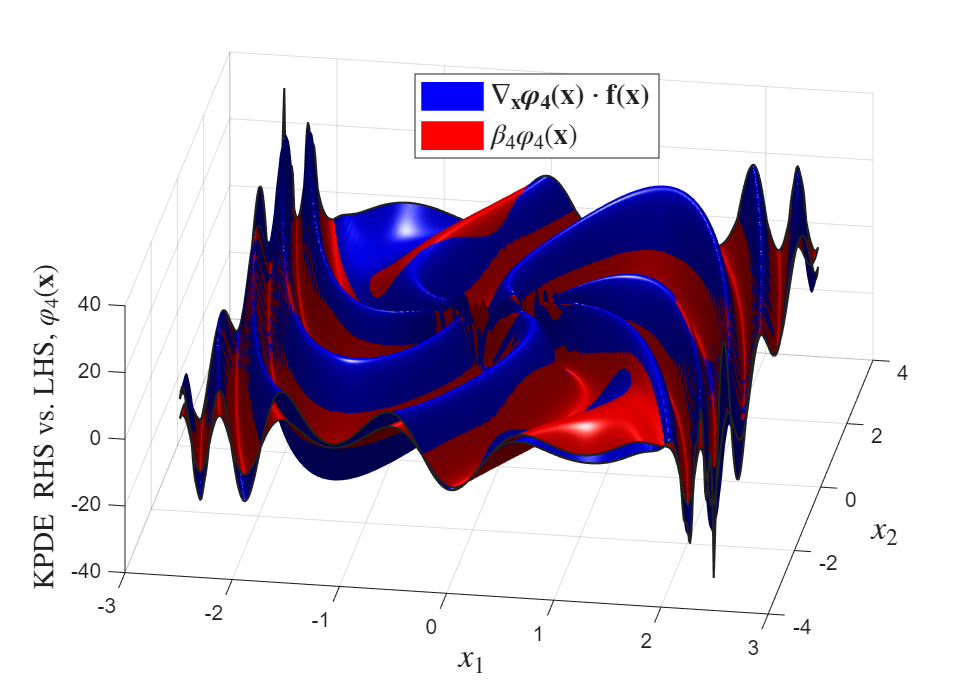}
        \caption{}
        \label{vdP_KPDE_3}
    \end{subfigure}
    \begin{subfigure}{0.45\textwidth}
        \centering
        \includegraphics[width=1\linewidth]{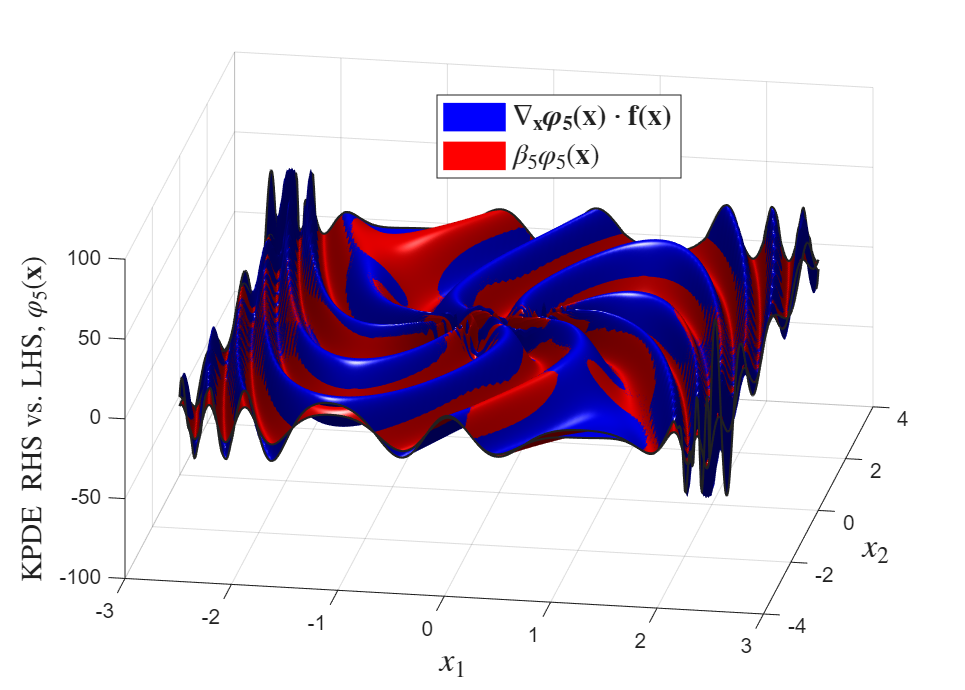}
        \caption{}
        \label{vdP_KPDE_4}
    \end{subfigure}
    \caption{Comparison of LHS (blue) and RHS (red) of the KPDE for eigenfunctions $\varphi_{2,...,5}$ of the van der Pol oscillator. The corresponding imaginary parts are (a) $\pm 2.48i$, (b) $\pm 4.12 i$, (c) $\pm 5.78i$, and (d) $\pm 7.42i$.}
    \label{van_der_pol_KPDE}
\end{figure*}

Figure~\ref{van_der_pol_KPDE} compares the unnormalized left- and right-hand sides of the KPDE for these four identified imaginary eigenfunctions. Notice the numerical artifacts near the origin, as the unstable fixed point causes singularities in the eigenfunctions. The increasing number of spiral branches corresponds to the odd-integer multiples of the fundamental eigenvalue $\lambda_1$. 

\subsection{Invariant subsets: Duffing system} \label{secDuffing}
The dissipative Duffing oscillator was selected to demonstrate identification across multiple basins of attraction and exploitation of state-space symmetry:
\begin{equation}
\begin{aligned}
    \dot{x}_1 &= x_2\,, \\
    \dot{x}_2 &= -\delta x_2 - x_1(b + a x_1^2)\,,
\end{aligned}
\end{equation}
with $\delta = 0.5$, $a = 1$ and $b = -1$.

With these parameters, the system has two stable fixed points at $[-1, 0]$ and $[1, 0]$ and a saddle point at $[0, 0]$. The stable fixed points have principal eigenvalues $\lambda_{1,2}^{s} = -0.25\pm i\sqrt{7.75}/2$, while the saddle has eigenvalues $\lambda_1^{u} = 0.781$ and $\lambda_2^{u} = -1.281$. The system admits two basins of attraction corresponding to the stable points, which are separated by the stable manifold of the saddle at the origin, forming an S-shaped separatrix. Since the system exhibits central symmetry, $\mathbf{f(-x) = -f(x)}$, the identification was performed using an $11 \times 21$ grid of initial conditions only in the $[0, 2] \times [-2, 2]$ half-plane, with the rest of the region reconstructed using the symmetry during refinement. Trajectories were simulated for 30~s with $\Delta t = 0.01$~s and sampling period $T_\mathrm{s} = 0.1$~s (301 samples).

To capture the basins of attraction, the zero eigenvalue $\lambda_0 = 0$ was prescribed together with the complex-conjugate eigenvalue pair $\lambda_{1,2}^{s}$ and the real eigenvalue $\lambda_2^{u}$. Seven additional complex-conjugate eigenvalue pairs were optimized, yielding $n_\lambda = 10$ in total. The optimization used a PSO population of 200, 2000 NM iterations, $\rho = 10^{-6}$, $\gamma = 10^{-5}$, and a $100 \times 100$ interpolation grid ($\mathbf{\Delta x} = [0.02, 0.04]$). The temporal subgrid size remained $6 \times 6$. Following Section~\ref{sec_basins}, each trajectory was classified based on its attracting fixed point, providing an approximate location of the separatrix. For the evaluation of $\mathcal{J}_\mathrm{KPDE}$, interpolation points within a neighborhood of the separatrix were excluded. The neighborhood was defined using manually constructed piecewise-linear boundaries.

\begin{figure}[!ht]
    \centering
    \includegraphics[width=1\linewidth]{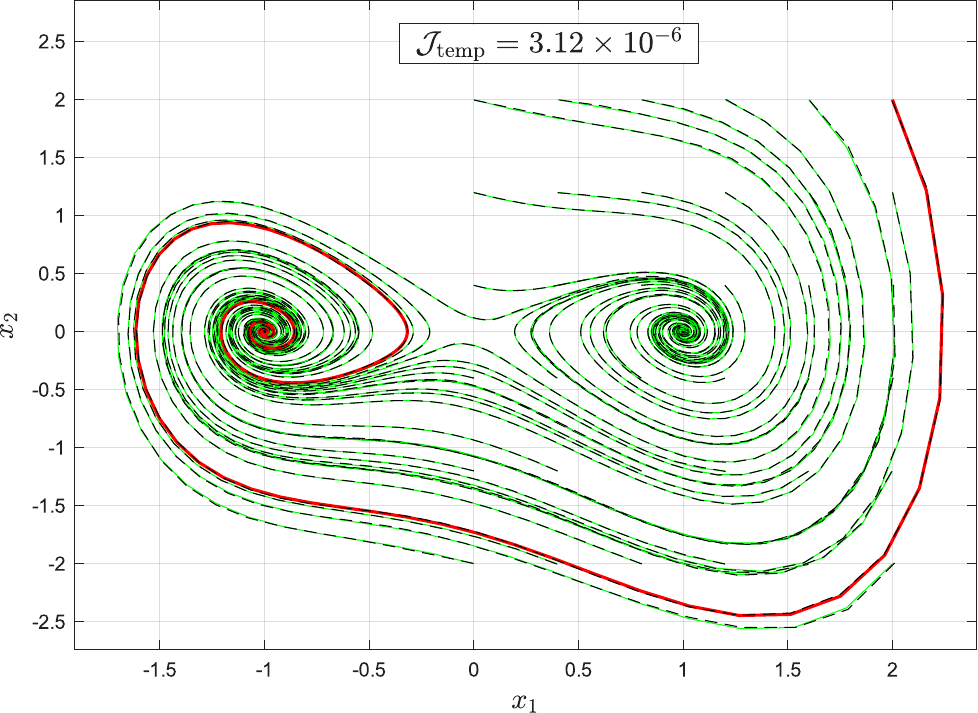}
    \caption{Reconstructed trajectories: ground truth (green solid) vs. reconstructed (black dashed), and the reference trajectory (red).}
    \label{duffing_fit_traj}
\end{figure}

The resulting costs were $\mathcal{J} = 6.9\times10^{-6}$, $\mathcal{J}_\mathrm{temp} = 3.12 \times 10^{-6}$, and $\mathcal{J}_\mathrm{KPDE} = 0.3792$, and the reconstructed trajectories on the sampled half-plane are shown in Fig.~\ref{duffing_fit_traj}. 

\begin{figure}[!ht]
    \centering
    \includegraphics[width=1\linewidth]{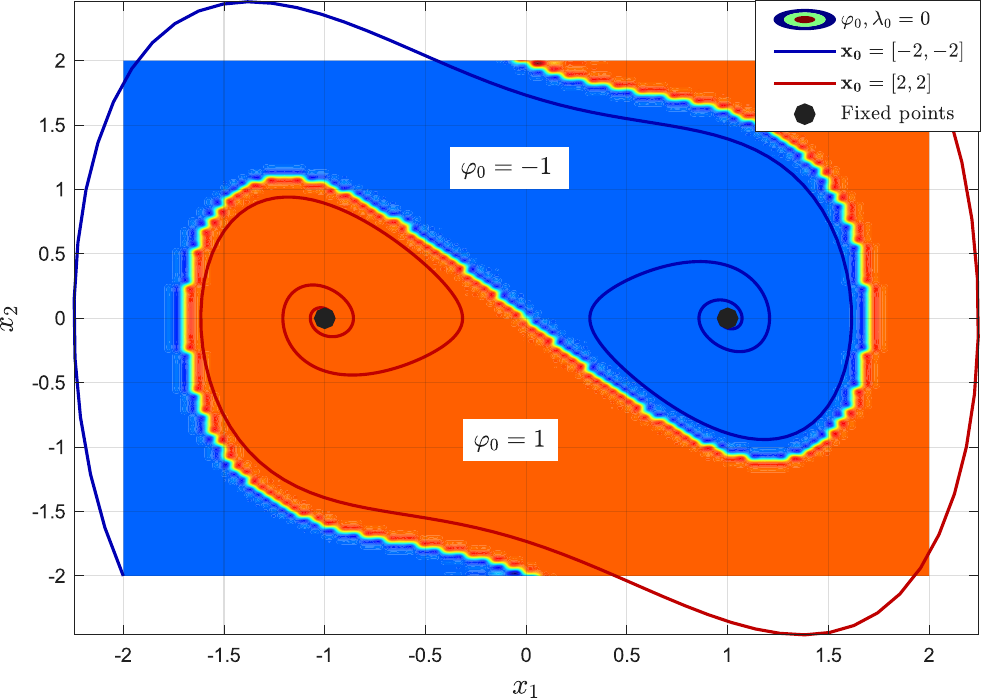}
    \caption{Refined zero-eigenvalue eigenfunction $\varphi_0$ distinguishing the two basins of attraction of the Duffing system. The transition between the approximately constant values identifies the S-shaped separatrix.}
    \label{duffing_invar_subs}
\end{figure}

The identified zero-eigenvalue eigenfunction $\varphi_0$ is shown in Fig.~\ref{duffing_invar_subs}. Its approximately piecewise-constant values distinguish the two basins of attraction. This is consistent with Koopman representations of systems containing multiple invariant sets~\cite{Pan2024}. 

\begin{figure}[!ht]
    \centering
    \includegraphics[width=1\linewidth]{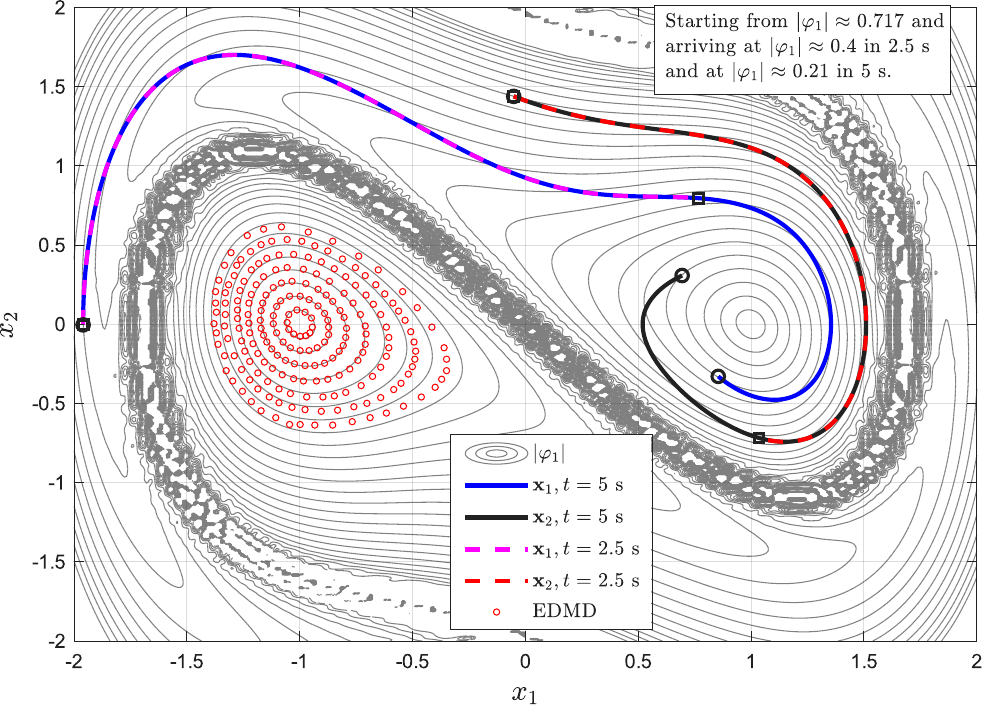}
    \caption{Level sets of $|\varphi_1|$ giving the isostables of both stable fixed points of the Duffing system. Two trajectories starting from the same level set approached another level set after the same time. The times were 2.5~s (blue and black dashed lines) and 5~s (cyan and violet solid lines). Red squares denote the isostables obtained by EDMD~\cite{Williams2015}.}
    \label{duffing_invar_isostab}
\end{figure}

The isostables obtained as level sets of $|\varphi_1(\mathbf{x})|$ are shown in Fig.~\ref{duffing_invar_isostab}. The eigenfunction was refined using a $200 \times 200$ initial-condition grid. For comparison, EDMD results reproduced following~\cite{Williams2015} with 1000 TPS RBFs are shown by red markers near the fixed point; the contour levels need not coincide because Koopman eigenfunctions are defined up to a nonzero scaling factor. 

Two trajectories initialized on the same level set $|\varphi_1| = 0.717$ reach approximately the same subsequent level sets, $|\varphi_1| = 0.4$ and $|\varphi_1| = 0.21$, after 2.5~s and 5~s, respectively, confirming that the level sets are approximate isostables. Here, the same eigenfunction describes both basins, while the separatrix remains apparent as a discontinuity between them.

\subsection{Unknown dynamics and control: two-spool turbojet engine}
The two-spool turbojet gas turbine engine (GTE) was selected as a practical nonlinear system with unknown, control-nonaffine dynamics. It operates in an admissible region bounded by safety limits corresponding primarily to surge, overpressure in the combustion chamber, combustor blowout, and overtemperature in the high-pressure turbine~\cite{MattiglyBook, Jaw2009}.

A nonlinear physics-based component-level model implemented in MATLAB R2025a and based on GasTurb 15~\cite{GasTurb} was used to generate an input-driven trajectory covering most of the admissible operating region and approaching the operational limits. Gaussian white noise was added to the simulated data, and all variables were normalized to $[0, 1]$.

The underlying governing equations for spool speeds $N_1$ and $N_2$ were discovered using the SINDy approach with linear and logistic functions as a basis and nonlinear parameter learning as described in~\cite{GrasevSpringer2026}. The resulting model is
\begin{align}
    \underbrace{\begin{bmatrix} \dot{N}_1 \\ \dot{N}_2 \end{bmatrix}}_{\mathbf{\dot{x}}} = 
    \underbrace{\begin{bmatrix} f_1 \\ f_2 \end{bmatrix}}_{\mathbf{f(x)}} + 
    \underbrace{\begin{bmatrix} g_{11} & g_{12} \\ g_{21} & g_{22} \end{bmatrix}}_{\mathbf{G(x)}} 
    \underbrace{\begin{bmatrix}
        W_\mathrm{f} \\ A_\mathrm{n}
    \end{bmatrix}}_{\mathbf{u}},
\end{align}
where $\mathbf{x}=[N_1\ N_2]^\intercal$ contains the normalized low- and high-pressure spool speeds, while $W_\mathrm{f}$ and $A_\mathrm{n}$ denote the fuel flow and nozzle area, respectively. The identified functions $\mathbf{f(x)}$ and $\mathbf{G(x)}$ are subsequently used as the drift and input vector fields for the SADFED model.

\subsubsection{Eigenfunction identification}
The autonomous drift $\mathbf{f(x)}$ was used to generate trajectories for SADFED identification from a $21 \times 21$ grid of initial conditions in $[-0.1, 1.1]^2$. The trajectories were simulated for 3~s with $\Delta t = 0.01$~s (301 samples). The fixed point corresponding approximately to the idle operating point lies sufficiently close to the origin that no additional state shift was required. 

Linearization at the fixed point yielded the principal eigenvalues $\lambda_{1,2} = -2.835 \pm 1.19i$, which were prescribed for identification. Four additional complex-conjugate eigenvalue pairs ($n_\varphi = 10$) were optimized using a PSO population of 50 followed by 2000 NM iterations. The optimization parameters were $\rho = 10^{-6}$ and $\gamma = 10^{-4}$, with a $100 \times 100$ interpolation grid used for KPDE evaluation ($\Delta\mathbf{x} = [0.012, 0.012]$).

\begin{figure}[!ht]
    \centering
    \includegraphics[width=1\linewidth]{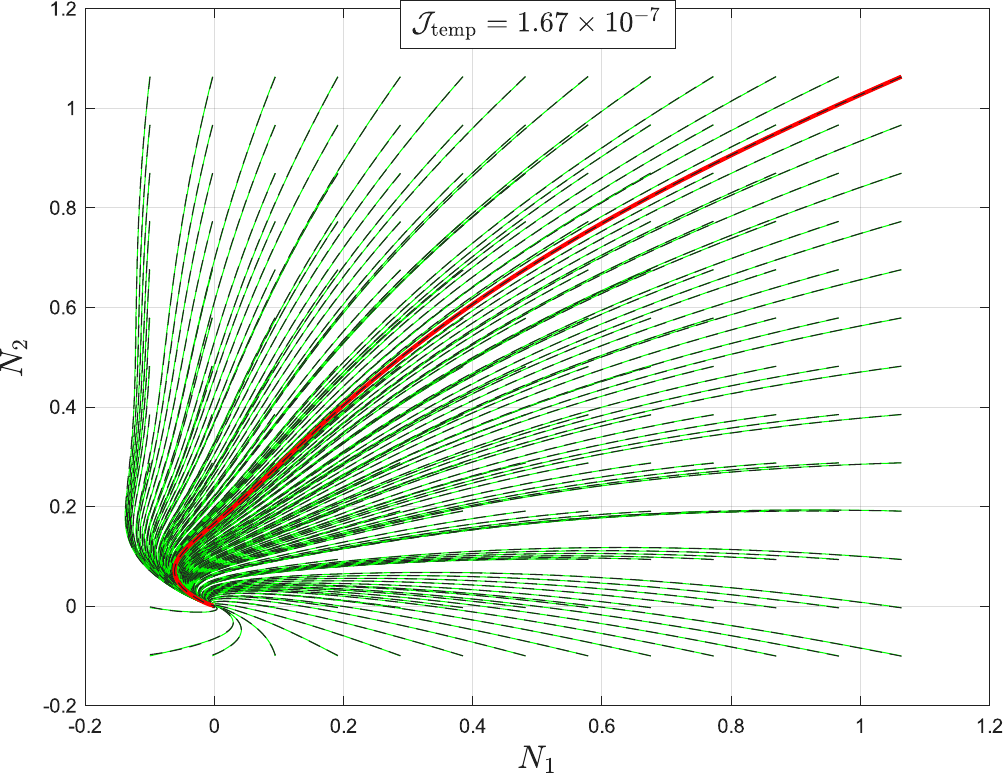}
    \caption{Reconstructed autonomous trajectories and the temporal cost of the two-spool GTE. Ground truth (green solid) vs. reconstruction (black dashed), and the reference trajectory (red).}
    \label{2spool_auton_fit}
\end{figure}

The resulting costs were $\mathcal{J}_\mathrm{temp} = 1.67\times10^{-7}$ and $\mathcal{J}_\mathrm{KPDE} = 0.0052$, and the reconstructed autonomous trajectories are shown in Fig.~\ref{2spool_auton_fit}.

\begin{figure*}[!ht] 
    \centering 
    \begin{subfigure}{0.45\textwidth}
        \centering
        \includegraphics[width=1\linewidth]{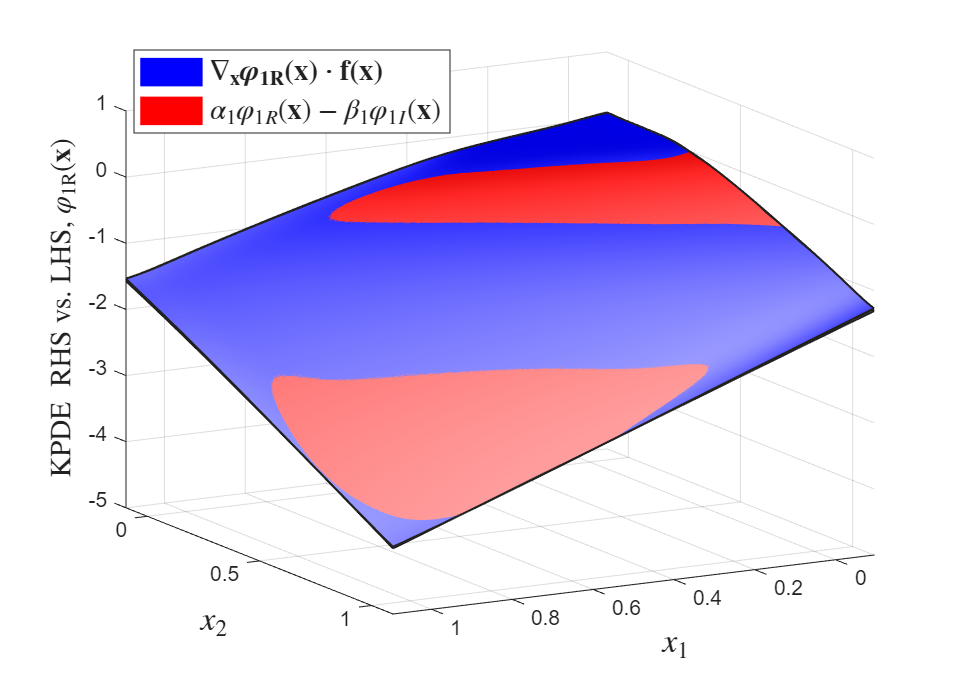}
        \caption{}
        \label{GTE_KPDE_1}
    \end{subfigure}
    \begin{subfigure}{0.45\textwidth}
        \centering
        \includegraphics[width=1\linewidth]{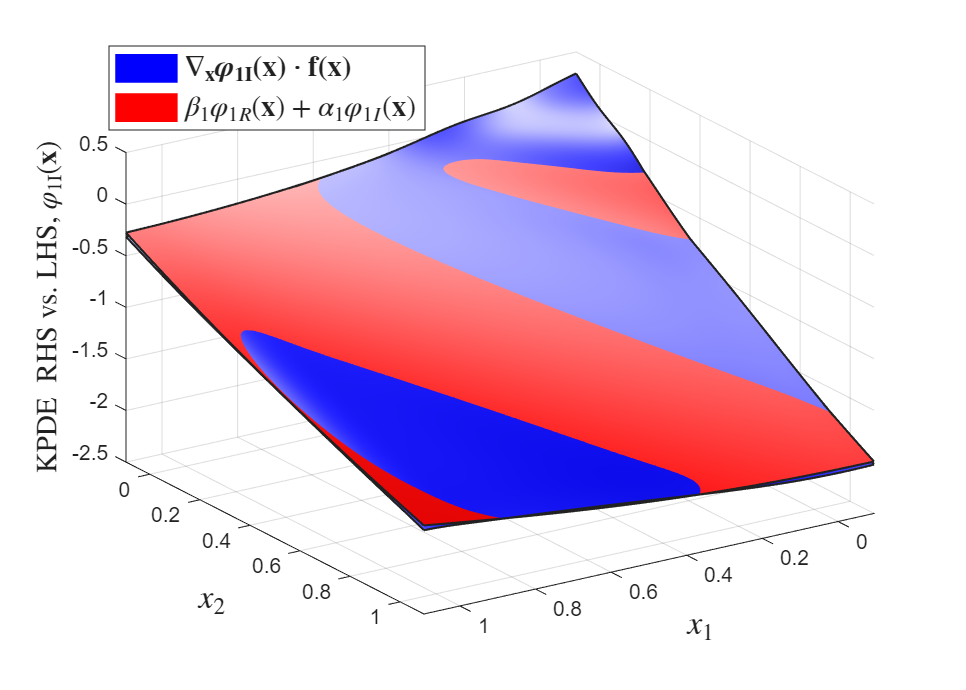}
        \caption{}
        \label{GTE_KPDE_2}
    \end{subfigure} 
    \begin{subfigure}{0.45\textwidth}
        \centering
        \includegraphics[width=1\linewidth]{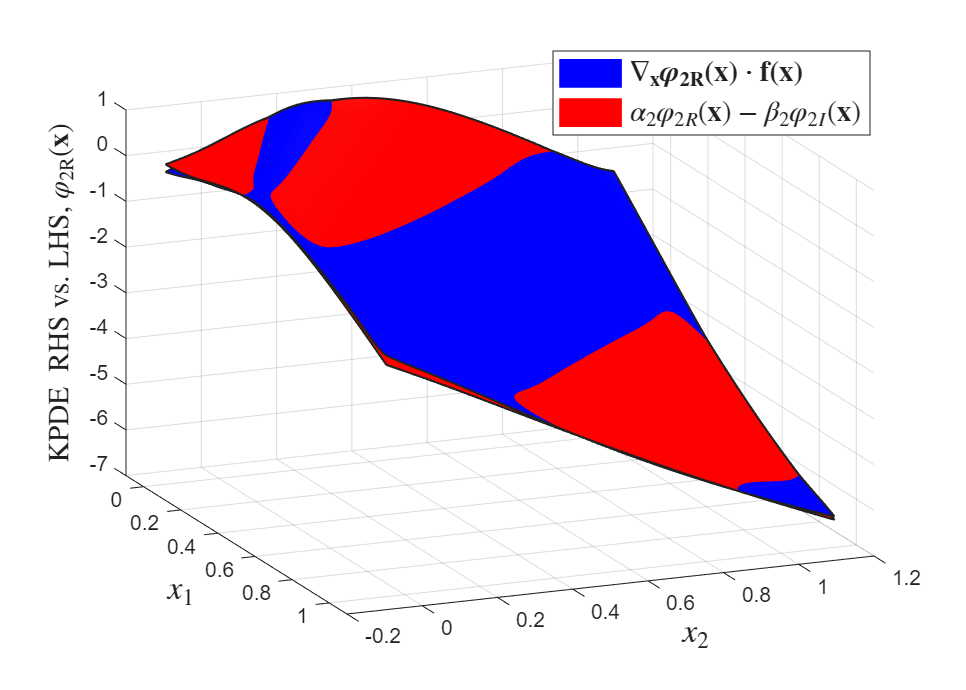}
        \caption{}
        \label{GTE_KPDE_3}
    \end{subfigure}
    \begin{subfigure}{0.45\textwidth}
        \centering
        \includegraphics[width=1\linewidth]{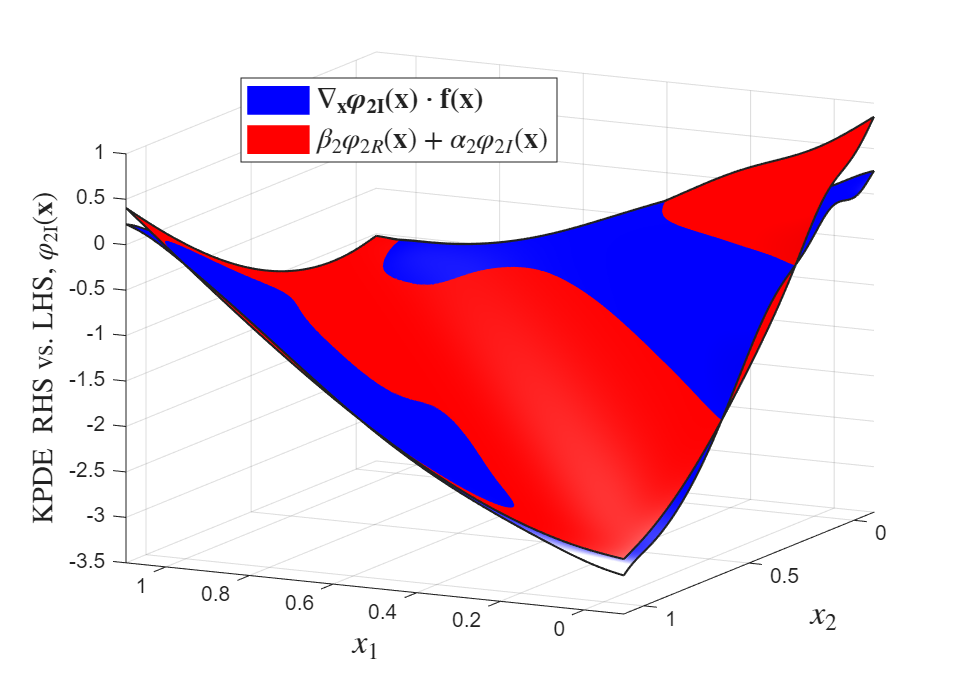}
        \caption{}
        \label{GTE_KPDE_4}
    \end{subfigure}
    \caption{Comparison of LHS (blue) and RHS (red) of the KPDE terms for the first 4 eigenfunctions of the two-spool GTE. The corresponding eigenvalues are (a,b) $\lambda_{1,2} = -2.835 \pm 1.19i$ and (c,d) $\lambda_{3,4} = -2.85 \pm 3.29i$.}
    \label{2spool_KPDE}
\end{figure*}

Figure~\ref{2spool_KPDE} compares KPDE left- and right-hand sides for the first four eigenfunctions, showing close agreement over the sampled region. 

\begin{figure}[!ht]
    \centering
    \includegraphics[width=1\linewidth]{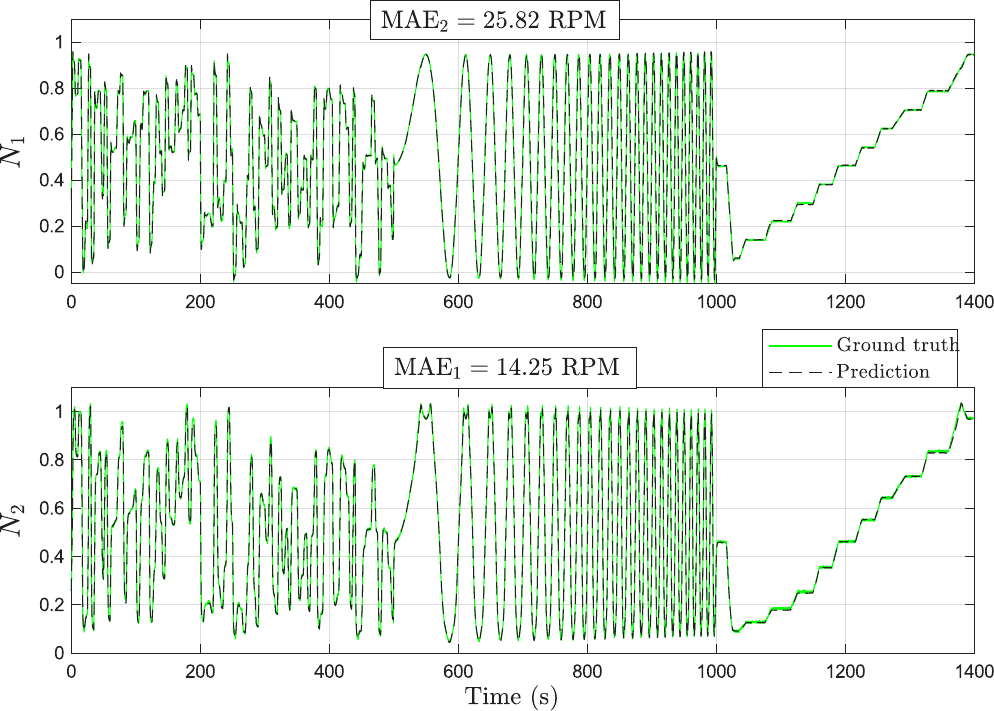}
    \caption{Fitted input-driven validation data of the two-spool GTE with the corresponding mean absolute errors.}
    \label{2spool_input_SS_fit}
\end{figure}

After fixing the identified eigenvalues, the eigenfunctions and their gradients were refined using a $100 \times 100$ initial-condition grid and a $500 \times 500$ interpolation grid. The lifted input dynamics were then computed directly as $\mathbf{\Gamma(x) = \nabla_x\Phi(x) G(x)}$. For efficient online computation, the entries of $\mathbf{\Gamma(x)}$ were approximated using 15 sigmoid basis functions according to~(\ref{eq_input_fit}). The surrogate parameters were optimized using the Adam optimizer. The resulting predictor was validated on an independent input-driven trajectory, as shown in Fig.~\ref{2spool_input_SS_fit}, yielding denormalized errors $\mathrm{MAE} = \mathrm{MAE}_1 + \mathrm{MAE}_2 = 40.07$~RPM.

\begin{figure}[!ht]
    \centering
    \includegraphics[width=1\linewidth]{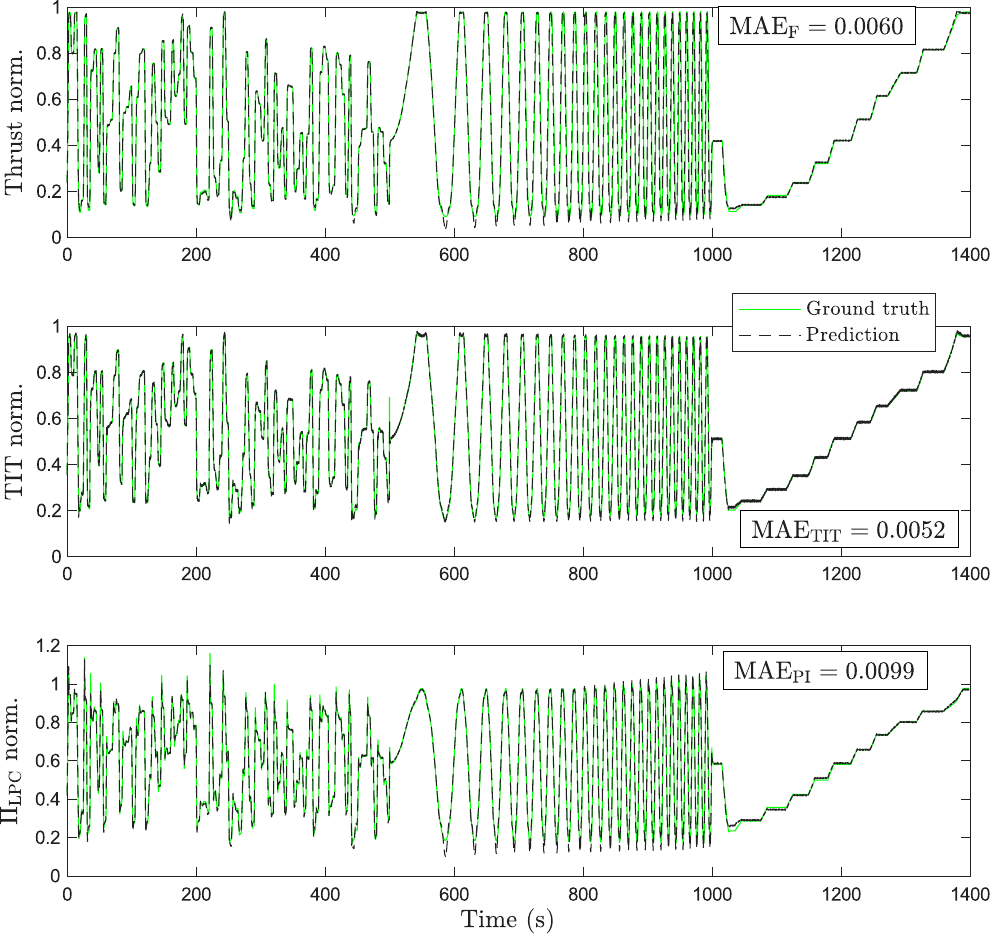}
    \caption{Fitted example output variables of the two-spool GTE: (top) thrust, (center) turbine inlet temperature, and (bottom) low-pressure compressor pressure ratio.}
    \label{2spool_F_TIT_PI_fit}
\end{figure}

Additional engine quantities can be reconstructed by projection onto the augmented lifted coordinates $[\mathbf{\Phi \ u}]^\intercal$. Figure~\ref{2spool_F_TIT_PI_fit} illustrates this for thrust, turbine inlet temperature, and low-pressure compressor pressure ratio.

\subsubsection{Kalman filter design} \label{sec_kalman}
A steady-state Kalman filter (KF) was designed in the lifted coordinates to improve the spool-speed estimates. Since $\mathbf{\Lambda}$ and $\mathbf{C}$ are time-invariant, a constant observer gain can be obtained from the algebraic Riccati equation for fixed process and measurement noise covariances, as shown in Appendix \ref{sec_A1}. An effective measurement noise covariance was estimated from the residuals between the measured states and their lifted reconstruction at each time step, $\mathrm{Cov}(\mathbf{x} - \mathbf{C_x\Phi(x)})$, while an effective process noise covariance was estimated from the residuals between the predicted eigenfunction dynamics and dynamics evaluated using the measured data $(\mathbf{x, \dot{x}})$: $\mathrm{Cov}\left( \mathbf{\nabla_x\Phi(x) \dot{x} - \Lambda\Phi(x)-\Gamma(x)u} \right)$. The resulting covariance matrices were used to tune $\mathbf{Q_o}$ and $\mathbf{R_o}$ and compute the steady-state Kalman gain.

\subsubsection{Gain-scheduled spool-speed controller design} \label{sec_GS_LQG}
A gain-scheduled tracking LQG controller was designed in the lifted state space to demonstrate the use of the identified SADFED model for control. Both spool speeds were controlled using fuel flow $W_\mathrm{f}$ and nozzle area $A_\mathrm{n}$. Their actuator dynamics were represented by first-order lags~\cite{Jaw2009}
\begin{align}\label{act_eqn}
    \underbrace{\begin{bmatrix} \dot{W}_\mathrm{f} \\[6pt] \dot{A}_\mathrm{n}\end{bmatrix}}_{\mathbf{\dot{x}_{act}}} = 
    \underbrace{\begin{bmatrix} -\frac{1}{T_\mathrm{f}} & 0 \\[6pt] 0 & -\frac{1}{T_\mathrm{a}}
    \end{bmatrix}}_{\mathbf{A_{act}}} 
    \underbrace{\begin{bmatrix} W_\mathrm{f} \\[6pt] A_\mathrm{n}
    \end{bmatrix}}_{\mathbf{x_{act}}} + \underbrace{\begin{bmatrix} \frac{1}{T_\mathrm{f}} & 0 \\[6pt] 0 & \frac{1}{T_\mathrm{a}} \end{bmatrix}}_{\mathbf{B_{act}}}
    \underbrace{\begin{bmatrix}
        W_\mathrm{f}^{c} \\[6pt] A_\mathrm{n}^{c}
    \end{bmatrix}}_{\mathbf{u_c}} \,,
\end{align}
where $T_\mathrm{f} = 0.06$ s and $T_\mathrm{a} = 0.1$ s are the fuel system and nozzle time constants, respectively, and $\mathbf{u_c} = [W_\mathrm{f}^{c}\ A_\mathrm{n}^{c}]^\intercal$ are the new control inputs. 

Augmenting the actuator states with the lifted dynamics gives the state-dependent system
\begin{align}
    \begin{bmatrix} \mathbf{\dot{\Phi}} \\
    \mathbf{\dot{x}_{act}}
    \end{bmatrix} = \begin{bmatrix} \mathbf{\Lambda} &  \mathbf{\Gamma(x)} \\ \mathbf{0} & \mathbf{A_{act}} \end{bmatrix}
    \begin{bmatrix} \mathbf{\Phi} \\ \mathbf{x_{act}} \end{bmatrix} + \begin{bmatrix} \mathbf{0} \\ \mathbf{B_{act}} \end{bmatrix} \mathbf{u_c} \,.
\end{align}

The operating region was sampled over the spool-speed space. At each scheduling point, $\mathbf{\Gamma(x)}$ was fixed to obtain a local linear time-invariant model, and the corresponding algebraic Riccati equation was solved, as shown in Appendix \ref{sec_A1}. The weight matrices were $\mathbf{Q_c} = \mathrm{blkdiag}(\mathbf{Q_\Phi, Q_{act}})$, where $\mathbf{Q}_\Phi = \mathbf{C_x^\intercal Q_x C_x}$, with $\mathbf{Q_x} = \mathrm{diag} (5, 5)$, $\mathbf{Q_{act}} = \mathrm{diag}(0.5, 0.1)$, and $\mathbf{R_c} = \mathrm{diag}(0.5, 0.1)$, penalizing mainly the fuel consumption. 

The resulting gain matrices were represented using the same sigmoid basis used for $\mathbf{\Gamma(x)}$, $\sigma(\mathbf{W_1 x + b_1})$, allowing efficient online evaluation, since it is also computed for $\mathbf{\Gamma}$. The gain matrix is thus
\begin{align} \label{eq_GS_LQG}
    \mathbf{K_{LQR}(x) = W_K \sigma(\mathbf{W_1 x + b_1})} + \mathbf{b_K}\,,
\end{align}
where $\mathbf{W_K}$ and $\mathbf{b_K}$ are the weights and biases. 

The resulting tracking law was
\begin{equation}
\begin{aligned}
    \mathbf{u_c} =\ \mathbf{u_{c,ss}} &-\mathbf{K_\Phi(\hat{\Phi} - \Phi_{ss})} \\ 
    &-\mathbf{K_{WA}(x_{act} - x_{act,ss}) - K_i \boldsymbol{\eta}}\,,
\end{aligned}
\end{equation}
where $\begin{bmatrix} \mathbf{K_\Phi} & \mathbf{K_{WA}}
\end{bmatrix} = \mathbf{K_{LQR}}$, $\boldsymbol{\eta}$ is a vector of control error integrals satisfying $\dot{\boldsymbol{\eta}} = \mathbf{x - x_r}$, and $\mathbf{(\cdot)_{ss}}$ denotes the desired steady-state values obtained from 
\begin{align} \label{eq_SS_NL_NH}
    \begin{bmatrix}
        \mathbf{\Lambda} & \mathbf{\Gamma(x_r)} & \mathbf{0} \\ 
        \mathbf{0} & \mathbf{A_{act}} & \mathbf{B_{act}} \\
        \mathbf{C_{x}} & \mathbf{0} & \mathbf{0}
    \end{bmatrix}
    \begin{bmatrix}
        \mathbf{\Phi_{ss}} \\ \mathbf{x_{act,ss}} \\ \mathbf{u_{c,ss}}
    \end{bmatrix} = 
    \begin{bmatrix}
        \mathbf{0} \\ \mathbf{0} \\ \mathbf{x_r}
    \end{bmatrix}\,,
\end{align}
with $\mathbf{x_r} = \begin{bmatrix} N_1^\mathrm{set} & N_2^\mathrm{set} \end{bmatrix}^\intercal$ the setpoint vector. 

\begin{figure*}
\centering 
    \begin{subfigure}{0.45\textwidth}
        \centering
        \includegraphics[width=0.85\linewidth]{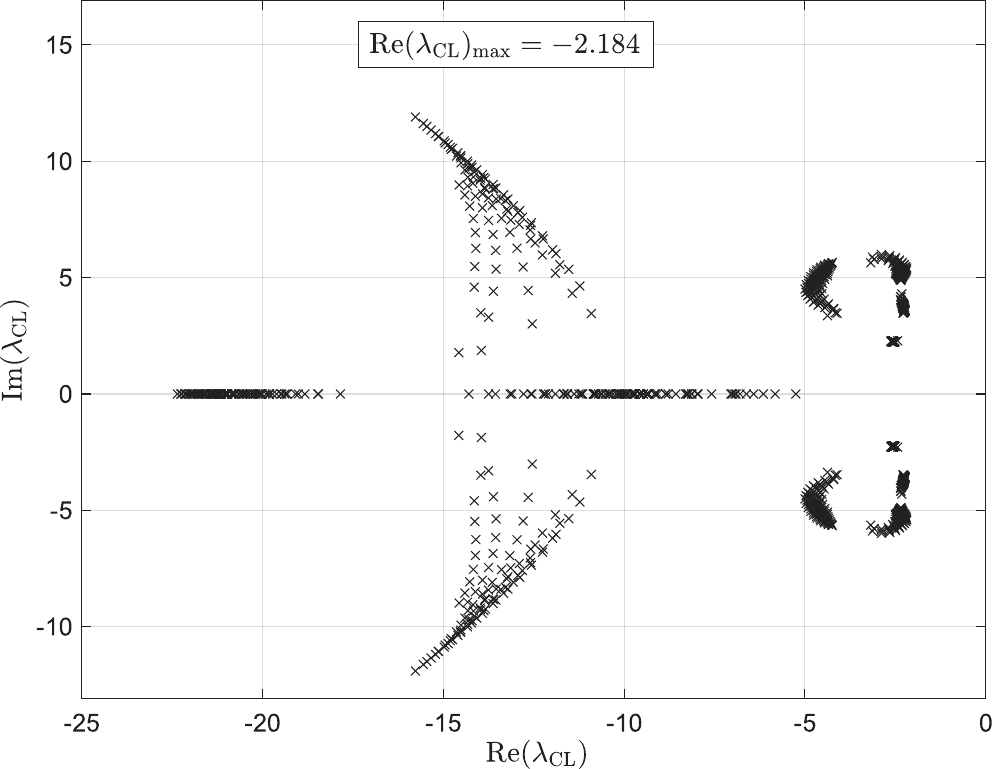}
        \caption{}
        \label{CL_eigs_complex}
    \end{subfigure}
    \begin{subfigure}{0.45\textwidth}
        \centering
        \includegraphics[width=1\linewidth]{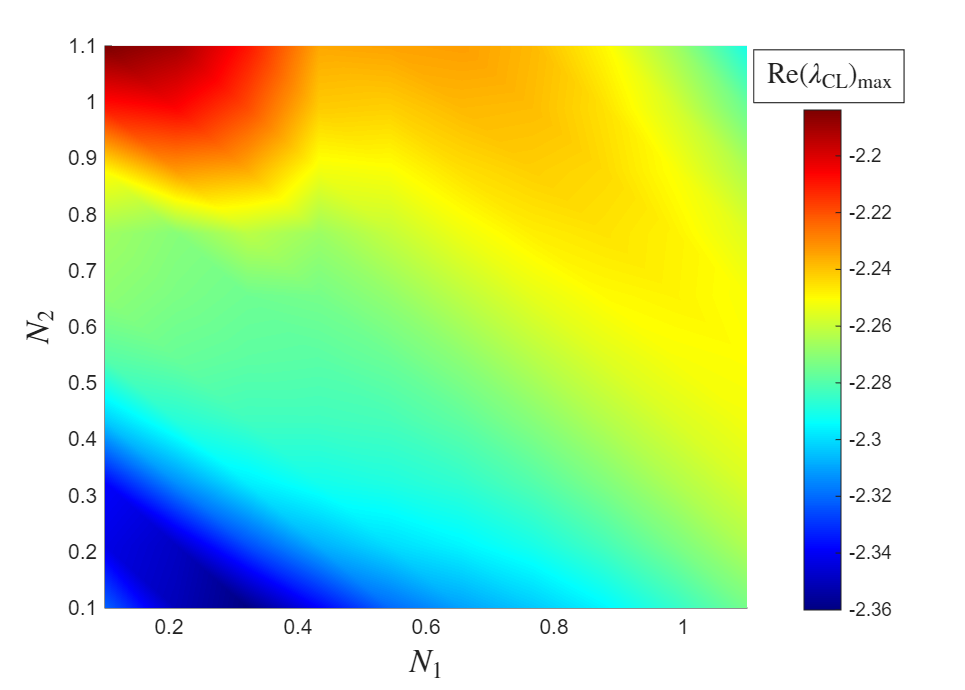}
        \caption{}
        \label{CL_eigs_distribution}
    \end{subfigure}
    \caption{The closed-loop eigenvalues, $\mathbf{\Lambda_{CL}}$: (a) distribution in the complex plane, and (b) maximum real parts vs. $N_1$ and $N_2$.}
    \label{CL_eigs}
\end{figure*}

Integral action was added to mitigate the steady-state offset caused by model uncertainty, with $\mathbf{K_i} = \mathrm{diag}(0.5, 0.5)$. All frozen-parameter closed-loop matrices evaluated on the scheduling grid were Hurwitz, with $\max(\Re(\lambda_{\mathrm{CL}})) = -2.184$. The resulting eigenvalues are shown in Fig.~\ref{CL_eigs}. This provides a local numerical stability check over the scheduling region. Input saturation was enforced using the engine limiters, and clamping anti-windup was applied during saturation. Further details of the control implementation are provided in Refs.~\cite{Jaw2009, GrasevSpringer2026}.

\subsubsection{Control results}
A classical proportional-integral (PI) controller was used as a well-established baseline. Its gains were tuned to provide safe full acceleration and deceleration with settling times below approximately 3~s and limited absolute overshoots (OS). The resulting gains were $K_\mathrm{p,1} = 0.5$ and $K_\mathrm{i,1} = 1$ for $N_1$, and $K_\mathrm{p,2} = 40$ and $K_\mathrm{i,2} = 60$ for $N_2$, relative to the design-point speeds $N_\mathrm{1,DP} = 9700$~RPM and $N_\mathrm{2,DP} = 13200$~RPM.

\begin{figure*}
\centering 
    \begin{subfigure}{0.45\textwidth}
        \centering
        \includegraphics[width=0.9\linewidth]{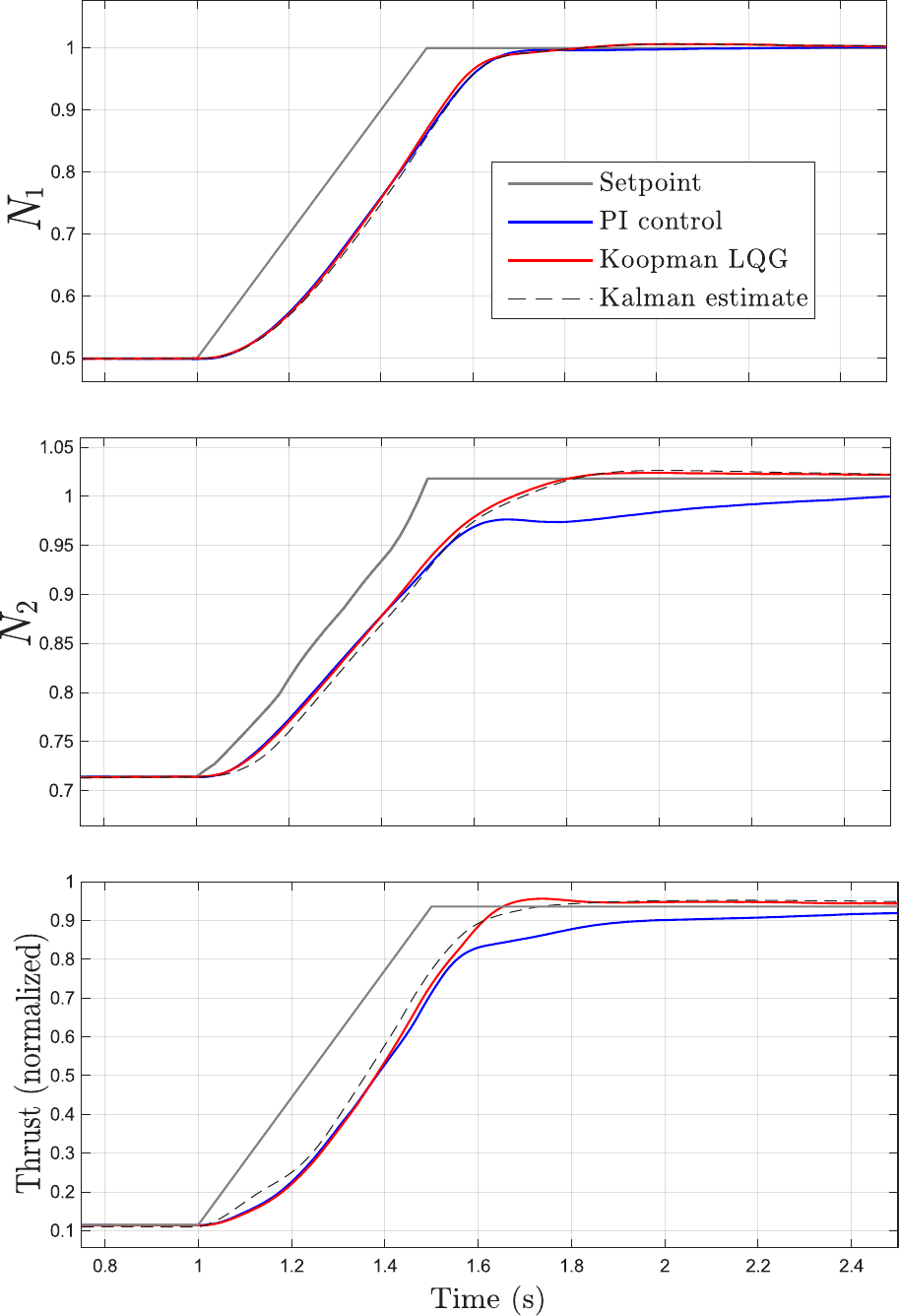}
        \caption{}
        \label{2spool_LQR_Accel}
    \end{subfigure}
    \begin{subfigure}{0.45\textwidth}
        \centering
        \includegraphics[width=0.9\linewidth]{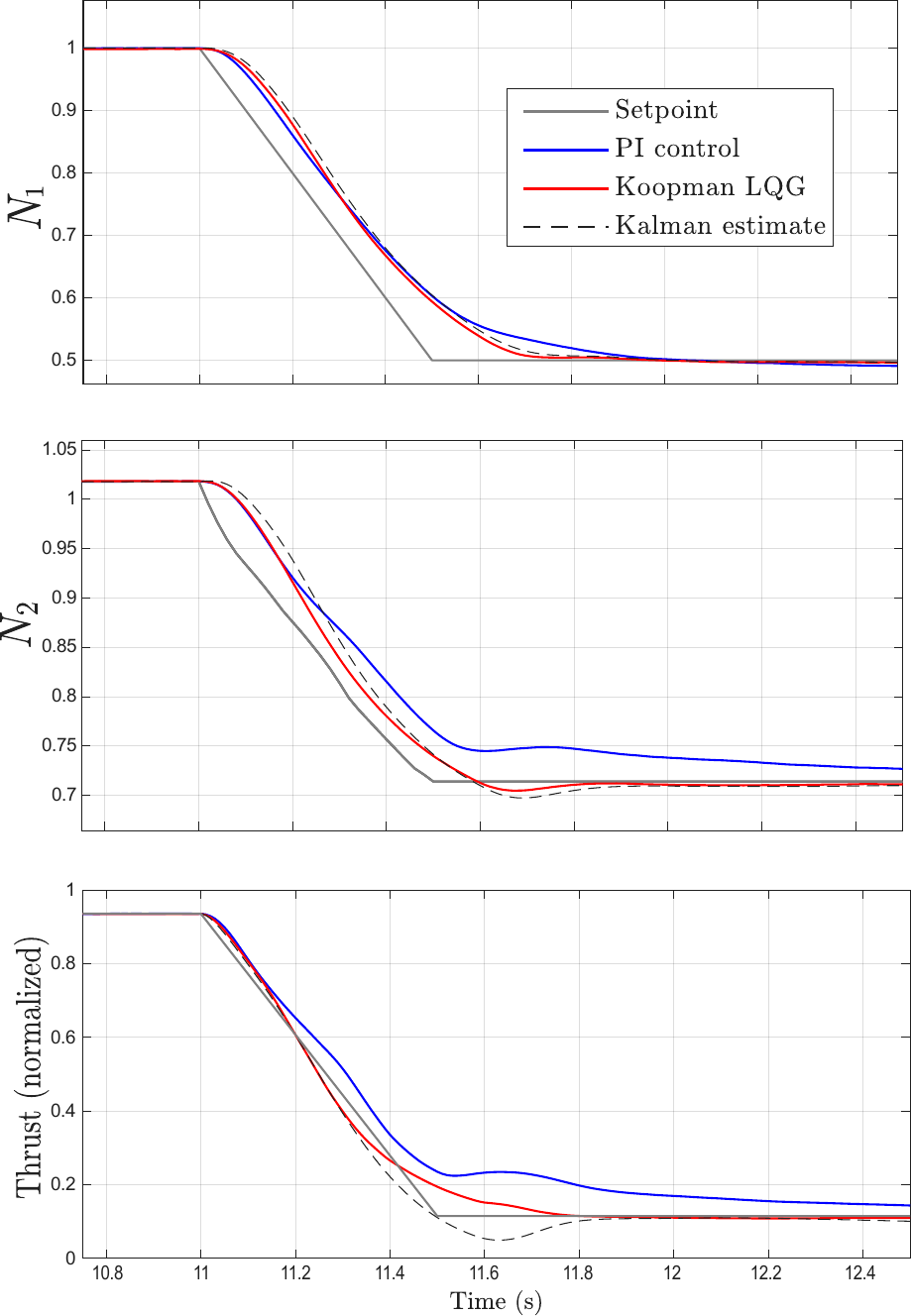}
        \caption{}
        \label{2spool_LQR_Decel}
    \end{subfigure}
    \caption{Example of tracking LQG (red) performance compared to a PI controller (blue) for (a) acceleration and (b) deceleration of the two-spool GTE. The black dashed lines correspond to the KF estimates for $N_1$ and $N_2$, and to the Koopman prediction of thrust.}
    \label{2spool_LQR_perf}
\end{figure*}

The responses are shown in Fig.~\ref{2spool_LQR_perf}, and quantitative metrics are summarized in Table~\ref{tab_ctrl}, with the integral of absolute error (IAE) computed as 
\begin{align}
    \mathrm{IAE}_y(t) = \int_0^t {| y(\tau) - y_\mathrm{set}(\tau) | \mathrm{d}\tau}\,,
\end{align}
where $y_\mathrm{set}$ is the setpoint.

\begin{figure*}
    \centering 
    \begin{subfigure}{0.45\textwidth}
        \centering
        \includegraphics[width=0.9\linewidth]{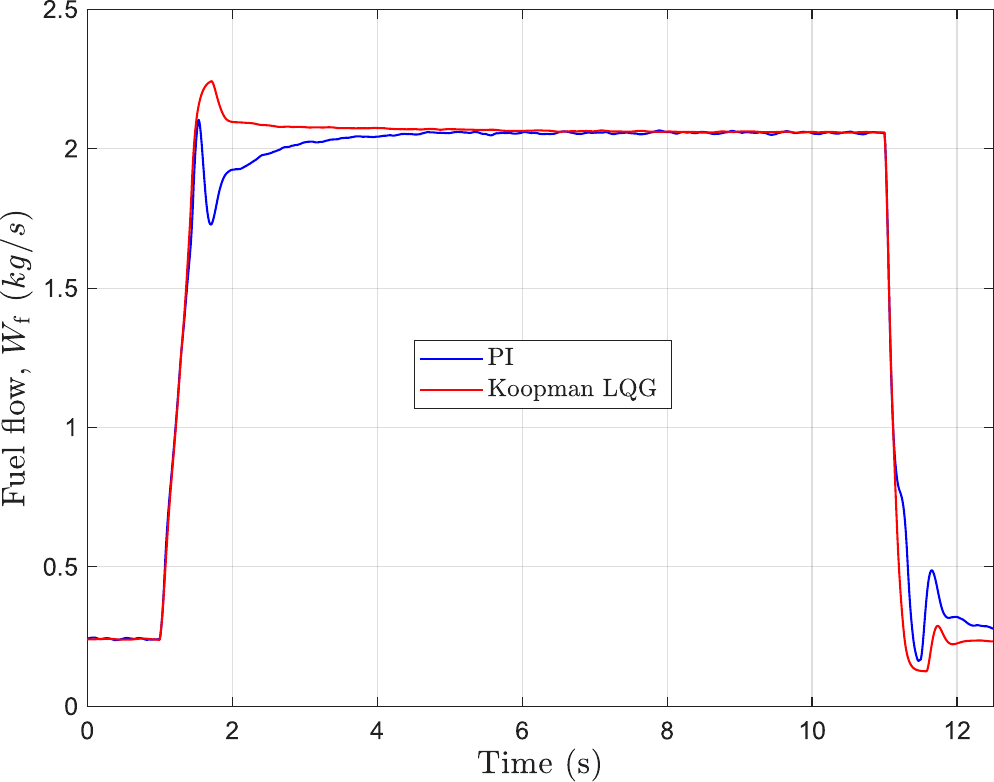}
        \caption{}
        \label{2spool_LQR_inputs_Wf}
    \end{subfigure}
    \begin{subfigure}{0.45\textwidth}
        \centering
        \includegraphics[width=0.9\linewidth]{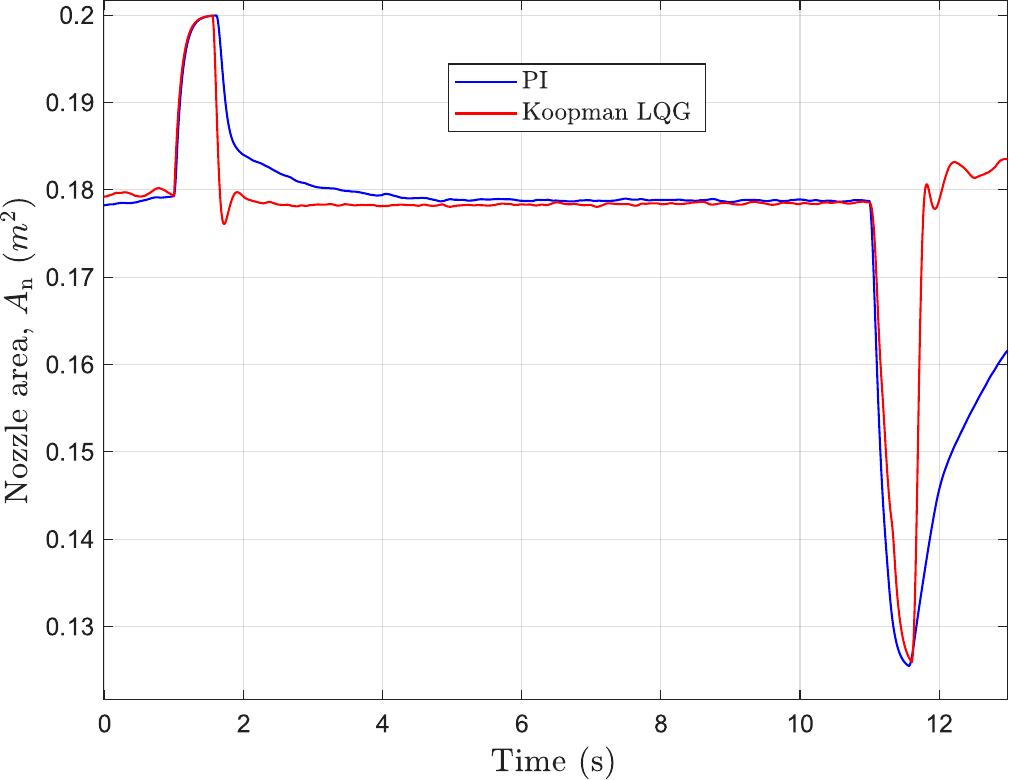}
        \caption{}
        \label{2spool_LQR_inputs_An}
    \end{subfigure}
    \caption{Comparison of (a) the fuel flow and (b) the nozzle area commanded by both controllers. The LQG controller effectively utilized both inputs compared to the PI controller.}
    \label{2spool_LQR_inputs}
\end{figure*}

The LQG exhibited lower IAE and reduced the thrust settling time from 1.55~s to 0.79~s during acceleration and from 2.02~s to 0.70~s during deceleration, while maintaining the absolute thrust OS below 2.5~\%. Fig.~\ref{2spool_LQR_inputs} shows the corresponding input commands. The LQG coordinates both control inputs, whereas the independent PI loops do not explicitly account for cross-coupling, necessitating additional decoupling measures and increasing the overall design complexity. The Kalman estimates of both spool speeds closely follow the true quantities. The reconstructed thrust is accurate during acceleration but shows larger error near the minimum operating condition during deceleration.

\begin{table}[!ht]
\centering
\caption{Comparison of PI and Koopman gain-scheduled tracking LQG controllers. The thrust settling times were evaluated using a $\pm2\,\%$ band relative to the thrust step magnitude.}\label{tab_ctrl}
\begin{tabular}{@{}p{0.30\columnwidth}p{0.25\columnwidth}p{0.25\columnwidth}@{}}
\toprule
Case & PI & LQG \\
\midrule
$\mathrm{IAE}_{N1}$ & 1400 RPM\,s & 1200 RPM\,s \\
$\mathrm{IAE}_{N2}$ & 1756 RPM\,s & 821.5 RPM\,s \\
$\mathrm{IAE}_{F}$  & 23077 F\,s  & 13684 F\,s  \\
\midrule
$t_{\pm2\,\%\,\Delta F}^{acc}$ & 1.55 s & 0.79 s \\
$t_{\pm2\,\%\,\Delta F}^{dec}$ & 2.02 s & 0.70 s \\
$F$ abs. OS   & 0.00 $\%$  & 2.46 $\%$  \\
$N_1$ abs. OS & 2.30 $\%$  & 1.23 $\%$  \\
$N_2$ abs. OS & 0.00 $\%$  & 3.00 $\%$  \\
\bottomrule
\end{tabular}
\end{table}

\section{Conclusion} \label{sec_conc}
This paper introduced SADFED, a dictionary-free framework for data-driven identification of Koopman eigenfunctions. The method uses a reference trajectory $\mathbf{x_{ref}}$ to define fixed reference Koopman modes $\mathbf{C_{ref}}$ and a transformed temporal basis $\mathbf{B}$. For candidate eigenvalue components $\alpha$ and $\beta$ collected in $\boldsymbol{\theta}$, the eigenfunction initial conditions $\mathbf{\Phi_0}$ are obtained through regularized LS, reducing the outer nonlinear optimization only to the eigenvalues. The temporal reconstruction cost $\mathcal{J}_\mathrm{temp}$ is combined with a numerically evaluated KPDE cost $\mathcal{J}_\mathrm{KPDE}$ to promote consistency of the identified eigenfunctions with the Koopman generator over the sampled state-space region. The eigenfunctions are represented numerically rather than through a prescribed dictionary, kernel, or neural-network architecture, making the core identification completely dictionary-free.

The paper also provides practical guidelines for the application of the SADFED. Reference trajectories should sufficiently excite the relevant modes and represent both the transient and the asymptotic/oscillatory dynamics, while the regularization and KPDE weights should be selected to balance numerical conditioning, temporal reconstruction, and KPDE consistency. These guidelines remain heuristic and problem-dependent in nature.

SADFED is expected to work best for systems with a point spectrum and an accurate low-rank Koopman representation, provided that the investigated state-space region is sufficiently sampled and the drift vector field is known or can be estimated. The method remains a nonconvex numerical identification procedure and therefore does not guarantee global convergence or uniqueness of the identified eigenvalues. Its computational cost increases with the state-space dimension because of trajectory sampling, interpolation, and numerical differentiation, while continuous spectra, strongly chaotic dynamics, noisy derivative estimates, and extrapolation beyond the sampled region remain challenging. Moreover, a small KPDE residual establishes generator consistency only over the sampled region.

The numerical examples demonstrated different aspects of the framework. The system with a quadratic manifold verified recovery of known analytical Koopman eigenfunctions, while the FitzHugh-Nagumo example demonstrated handling nontrivial fixed points, sensitivity analyses, construction of lifted input dynamics from the eigenfunction gradient, and comparison with EDMDc. For the van der Pol oscillator, SADFED obtained the five prominent limit-cycle harmonic modes and provided approximate isochrons. The Duffing example demonstrated identification across multiple basins of attraction, a zero-eigenvalue indicator eigenfunction, isostables, and exploitation of state-space symmetry. Finally, the two-spool turbojet example illustrated the complete modeling chain from an identified control-affine proxy model to Koopman eigenfunctions, gradient-based lifted input dynamics, state estimation, and gain-scheduled LQG control.

Future work will focus on automated selection of reference trajectories, extension to continuous spectra, improved numerical representation of eigenfunctions and their gradients, and further computational improvements for moderately higher-dimensional systems. Further investigation of observer and control design based on the resulting Koopman models, including broader applications to gas-turbine engines, is ongoing.

\backmatter

\bmhead{Acknowledgements}

\bmhead {Author contribution}
The author confirms sole responsibility for the following: study conception and design, simulation and data collection, analysis and interpretation of results, and manuscript preparation.

\bmhead {Funding}
The author declares that this research was supported by the infrastructure of the University of Defence, Brno, Czech Republic, within the framework of DZRO-FVT22-AIROPS "Long Term Organization Development Plan - Conduction of airspace operations" of the Ministry of Defence of the Czech Republic, and by the Student Research Program SV 26-206/1 of the Ministry of Education, Youth and Sports of the Czech Republic.

\bmhead{Data availability}
Codes and datasets supporting the results can be found in the supplementary materials. Additional materials will be made available upon reasonable request.

\section*{Declarations} \label{sec_decl}

\bmhead {Conflict of interest}
The author has no conflicts of interest and no relevant financial or non-financial interests to disclose.

\begin{appendices}

\section{Riccati equations for the gain-scheduled LQG design}
\label{sec_A1}

This appendix summarizes the Riccati equations used in the estimator and controller design described in Sections~\ref{sec_kalman} and~\ref{sec_GS_LQG}.

For the steady-state Kalman filter in the lifted coordinates, the observer gain is
\begin{equation}
    \mathbf{L = P_o C^\intercal R_o}^{-1},
\end{equation}
where $\mathbf{P}_\mathbf{o}$ is a solution of the continuous-time observer algebraic Riccati equation~\cite{Friedland1987}
\begin{equation}
    \mathbf{\Lambda P_o + P_o \Lambda^\intercal + G_o Q_o G_o^\intercal - P_o C^\intercal R_o^{-1} C P_o = 0},
\end{equation}
where $\mathbf{G_o = I}$, and $\mathbf{R_o}$ and $\mathbf{Q_o}$ are tuned based on the effective covariance matrices described in Section~\ref{sec_kalman}.

For controller design, the lifted and actuator dynamics are frozen at each scheduling point $\mathbf{x}^{(s)}$, yielding
\begin{align}
    \mathbf{A}^{(s)} &= \begin{bmatrix} \mathbf{\Lambda} & \mathbf{\Gamma}(\mathbf{x}^{(s)}) \\ \mathbf{0} & \mathbf{{A}_{act}}
    \end{bmatrix}\,, & \mathbf{B_a} &= \begin{bmatrix} \mathbf{0} \\ \mathbf{B_{act}} \end{bmatrix}\,.
\end{align}
The corresponding LQR gain is
\begin{equation} \label{SUP-4}
    \mathbf{K}^{(s)} = \mathbf{R_c}^{-1} \mathbf{B_a}^\intercal \mathbf{P}_\mathbf{c}^{(s)}\,,
\end{equation}
where $\mathbf{P}_\mathbf{c}^{(s)}$ is a solution of the continuous-time controller algebraic Riccati equation
\begin{equation}
\begin{aligned}
    (\mathbf{A}^{(s)})^\intercal \mathbf{P}_\mathbf{c}^{(s)} & + \mathbf{P}_\mathbf{c}^{(s)} \mathbf{A}^{(s)} + \mathbf{Q_c} \\ &- \mathbf{P}_\mathbf{c}^{(s)} \mathbf{B_a} \mathbf{R_c}^{-1} \mathbf{B_a}^\intercal \mathbf{P}_\mathbf{c}^{(s)} = \mathbf{0}\,,
\end{aligned}
\end{equation}
where $\mathbf{Q_c}$ and $\mathbf{R_c}$ are the augmented state and input weighting matrices. The resulting gains $\mathbf{K}^{(s)}$ are approximated over the scheduling region using the sigmoid representation given in Section~\ref{sec_GS_LQG}.

\end{appendices}

\bibliography{bibliography}

@book{MattiglyBook,
    author		= "J. D. Mattingly and K. M. Boyer",
    title			= "Elements of Propulsion: Gas Turbines and Rockets, Second Edition",
    address		= "Reston, Virginia",
    publisher		= "American Institute for Aeronautics and Astronautics",
    year			= "2016"
}

@article{Koopman1931,
    ISSN = {00278424, 10916490},
    URL = {http://www.jstor.org/stable/86114},
    author = {B. O. Koopman},
    journal = {Proceedings of the National Academy of Sciences of the United States of America},
    number = {5},
    pages = {315--318},
    publisher = {National Academy of Sciences},
    title = {{Hamiltonian Systems and Transformations in Hilbert Space}},
    volume = {17},
    year = {1931}
}

@article{Koopman1932,
    ISSN = {00278424, 10916490},
    URL = {http://www.jstor.org/stable/86259},
    author = {B. O. Koopman and J. {von Neumann}},
    journal = {Proceedings of the National Academy of Sciences of the United States of America},
    number = {3},
    pages = {255--263},
    publisher = {National Academy of Sciences},
    title = {{Dynamical Systems of Continuous Spectra}},
    volume = {18},
    year = {1932}
}

@book{Mauroy2020,
    editor		= "A. Mauroy and I. Mezić and Y. Susuki",
    title			= "The Koopman Operator in Systems and Control",
    address		    = "Cham, Switzerland",
    publisher		= "Springer Nature",
    year			= "2020",
    issn            = "0170-8643",
    isbn            = "978-3-030-35713-9",
    doi             = "10.1007/978-3-030-35713-9"
}

@book{BruntonBook2019, 
    address={Cambridge}, 
    title={Data-Driven Science and Engineering: Machine Learning, Dynamical Systems, and Control}, 
    publisher={Cambridge University Press}, 
    author={Brunton, Steven L. and Kutz, J. Nathan}, 
    year={2019}
}

@article{Brunton2021,
    author = {Brunton, Steven L. and Budi\v{s}i\'{c}, Marko and Kaiser, Eurika and Kutz, J. Nathan},
    title = {Modern {Koopman} Theory for Dynamical Systems},
    journal = {SIAM Review},
    volume = {64},
    number = {2},
    pages = {229--340},
    year = {2022},
    doi = {10.1137/21M1401243},
}

@article{Kutz2018-EDMD-PDEs,
    author = {Nathan Kutz, J. and Proctor, J. L. and Brunton, S. L.},
    title = {Applied Koopman Theory for Partial Differential Equations and Data-Driven Modeling of Spatio-Temporal Systems},
    journal = {Complexity},
    volume = {2018},
    number = {1},
    pages = {6010634},
    doi = {https://doi.org/10.1155/2018/6010634},
    year = {2018}
}

@article{Otto2019-AE-PDEs,
    author = {Otto, Samuel E. and Rowley, Clarence W.},
    title = {Linearly Recurrent Autoencoder Networks for Learning Dynamics},
    journal = {SIAM Journal on Applied Dynamical Systems},
    volume = {18},
    number = {1},
    pages = {558-593},
    year = {2019},
    doi = {10.1137/18M1177846},
}

@article{Brunton2016InvSubs,
    doi = {10.1371/journal.pone.0150171},
    author = {Brunton, Steven L. AND Brunton, Bingni W. AND Proctor, Joshua L. AND Kutz, J. Nathan},
    journal = {PLOS ONE},
    publisher = {Public Library of Science},
    title = {Koopman Invariant Subspaces and Finite Linear Representations of Nonlinear Dynamical Systems for Control},
    year = {2016},
    month = {02},
    volume = {11},
    pages = {1--19},
    number = {2},
}

@article{Surana2020,
    title = {Koopman {Operator} {Framework} for {Time} {Series} {Modeling} and {Analysis}},
    volume = {30},
    issn = {1432-1467},
    doi = {10.1007/s00332-017-9441-y},
    number = {5},
    journal = {Journal of Nonlinear Science},
    author = {Surana, Amit},
    month = oct,
    year = {2020},
    pages = {1973--2006},
}

@article{Iacob2024,
title = {Koopman form of nonlinear systems with inputs},
journal = {Automatica},
volume = {162},
pages = {111525},
year = {2024},
issn = {0005-1098},
doi = {10.1016/j.automatica.2024.111525},
author = {Lucian Cristian Iacob and Roland Tóth and Maarten Schoukens}
}

@article{GrasevAccess2025,
    author={Grasev, David},
    title={{Data-Driven Identification of Gas Turbine Engine Dynamics via Koopman Operator Genetic Algorithm}}, 
    journal={IEEE Access},
    volume={13},
    number={1},
    pages={91972-91988},
    year={2025},
    doi={10.1109/ACCESS.2025.3573472}
}

@article{GrasevSpringer2026,
	title = {Koopman eigenfunction-based identification and optimal nonlinear control of turbojet engine},
	volume = {114},
	issn = {1573-269X},
	doi = {10.1007/s11071-025-12070-7},
	number = {3},
	journal = {Nonlinear Dynamics},
	author = {Grasev, David},
	month = feb,
	year = {2026},
	pages = {205},
}

@article{Korda2020,
    author={Korda, Milan and Mezić, Igor},
    journal={IEEE Transactions on Automatic Control}, 
    title={{Optimal Construction of Koopman Eigenfunctions for Prediction and Control}}, 
    year={2020},
    volume={65},
    number={12},
    pages={5114-5129},
    doi={10.1109/TAC.2020.2978039}
}

@article{Williams2015,
    author =    {M.O. Williams and I.G. Kevrekidis and C.W. Rowley},
    title = {{A Data–Driven Approximation of the Koopman Operator: Extending Dynamic Mode Decomposition}},
    journal = {J Nonlinear Sci},
    volume = {25},
    pages = {1307–-1346},
    year = {2015},
    doi = {10.1007/s00332-015-9258-5},
}

@article{Williams2015-2,
    author =    {Matthew O. Williams and Clarence W. Rowley and Ioannis G. Kevrekidis},
    title = {A kernel-based method for data-driven {Koopman} spectral analysis},
    journal = {Journal of Computational Dynamics},
    volume = {2},
    number = {2},
    pages = {247--265},
    year = {2015},
    doi = {10.3934/jcd.2015005},
}

@inproceedings{Jiang2022,
    author={Jiang, Wei and Zhang, Xing long and Zuo, Zhen and Shi, Meiping and Su, Shaojing},
    booktitle={2022 IEEE/RSJ International Conference on Intelligent Robots and Systems (IROS)}, 
    title={{Data-driven Kalman Filter with Kernel-based Koopman Operators for Nonlinear Robot Systems}}, 
    year={2022},
    volume={},
    number={},
    pages={12872-12878},
    doi={10.1109/IROS47612.2022.9981408}
}

@misc{Ishikawa2024,
    title={{Koopman operators with intrinsic observables in rigged reproducing kernel Hilbert spaces}}, 
    author={Isao Ishikawa and Yuka Hashimoto and Masahiro Ikeda and Yoshinobu Kawahara},
    year={2024},
    archivePrefix={arXiv},
    primaryClass={math.DS},
    doi = {10.48550/arXiv.2403.02524}
}

@article{Jovanovich2014,
    title = {Sparsity-promoting dynamic mode decomposition},
    volume = {26},
    issn = {1070-6631},
    doi = {10.1063/1.4863670},
    number = {2},
    journal = {Physics of Fluids},
    author = {Jovanović, Mihailo R. and Schmid, Peter J. and Nichols, Joseph W.},
    month = feb,
    year = {2014},
    pages = {024103},
}

@article{Schlosser2022,
    author = {Schlosser, Corbinian and Korda, Milan},
    title = {{Sparsity Structures for Koopman and Perron--Frobenius Operators}},
    journal = {SIAM Journal on Applied Dynamical Systems},
    volume = {21},
    number = {3},
    pages = {2187--2214},
    year = {2022},
    doi = {10.1137/21M1466608}
}

@article{Klus2020,
    author =     {Stefan Klus and Feliks Nüske and Sebastian Peitz and Jan-Hendrik Niemann and Cecilia Clementi and Christof Schütte},
    title = {{Data-driven approximation of the Koopman generator: Model reduction, system identification, and control}},
    journal = {Physica D: Nonlinear Phenomena},
    volume = {406},
    pages = {132416},
    year = {2020},
    issn = {0167-2789},
    doi = {10.1016/j.physd.2020.132416},
}

@article{Colbrook2023,
    title={{Residual dynamic mode decomposition: robust and verified Koopmanism}}, 
    volume={955}, 
    DOI={10.1017/jfm.2022.1052}, 
    journal={Journal of Fluid Mechanics}, 
    author={Colbrook, Matthew J. and Ayton, Lorna J. and Szőke, Máté}, 
    year={2023}, 
    pages={A21}
}

@article{Pan2024,
    title = {On the lifting and reconstruction of nonlinear systems with multiple invariant sets},
    volume = {112},
    issn = {1573-269X},
    doi = {10.1007/s11071-024-09581-0},
    number = {12},
    journal = {Nonlinear Dynamics},
    author = {Pan, Shaowu and Duraisamy, Karthik},
    month = jun,
    year = {2024},
    pages = {10157--10165},
}

@article{Proctor2018,
    author = {Proctor, Joshua L. and Brunton, Steven L. and Kutz, J. Nathan},
    title = {{Generalizing Koopman Theory to Allow for Inputs and Control}},
    journal = {SIAM Journal on Applied Dynamical Systems},
    volume = {17},
    number = {1},
    pages = {909--930},
    year = {2018},
    doi = {10.1137/16M1062296},
}

@article{BruntonHAVOK,
    title = {Chaos as an intermittently forced linear system},
    volume = {8},
    issn = {2041-1723},
    doi = {10.1038/s41467-017-00030-8},
    number = {1},
    journal = {Nature Communications},
    author = {Brunton, Steven L. and Brunton, Bingni W. and Proctor, Joshua L. and Kaiser, Eurika and Kutz, J. Nathan},
    month = may,
    year = {2017},
    pages = {19},
}

@ARTICLE{Li2024HAVOK,
    author={Li, Xiuting and Zhang, Hai-Tao and Huang, Xiang and Li, Linlin and Zhu, Li-Min and Ding, Han and Yuan, Ye},
    journal={IEEE Transactions on Systems, Man, and Cybernetics: Systems}, 
    title={{Data-Driven Koopman Learning and Prediction of Piezoelectric Tube Scanner Hysteresis}}, 
    year={2024},
    volume={54},
    number={6},
    pages={3631-3641},
    doi={10.1109/TSMC.2024.3368570}
}

@inproceedings{Li2020,
    title={{Learning Compositional Koopman Operators for Model-Based Control}},
    author={Yunzhu Li and Hao He and Jiajun Wu and Dina Katabi and Antonio Torralba},
    booktitle={International Conference on Learning Representations},
    year={2020},
    doi={10.48550/arXiv.1910.08264}
}

@article{Takeishi2017,
    author       = {Naoya Takeishi and
                  Yoshinobu Kawahara and
                  Takehisa Yairi},
    title        = {{Learning Koopman Invariant Subspaces for Dynamic Mode Decomposition}},
    year         = {2017},
    doi          = {10.48550/arXiv.1710.04340}
}

@article{Jin2024,
    title = {{Extended Dynamic Mode Decomposition with Invertible Dictionary Learning}},
    journal = {Neural Networks},
    volume = {173},
    pages = {106177},
    year = {2024},
    issn = {0893-6080},
    doi = {10.1016/j.neunet.2024.106177},
    author = {Yuhong Jin and Lei Hou and Shun Zhong}
}

@article{Lusch2018,
    author =    {B. Lusch and J.N. Kutz and Steven L. Brunton},
    title = {Deep learning for universal linear embeddings of nonlinear dynamics},
    journal = {Nat Commun},
    volume = {9},
    pages = {4950},
    year = {2018},
    doi = {10.1038/s41467-018-07210-0},
}

@article{Kaiser2021,
    doi = {10.1088/2632-2153/abf0f5},
    year = {2021},
    month = {jun},
    publisher = {IOP Publishing},
    volume = {2},
    number = {3},
    pages = {035023},
    author = {Kaiser, Eurika and Kutz, J Nathan and Brunton, Steven L},
    title = {Data-driven discovery of {Koopman} eigenfunctions for control},
    journal = {Machine Learning: Science and Technology},
}

@article{Mezic2012,
    author = "Mezić, Igor",
    title = {{Analysis of Fluid Flows via Spectral Properties of the Koopman Operator}}, 
    journal= "Annual Review of Fluid Mechanics",
    year = "2013",
    volume = "45",
    pages = "357--378",
    doi = "https://doi.org/10.1146/annurev-fluid-011212-140652",
    publisher = "Annual Reviews",
    issn = "1545-4479",
    type = "Journal Article",
}

@article{Mezic2020_Spectral,
    author =    {Igor Mezić},
    title = {{Spectrum of the Koopman Operator, Spectral Expansions in Functional Spaces, and State-Space Geometry}},
    journal = {J Nonlinear Sci},
    volume = {30},
    pages = {2091–-2145},
    year = {2020},
    doi = {https://doi.org/10.1007/s00332-019-09598-5},
}

@article{Mauroy2012,
    author = {Mauroy, A. and Mezić, I.},
    title = {{On the use of Fourier averages to compute the global isochrons of (quasi)periodic dynamics}},
    journal = {Chaos: An Interdisciplinary Journal of Nonlinear Science},
    volume = {22},
    number = {3},
    pages = {033112},
    year = {2012},
    month = {07},
    issn = {1054-1500},
    doi = {10.1063/1.4736859},
}

@article{Mauroy2013,
    title = {{Isostables, isochrons, and Koopman spectrum for the action–angle representation of stable fixed point dynamics}},
    journal = {Physica D: Nonlinear Phenomena},
    volume = {261},
    pages = {19--30},
    year = {2013},
    issn = {0167-2789},
    doi = {https://doi.org/10.1016/j.physd.2013.06.004},
    author = {A. Mauroy and I. Mezić and J. Moehlis},
}

@misc{Liu2022,
    title={{Physics-Informed Koopman Network}}, 
    author={Yuying Liu and Aleksei Sholokhov and Hassan Mansour and Saleh Nabi},
    year={2022},
    archivePrefix={arXiv},
    primaryClass={cs.LG},
    doi = {https://doi.org/10.48550/arXiv.2211.09419}
}

@article{Cibulka2025,
    author = {Cibulka, V\'{\i}t and Korda, Milan and Hani\v{s}, Tom\'{a}\v{s}},
    title = {{Dictionary-Free Koopman Model Predictive Control with Nonlinear Input Transformation}},
    journal = {SIAM Journal on Applied Dynamical Systems},
    volume = {24},
    number = {1},
    pages = {94--130},
    year = {2025},
    doi = {10.1137/23M1597174}
}

@inproceedings{Kennedy1995,
    author={Kennedy, J. and Eberhart, R.},
    booktitle={Proceedings of ICNN'95 - International Conference on Neural Networks}, 
    title={Particle swarm optimization}, 
    year={1995},
    volume={4},
    pages={1942-1948},
    doi={10.1109/ICNN.1995.488968}
}

@article{Nelder1965,
  title={{A Simplex Method for Function Minimization}},
  author={John A. Nelder and Roger Mead},
  journal={Comput. J.},
  year={1965},
  volume={7},
  pages={308-313},
  doi = {10.1093/COMJNL/7.4.308}
}

@book{Jaw2009,
    author = {L. C. Jaw and J. D. Mattingly},
    title = {Aircraft Engine Controls: Design, System Analysis, And Health Monitoring},
    publisher = {Amer Inst Aero \& Astro},
    year = {2009},
    address = {Reston, Virginia}
}

@manual{GasTurb,
  title        = {GasTurb 15: Design and Off-Design Performance of Gas Turbines},
  author       = {Joachim Kurzke},
  year         = 2026,
  address      = {Aachen, Germany},
  note         = {Available at \url{https://www.gasturb.com/Downloads/Manuals/GasTurb15.pdf}},
  organization = {GasTurb GmbH},
}

@misc{Kingma2015,
    title={{Adam: A Method for Stochastic Optimization}}, 
    author={Diederik P. Kingma and Jimmy Ba},
    year={2015},
    archivePrefix={arXiv},
    primaryClass={cs.LG},
    doi = {10.48550/arXiv.1412.6980}
}

@book{Friedland1987,
    author		= "Bernard Friedland",
    title		= {Control System Design: An Introduction to State-Space Methods},
    subtitle    = "An Introduction to State-Space Methods",
    address		= "Mineola, NY",
    publisher	= "Dover Publications",
    year		= "1987"
}

@article{Annaswamy2021,
    title = {A historical perspective of adaptive control and learning},
    journal = {Annual Reviews in Control},
    volume = {52},
    pages = {18--41},
    year = {2021},
    issn = {1367-5788},
    doi = {https://doi.org/10.1016/j.arcontrol.2021.10.014},
    author = {Anuradha M. Annaswamy and Alexander L. Fradkov}
}

\end{document}